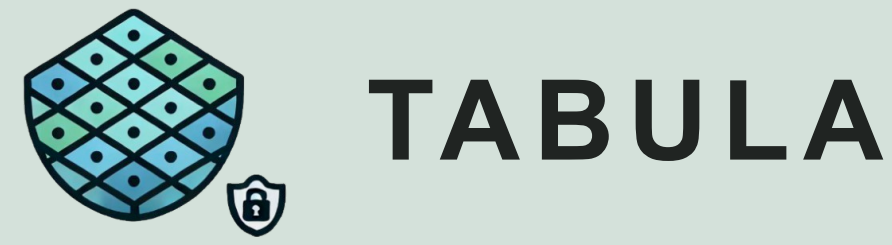

# Predictive single cell foundation model for gene regulation and aging with privacy-preserving tabular learning

Jiayuan Ding[1,2,$,*], Jianhui Lin[3,$], Ziyang Miao[3,$], Nils Mechtel[4,5,$], Shiyu Jiang[1], Yixin Wang[6], Zhaoyu Fang[3], Jorge D. Martin-Rufino[7,8], Chen Weng[9], Reuben Saunders[9], Weize Xu[1], Jonathan S. Weissman[9,10,11], Min Li[3,*], Jiliang Tang[2,*], Wei Ouyang[4,5,*], Yuancheng Ryan Lu[9,*], Xiaojie Qiu[1,12,13,14,*]

[$]These authors contributed equally to this work.

**Corresponding authors**: Jiayuan Ding (dingjia5@msu.edu), Min Li (limin@mail.csu.edu.cn), Jiliang Tang (tangjili@msu.edu), Wei Ouyang (weio@kth.se), Yuancheng Ryan Lu (ylu@wi.mit.edu), Xiaojie Qiu (xiaojie@stanford.edu)

## Abstract

Pre-trained foundation models (FMs) have begun transforming single-cell genomics, but scaling them raises privacy concerns. Moreover, unlike text data, single-cell data is unordered and exhibits a unique tabular structure that current single-cell FMs overlook. We introduce Tabula, a privacy-preserving FM designed with federated learning (FL) that explicitly models the tabular structure of single-cell data. To deploy Tabula, we further developed Chiron, a decentralized AI agent-enabled platform for collaborative training across institutions without sharing raw data. Beyond strong performance across downstream benchmarks, Tabula reveals combinatorial regulatory logic across diverse biological systems, including hematopoiesis, pancreatic endogenesis, neurogenesis, and cardiogenesis. Using a new scRNA-seq dataset of paired young and aged human fibroblasts, Tabula nominates rejuvenation factors through age- and identity score-guided *in silico* prioritization, outperforming conventional approaches. Thus, Tabula represents an important advance in single-cell foundation modeling by integrating tabular learning with FL, paving the way toward privacy-preserving virtual cells for human health.

**Code**: github.com/aristoteleo/tabula **Platform**: chiron.aicell.io

## Affiliations

[1]Department of Genetics, Stanford University, Stanford, CA, USA. [2]Department of Computer Science and Engineering, Michigan State University, East Lansing, MI, USA. [3]School of Computer Science and Engineering, Central South University, Changsha, China. [4]Department of Applied Physics, KTH Royal Institute of Technology, Stockholm, Sweden. [5]Science for Life Laboratory (SciLifeLab), KTH Royal Institute of Technology, Stockholm, Sweden. [6]Department of Bioengineering, Stanford University, Stanford, CA, USA. [7]Division of Hematology/Oncology, Boston Children's Hospital and Department of Pediatric Oncology, Dana-Farber Cancer Institute, Harvard Medical School, Boston, MA, USA. [8]Broad Institute of MIT and Harvard, Cambridge, MA, USA. [9]Whitehead Institute for Biomedical Research, Cambridge, MA, USA. [10]Department of Biology and Howard Hughes Medical Institute, Massachusetts Institute of Technology, Cambridge, MA, USA. [11]David H. Koch Institute for Integrative Cancer Research and Massachusetts Institute of Technology, Cambridge, MA, USA. [12]Basic Sciences and Engineering Initiative, Betty Irene Moore Children's Heart Center, Lucile Packard Children's Hospital, Stanford, CA, USA. [13]Department of Computer Science, Stanford University, Stanford, CA, USA. [14]Stanford Cardiovascular Institute, Stanford University, Stanford, CA, USA.

# Introduction

One remarkable mystery in the multicellular system is how the same genome can generate a vast diversity of cell types and states during development, disease and aging. Distinct gene expression patterns and complex pairwise or higher-order regulatory interactions underlie such diversities. Imagine each cell state as a point in the transcriptomic state space, where each dimension corresponds to a gene[1]. We reason that the feasible space of cell states in any organism doesn't occupy the entire state space but is generally considered to lie on a manifold, shaped by the governing regulatory networks.

Although genome-wide measurement of regulatory networks remains largely elusive, rapid advances in single-cell genomics[2] and collaborative consortium efforts, such as CZI, the Human Cell Atlas[3], and HuBMAP[4], have enabled us to map a significant portion of this single-cell manifold across human biology. Recent breakthroughs in natural language processing (NLP)[5–7], computer vision[8,9], and, more recently, protein language models[10–12] have shown that attention-based Transformers[13] can learn the fundamental structure of data through unsupervised self-learning on large datasets, rather than relying on explicit modeling of grammar, visual geometry, biophysical forces. The accumulation of the massive single cell data thus promises a new era of human biology, enabling the development of single cell foundation models to gain fundamental understanding of gene regulation—critical for virtually all cellular processes.

Over the past three years, we have witnessed the emergence of single cell foundation models, notably Geneformer[14], scFoundation[15], CellPLM[16], scGPT[17], UCE[18], and NicheFormer[19]. However, current models fail to explicitly account for the tabular structure of the scRNA-seq data, often naively converting gene expression within a single cell to gene sequences to mimic word sequences used in NLP. While this adapts the NLP paradigm to single cell data, it overlooks critical features of the underlying data structure of scRNA-seq. Therefore, there is an urgent need to develop novel pretraining approaches to explicitly model the tabular structure of single cell data to enhance the foundation model pretraining. In addition, it has been shown previously that scRNA-seq data is susceptible to privacy breaches through linking attacks[20,21], where information from one study can be linked to another to identify private data. This privacy concern becomes even more pronounced as we begin creating foundation models with tens of millions of single cells and datasets from thousands of individuals or tens of institutes.

To address these critical gaps, we developed Tabula, a privacy-preserving federated foundation model tailored for scRNA-seq data through tabular learning. Tabula delivers three major innovations. First, it introduces federated learning (FL)[22] into the foundation model training, allowing for collaborative model training across multiple clients without compromising the privacy of the individual data. The distributed computing framework also significantly reduces local training cost. Furthermore, Tabula combines a shared global transformer with client-specific components to capture both cross-client and tissue-specific features, significantly improving zero-shot performance on client-specific downstream tasks. Second, we developed Chiron, a decentralized platform that realizes federated training, enabling institutions to collaboratively train, monitor, publish, and reuse single-cell foundation models and potentially others through a unified infrastructure without sharing raw data. Third, unlike earlier single cell foundation models[14,15,17,23], which treated single cell data analogous to natural language data (e.g., viewing cells as sentences and genes as words), Tabula redefines this paradigm by explicitly modeling the tabular structure of single cell data with tabular learning. Extensive benchmarking demonstrates that Tabula consistently outperform state-of-the-art foundation models on conventional gene-wise and cell-wise downstream tasks, including cell type annotation, batch correction, gene imputation, single- and double-perturbation response prediction and reverse perturbation predictions[24]. Importantly, we show Tabula accurately recovers pairwise and combinatorial gene regulations across diverse biological systems, including hematopoiesis, pancreatic endogenesis, neurogenesis and cardiogenesis where extensive biochemical studies have established ground-truth core regulatory networks. Furthermore, we show that Tabula moves beyond conventional genomics benchmarks to enable therapeutic discovery, nominating candidate rejuvenation factors for human skin aging through age- and identity-guided in silico prioritization. These results represent a significant step toward building a mechanistic and predictive single cell foundation model for understanding fundamental mechanisms of gene regulation and human health. Tabula and our decentralized training platform Chiron can be freely accessible at https://github.com/aristoteleo/tabula and https://chiron.aicell.io.

# RESULTS

## Tabula redefines single cell foundation models through tabular and federated learning

Pioneer single cell foundation models such as Geneformer[14] or scGPT[17] show promises of LLM in advancing single cell biology. However, these frameworks and recent new developments, e.g. UCE[18], CellPLM[16], NicheFormer[25] etc., rely on centralized learning that trains a unified model directly on aggregated data, thus raising concerns about data privacy and ethics[20,21]. Furthermore, although it is obvious that single cell data doesn't have an intrinsic order, most of these approaches nevertheless directly borrow conventional large language model techniques originally designed for sequential text data to single cell data after transforming the unordered genes across single cells to gene sequences with some rather naive ordering strategies **(Fig. 1a)**. For example, Geneformer generates a gene pseudo-order by ranking the normalized gene expression within each cell, followed by masked language modeling[5]. Similarly, scGPT first bins genes and then uses attention maps from the learned transformer to create a sequence order by prioritizing genes with high attention scores for next-token prediction using autoregressive modeling[26]. Likewise, tGPT[27] creates a pseudo-order of genes by sorting the normalized gene expression within each cell and employs autoregressive modeling to predict the subsequent gene token.

To explicitly model the tabular structure while preserving data privacy, we propose Tabula, a novel single cell foundation model that integrates tabular and federated learning **(Fig. 1a)**. First, we leverage federated learning[22] to allow Tabula to have different clients and to train a client-specific model locally on private data without sharing raw data across institutions, thus eliminating privacy concerns for potential single cell datasets generated from patient or proprietary samples. Second, instead of treating gene expression data as ordered sequences, we explicitly model its intrinsic tabular structure using self-supervised tabular learning[28]. Specifically, we treat each cell of the gene expression matrix as a permutation-invariant row of genes, where each column corresponds to a specific gene, and the expression value of an element of the matrix comes from the corresponding cell in the row and gene in the column. The column feature can also include metadata, such as tissue type or cell type, providing additional biological context. Tabula leverages a large-scale pretraining dataset comprising 15 million single cells across diverse tissues (**Supplementary Fig. 1a and b**), following a scaling law that demonstrates significant performance improvements as the pretraining dataset size increases (**Supplementary Fig. 1c and d**). Additionally, Tabula can generate tissue-specific embedders that encode distinctive features of individual tissues, allowing for better representation of tissue-specific biological characteristics (**Supplementary Fig. 2**). Unlike sequence-based models that impose an artificial order on genes, Tabula's tabular design treats each gene as an independent attribute, respecting their unordered nature. By incorporating both cell-wise and gene-wise learning (**Methods**), Tabula effectively models both gene and cell relationships of the tabular single cell data. These innovative designs in Tabula necessitate new training loss for tabular data, as well as model architecture and pretraining scheme.

To handle the intrinsic tabular structure of scRNA-seq dataset, Tabula introduces the self-supervised tabular modeling framework that consists of both gene-wise reconstruction learning and cell-wise contrastive learning (**Fig. 1b**). Both training components start with generating a corrupted representation of the scRNA-seq data. For gene-wise reconstruction, 60% of genes in each cell are firstly randomly selected, expression of these selected genes are then replaced with values sampled from each gene's marginal distributions across cells based on a negative binomial distribution (**Methods**). Then gene-wise reconstruction tries to restore original expression values (columns of the *cell by gene* table) from their corrupted views. In contrast, the cell-wise contrastive learning tries to align or separate positive or negative cell pairs in latent spaces, respectively. Specifically, the gene expression data of each cell, including original and corrupted views, is firstly embedded into latent dimension. Then contrastive learning is used to ensure positive cell pairs, representing both views from the same cell, colocalize while negative pairs, representing original or corrupted views from distinct cells, overall separate in the latent space. Therefore, while gene-wise reconstruction ensures robust gene-level representations, the cell-wise contrastive learning ensures both intra-cellular consistency but also inter-cell variability, thus properly modeling the tabular structure of single cell cell-by-gene matrix along both the gene and cell axis.

Additionally, to address the data privacy concern, Tabula integrates federated learning that relies on distributed training across clients without sharing raw data (**Fig. 1c**). Tabula's federated learning framework builds upon a unified transformer with the architecture shared between both the client model and global model but trained separately. Tabula's pretraining leverages large-scale, privacy-protected datasets, each tied to a specific client, such as a hospital, institute or tissue type (e.g., lung, brain, heart), as demonstrated in this study. Raw data remains localized at client sites, while only local tabular transformer weights are shared with a central server to update the global tabular transformer model. Each client, containing a client-specific embedder and projection heads (**Fig. 1c**), transforms cell-by-gene tables into embeddings, which are processed by the tabular transformer to compute gene-wise and cell-wise learning losses. The training process involves iterative updates of the local and global tabular transformers in alternating cycles. At the client level, local transformers learn client-specific, e.g. tissue-specific, patterns which are then uploaded to the central server. At the center level, the global transformer aggregates these uploaded weights to update the global model and then broadcasts the aggregated weights back to local clients to improve the generalized model across clients. This continuous communication between the central server and the client ensures exchange of knowledge across diverse data sources without data sharing, thus preserving data confidentiality. By training on distributed clients using tabular learning loss, Tabula establishes a novel foundation model for single cell data that is privacy-preserving, capable of capturing tissue-specific and cross-tissue biological variations while respecting the intrinsic tabular structure of the scRNA-seq data.

Following pretraining, the global Tabula transformer can be fine-tuned for a broad suite of downstream tasks (**Fig. 1c**). We systematically benchmarked Tabula across classical tasks, including cell type annotation, gene imputation, batch correction, multi-omics integration, and genetic perturbation prediction (covering single-gene, double-gene, and unseen perturbation scenarios)[24]. These analyses showed that Tabula outperformed representative sequence-based single cell foundation models, including scGPT[17], Geneformer[14] and scFoundation[15], across both cell-wise and gene-wise tasks, despite being pretrained on fewer cells (**Supplementary Fig. 1-6**). Beyond these conventional benchmarks, Tabula uniquely enables novel biological discovery tasks, including zero-shot pairwise gene regulation inference, combinatorial regulatory logic discovery, and aging and rejuvenation factor discovery. As detailed in the following sections, these capabilities culminate in the nomination of candidate rejuvenation factors such as *CPE*, *POSTN*, and *OLFM2* from matched young and aged human fibroblast transcriptomes, paving the way toward privacy-preserving virtual cells for human health.

Taken together, Tabula is a novel single cell foundation model that relies on tabular learning and federated learning to better model the intrinsic structure of the scRNA-seq data while ensuring privacy preservation, promising in gaining mechanistic and predictive insights, as described in the following.

## Chiron, a decentralized platform that makes federated training and reuse of single cell foundation models practical for any institution

Tabula's federated design only delivers its data privacy benefits if institutions can adopt it without a dedicated infrastructure team. The conventional path to federated training requires harmonizing data preprocessing across sites, deploying and orchestrating servers, managing authentication and per-dataset access control, and assembling a monitoring layer. Each step is a barrier to entry, especially for the clinical and academic groups that own the most valuable single cell datasets but rarely operate the machine learning infrastructure needed to train them. To remove this barrier, we developed Chiron (https://chiron.aicell.io), an open decentralized training platform built on top of the BioEngine distributed computing infrastructure[29] that brings the full federated training life cycle into a single browser interface (**Fig. 2a**). The same Chiron actions are also exposed as a built-in AI agent skill that any compatible AI agent can call to set up workers and data, launch and monitor a training run, and browse and reuse published models. The Chiron Model Hub catalogues every published Tabula checkpoint with versioning and a public landing page, so any user can browse, fork, retrain, or republish a model. The Chiron training interface guides an operator through configuring a federated session, joining it with a chosen set of trainers, monitoring live progress, and publishing the trained model back to the Hub, closing a loop in which every collaborator's contribution becomes a community resource.

Each participating institution joins the federation by running an independent BioEngine worker on its own hardware (**Fig. 2b**). To take the engineering burden off the operator, the Chiron interface ships a browser-based setup wizard that asks a short form (worker name, path to the local dataset directory, container runtime, CPU, GPU and memory allocation) and emits a one-line launch command tailored to a sandbox or container environment, based on Docker Compose[30], Podman[31], Singularity / Apptainer[32], so the same worker image runs unchanged on local hardware, cloud, or on-premise environments. The launched worker hosts two isolated containers: a local data server that exposes the institution's private single cell datasets only over the worker's container-internal network through authenticated streaming with per-dataset access control, and a Tabula trainer that holds the GPU-bound model. Raw single cell data never moves over external channels. On first read the data server applies the same preprocessing protocol used for Tabula pretraining, selecting the 1,200 most variable genes per dataset as the trainer-ready input, and binning the expression values, so that every trainer across the federated network operates on data prepared identically regardless of where it was uploaded or who prepared it.

Once trainers are deployed across the participating institutions, a central orchestrator coordinates the federated session (**Fig. 2c**). The orchestrator can run on any registered worker regardless of whether it also hosts local data. Each round of training proceeds in four steps. The orchestrator broadcasts the current global transformer weights to every registered trainer, each trainer trains for one local epoch on its private dataset, the updated transformer weights are returned to the orchestrator, the orchestrator performs a FedAvg[33] aggregation to produce a new global model, and the cycle repeats for the configured number of rounds. Across the entire session the orchestrator only ever sees transformer weights and scalar metrics, never the raw single cell data. While a session runs, the Chiron training interface displays per-round training and validation loss curves and downstream task evaluation metrics such as cell type annotation accuracy and gene perturbation prediction, so the operator can assess convergence in real time and stop or extend the session as needed. At the end of the run the orchestrator publishes the FedAvg-aggregated transformer to the Chiron Model Hub for any other Chiron user to reuse, and individual trainers can additionally publish their full client-specific models so that the community accumulates both shared and client-specialized checkpoints.

Across these use patterns (**Fig. 2d**), Chiron is designed as a general framework for federated training of single cell foundation models, beyond the Tabula deployment shown here. The orchestrator and manager applications are not bound to Tabula-specific assumptions and treat the trainer as a Flower-based[34] client over BioEngine's RPC layer, so a different foundation model can join the federation by shipping a model-specific trainer image built using Chiron's trainer template. The trainer template documented in the platform skill makes the contract explicit for external contributions, and other single cell foundation models such as scGPT[17] and Geneformer[14] can be added to the federation through the same path. Taken together, Chiron realizes decentralized training of single cell foundation models as a deployable system that institutions can adopt without an infrastructure project, while preserving the central guarantee that raw single cell data never leaves the institution that owns it.

## Tabula accurately recovers pairwise and high-order combinatorial regulation across diverse biological systems with zero-shot prediction

Although the single cell foundation model has shown to be promising in many different tasks, a true testimony of a valuable single cell foundation model should prove its accuracy, consistency and generalizability when applied under zero-shot setting, e.g., completely new situations without prior specific training, and should confirm its ability in deriving mechanistic biological insights from a very diverse set of biological systems. To demonstrate this, we take on a novel and challenging task by validating Tabula's performance in recovering pairwise and higher-order combinatorial gene regulation across four distinct developmental systems, ranging from hematopoiesis[1,35], cardiogenesis[36,37], neurogenesis[38,39], and pancreatic endogenesis (E17)[40,41], that have accumulated very well validated regulatory networks through decades of biochemical research.

To enable Tabula to predict pairwise and combinatorial regulations, we proposed a novel strategy based on zero-shot prediction. For the pairwise regulation prediction, we iteratively *in silico* perturb the expression of a regulator gene, e.g., $g_1$, with our Tabula model, while tracking the resulting impact on its target ($g_2$) over multiple iterations (**Fig. 3a**). Then the dynamic trajectories of both the regulator and target are characterized, which is further used to determine activation, repression, or neutrality regulation mode based on expression distribution shifts (**Fig. 3a-iv**) (**Fig. 3a-v**) of the target, as well as the proportion pattern of different regulation modes across cells, before and after the iterative perturbation. For example, in the illustrated diagram, $g_2$ expression decreases after in silico activating $g_1$ expression, thus indicating a repression from $g_1$ to $g_2$. A combinatorial regulation involving multiple regulators to the same target can be often simplified as a boolean table. Therefore, for the combinatorial regulation predictions, Tabula first enumerates all possible boolean logic gates of the combinatorial regulation and then follows the procedure from pairwise regulation prediction to perform iterative perturbation prediction based on the logic gate. Specifically, if the value of a regulator is set to be zero or one on the logic gate, the regulator will decrease or increase the gene rank by one. For example, we show Tabula is able to accurately predict the combinatorial regulation from GATA1 and FLI-1 to EKLF as it only increases EKLF's gene expression after we increase GATA-1's expression order by 1 while decreasing FLI-1's order by 1 iteratively but not others. This simple but powerful strategy provides a general approach for Tabula to predict complex gene regulatory scenarios.

After implementing the regulation prediction framework in Tabula, we apply it on four well validated regulatory networks involving hematopoiesis, cardiogenesis, neurogenesis and pancreatic endogenesis by zero-shot prediction without any fine-tuning. The core hematopoiesis network, one of the most studied developmental systems, involves a total of 11 key transcription factors and 29 edges, which specifies cell fates of seven hematopoietic lineages (**Supplementary Fig. 7a and c**). The core cariogenesis network involves nine key TFs and signaling factors (e.g. BMP2, FGF8), and 16 edges, which controls the cell fate for the first heart field and second heart field (**Supplementary Fig. 7b and d**). The core neurogenesis network involves only seven transcription factors and a total of nine edges, which dedicates six neural lineages (**Supplementary Fig. 7e and f**). Lastly the core pancreatic endogenesis network includes eight TFs, 22 edges and involves eight pancreatic lineages across three time points at embryonic day 12 or E12, E14 and E17 (**Supplementary Fig. 8**). Surprisingly, Tabula's simple in silico perturbation approach achieves an overall high predictive accuracy of > 0.83 across four systems on pairwise regulation prediction without the need for any fine-tuning, leveraging the same pretrained model. Taking the three-gene motif of *GATA1*, *FLI1*, and *KLF1* within the core hematopoiesis network[135] (**Fig. 3c and d**) as an example, Tabula accurately predicts both the expression dynamics over iterations (**Fig. 3c**) and distribution shifts pre- and post-perturbation on either activation or repression regulations (**Fig. 3d**). Furthermore, Tabula accurately recovers 24 out of 29 pairwise interaction pairs by either expression distribution shifts (default setting) or cross-cell regulation proportions (**Fig. 3e**). Moreover, the wrongly predicted perturbation pairs often involve genes with significantly lower gene expression compared to other genes (data not shown), underscoring Tabula's reliability in determining regulatory modes.

It has long been known that the large cell type diversity results from the innumerable combinatorial gene regulations of a limited set of genes. Although massively parallel genetic screening, such as Perturb-seq, have made great strides in studying genetic interactions, the exponential increase in the possible higher order interactions render the effort in comprehensively profile combinatorial regulation an impossible mission. It is thus a great contribution if we are able to demonstrate foundation models are capable of revealing higher-order interactions. We thus take on the challenge to demonstrate Tabula's ability to predict combinatorial regulation (CR) from the hematopoiesis system and the cardiogenesis system, both have very well validated combinatorial regulation. Overall, for each system, Tabula's capability in revealing combinatorial rules (CRs) is evaluated against all boolean logic gates. For the cardiogenesis system, there are four known CRs. Our analysis demonstrated that Tabula achieved 100% consistency with the ground truth for CR1 and CR2, 87.5% for CR3, and 12.5% for CR4 (**Fig. 3f**) (**Supplementary Fig. 9a**). We next demonstrate the predicted dynamics of target gene expression for all possible regulator states across all in silico perturbation iterations. For example, FGF8 (CR1) shows two distinct trajectories for all combinations of regulator states (e.g., 0-0-0, 0-1-0, etc.), with the predicted expression levels distribution before and after in silico perturbation 100% aligning with the ground truth provided in the Boolean tables. For CR3, while the majority of predicted expression dynamics follow the same trend and align with the Boolean tables, one regulator state (0-0-0) shows a deviation and fails to capture the dynamics. For the hematopoiesis system, there are nine CRs[4041]. Tabula's predictions achieve 100% consistency for CR1, CR6, CR8, and CR9, while CR3 and CR4 reached 81.25% and 75% consistency, respectively (**Fig. 3g**) (**Supplementary Fig. 9b**).

Taken together, we have demonstrated Tabula's ability to accurately reveal pairwise and combinatorial gene regulation, generalizable across a large range of diverse biological systems, establishing one of the first testament in gaining mechanistic and predictive insight for a single cell foundation model.

## Tabula predicts skin rejuvenation factors through age- and identity score-guided in silico prioritization

Aging has become a global biomedical issue and the primary risk factor for many chronic diseases, yet effective and practical anti-aging strategies remain limited, largely nominated through human creativity and informed guessing[42]. We next explore if Tabula could be deployed to tackle this problem.

Skin fibroblasts, contributing to aging hallmarks such as skin thinness and hair loss in vivo. It fulfills several key characteristics to study aging: (1) it is a physiologically relevant cell type that develops aging signature; (2) it must grow well in vitro to allow cell number expansion; (3) for validation, it must be easily transduced by lentivirus. Given these considerations, we collected human primary fibroblasts from a unique resource, called Baltimore Longitudinal Study of Aging (BLSA), where Skin biopsies can be taken from the same donor across years. These matched sets of cells from a single donor at young and mid-age eliminate confounds from different donor sources and allow us to study age-specific cellular deregulations. To avoid artifact from a single donor, we've picked 3 sets of fibroblasts from 3 donors that have widest time range (>17 years), the “younger” fibroblasts (A1; B1; H1) were collected from patients who were approximately 30 years of age while matched “older” fibroblasts (A2; B2) were collected from the same patients when they were 50 years of age, or from their dad (I2, dad of H1) whose age is 37 older (**Fig.4a, b**).

We performed a scRNA-seq experiment to profile these lines (**Fig.4b**) and identified around 60 genes, that involved in sterol homeostasis and protein transportation, changed significantly during the aging process and their expressions formed the **Age score** (**Supplementary Fig.10a**, **Supplementary Table 1**). Another 19 fibroblast marker genes (e.g. *COL3A1*, *THY1*, *FSP-1*, see **Supplementary Table 1**) are used to build **Identity score**[43]. As expected, aged fibroblast lines have higher age scores and lower identity scores compared to young lines, in both input space and output space (**Fig.4c and d**, **Supplementary Fig.10b**), consistent with previous findings[44].

For each fibroblast age pair, we selected the 1,200 most highly variable genes (HVGs) between young and old cells and used their expression profiles as input to a discriminator multilayer perceptron (MLP; see **Methods**). This feedforward neural network was trained to predict an age score and an identity score. Training minimized the mean squared error (MSE) between the predicted scores and the corresponding ground-truth scores, updating model parameters over successive epochs until convergence.

To nominate candidate rejuvenation factors, we embedded aged cells into a frozen Tabula model that reconstructs cell states and used an iterative optimization procedure to search for perturbations that shift these reconstructions toward a younger phenotype. At each optimization step, reconstructed cells were evaluated by the above discriminator MLP with frozen model parameters that outputs two quantities: an age score and an identity score. The objective was to decrease the age score toward a lower boundary corresponding to simulated young cells (set to 17-year-old, our youngest donor), while simultaneously increasing the identity score toward an upper boundary defined as the median identity score of young cells (**Fig. 4e, Supplementary Fig.11a, 11b**).

After optimization, we aggregated gene-level contributions across trajectories using an accumulated path score, and ranked genes accordingly to prioritize candidate rejuvenation factors (**Fig. 4a)**. For looking at how much candidates are common between lines, the union of top 500 candidate rejuvenation genes in each cell line are compared, a substantial fraction of candidates is shared across all three groups, whereas additional subsets are shared between pairs of groups or remain group-specific, indicating that Tabula captures common transcriptional programs associated with rejuvenation, while also preserving donor-specific components (**Fig. 4f**).

We then used mSALT[45], a mouse atlas of aging and longevity interventions, to evaluate whether candidate rejuvenation factors are associated with transcriptional programs observed across lifespan-extending interventions. This atlas allows us to evaluate Tabula-nominated rejuvenation factors across multiple dimensions: whether Tabula is more robust to nominate from input or output space, and whether it performed better from Patient specific HVG, or Patient union HVGs (**Supplementary Fig. 12a**). Union HVGs pools cells from all three lines and selects 1,200 HVGs as input for Tabula ranking in both input space and output space.

We defined the baseline as the expected fraction of mSALT-positive genes in randomly selected gene sets and performed a top-k enrichment analysis (**Methods**). We observe that performance peaks at small k, with the top 10 genes achieving the highest enrichment of positive genes (**Supplementary Fig. 12b**). Notably, Tabula consistently outperforms the baseline in both input and output spaces, achieving >45% enrichment and up to 151% enrichment. These enrichment levels also exceeded the overall probability of a young-associated gene being identified as an mSALT-positive gene (**Fig.4g**, **Supplementary Fig. 12c,d**), further demonstrating that Tabula improves the biological relevance and positive enrichment of top-ranked rejuvenation factor candidates. The top ranked genes are *CPE* in Patient specific HVG input space, *POSTN* in the Patient specific HVGs output space, *OLFM2* in Union output space (**Fig.4h**). *Cpe* is upregulated by six individual longevity interventions, including Ames, Snell, and Little—three of the most robust models of lifespan extension (**Supplementary Fig.13a**). *Olfm2* is upregulated in acarbose-treated and Ames dwarf mice (**Supplementary Fig.13b**). *Postn* is upregulated in Ames, CR, GHRKO, Little, Myc deficiency interventions (**Supplementary Fig.14**). Interestingly, the loss or downregulation of *CPE* is associated with the aging process in human dermal fibroblasts[46]. *OLFM2* also links to skin-aging-related changes in human dermal fibroblasts, as it is associated with extracellular matrix-supportive fibroblast programs regulated by SPARC and LRG1, both of which decline in aged skin[47,48]. Likewise, reduced *POSTN* expression has been observed in aged skin and is associated with compromised collagen organization and extracellular matrix homeostasis[49].

Finally, we asked whether the perturbations nominated by Tabula overlap with, or are orthogonal to, existing rejuvenation strategies such as reprogramming factor-based paradigms. Partial epigenetic reprogramming with Yamanaka factors (*OCT4*, *SOX2*, *KLF4*), as a particularly influential rejuvenation strategy[42,50,51]. To experimentally measure perturbation effects of single and dual factors, we followed a prior example[52] and developed a novel inducible overexpression vector that is compatible with Perturb-seq[53] (**Supplementary Fig.15a**), to capture genetic perturbation and its transcriptional consequences, allowing simultaneous testing of single or dual combinations of three reprogramming factors (*OCT4*, *SOX2*, *KLF4*) (**Fig. 4i**). We applied these lentiviral constructs to three aged fibroblast lines and performed Perturb-seq analysis (**Fig. 4j; Supplementary Fig. 15b**). The analysis showed that, in fibroblasts, overexpression of either *OCT4* or *SOX2* alone was sufficient to reduce the age score (**Fig. 4k**), whereas *KLF4* appeared to increase it. Notably, all perturbations reduced the identity score, underscoring the importance of introducing the Tabula framework in nominating rejuvenation factors while preservation of cell identity. Interestingly, *OCT4* and *SOX2* did not alter expression of *CPE* or *OLFM2*, suggesting that *CPE* and Olfm2 represent rejuvenation strategies orthogonal to partial reprogramming. In contrast, mt-CO2, a top-10 candidate identified from the Patient specific HVGs input space analysis, was also induced by *OCT4* and *SOX2* (**Fig. 4l**), indicating that some Tabula-nominated interventions can converge with mechanisms engaged by partial reprogramming. Together, this suggests Tabula can be useful for systematically nominating candidate rejuvenation factors, rather than relying only on prior intuition or manual hypothesis generation.

# DISCUSSION

Advances in genomics now enable us to profile single cells at unprecedented scale, resolution and diversity across the spectrum of human biology. The expanding single cell dataset provides increasingly rich and diverse data, allowing the creation of virtual cell models to enhance our understanding of human biology with generative AI[54]. Early efforts in this direction, such as Geneformer, scGPT and scFoundation, have demonstrated great promise. However, as we begin training foundation models on a large corpus of human data, data privacy becomes a significant concern. Moreover, while it is convenient to construct gene orders across cells to mimic word orders in sentences for subsequent transformer-based pretraining as employed in these efforts, the unique features of single cell data demand novel design.

We developed Tabula, the first single cell foundation model that explicitly models the tabular structure of scRNA-seq data and integrates federated learning to address privacy concerns. Tabula is unique in two key ways. First, it uses a tabular learning framework to explicitly model the tabular structure of single cell data. Unlike other approaches that preprocess gene expression into gene sequences, Tabula represents each cell as a permutation-invariant row of genes. Furthermore, rather than relying on mask language modeling, Tabular's pretraining comprises of two objectives, namely gene-wise learning, which reconstruct original gene expression from a corrupted view, and cell-wise learning, which uses contrastive learning in the latent space to distinguish between positive (cell pairs from the same origin, with one generated from corrupted data) and negative (different cells) cell pairs. Extensive benchmarking demonstrates that tabular learning outperforms MLM- and autoregression-based approaches on a range of cell- and gene-level downstream tasks, e.g. cell type annotation, batch correction, gene imputation and genetic perturbation prediction (covering single-gene, double-gene, and unseen perturbation scenarios). Second, Tabula employs federated learning[22], allowing decentralized clients to train independently without sharing raw data, thereby addressing the privacy concerns. Thanks to the architectural improvement, we show that Tabula is able to allow impressive results in recovering pairwise and higher-order regulation across diverse biological systems. This work establishes one of the first pieces of evidence for the single cell foundation model successfully recovering ground-truth regulatory networks. Beyond these core capabilities, we further demonstrate the potential of Tabula to address biomedically important problems, exemplified by the nomination of rejuvenation factors for human skin aging. Through application to matched young and aged fibroblast transcriptomes and the development of age- and identity score-guided in silico prioritization method, Tabula can identify candidate rejuvenation genes at efficiency outperforming conventional approaches and are supported by longevity intervention signatures, establishing single-cell foundation models as a practical discovery engine to assist longevity research.

Beyond Tabula itself, Chiron contributes an AI agent-enabled, open, decentralized platform that brings federated training of single cell foundation models within practical reach of any institution. By bundling worker orchestration, training control, downstream-task monitoring, model publishing, and AI agent integration into a single browser-driven workflow, Chiron lowers the barrier to participation from a multi-month infrastructure project to a one-command worker launch. We expect this to be most valuable in clinical and consortium settings where data residency is non-negotiable. Furthermore, we anticipate that trainers for other single cell foundation models contributed alongside Tabula will broaden the platform from a Tabula-specific deployment to a shared federation infrastructure for the field. As more groups contribute checkpoints back to the Chiron Model Hub, the platform should accumulate into a shared resource for collaborative single cell modelling that complements, rather than replaces, single-institution releases on existing model hubs.

Although Tabula represents an important advancement in single cell foundation models, it is not without limitations. In its federated learning framework, all clients share a general transformer model, while distinct embedders are used for client-specific data. This design enables handling heterogeneous data distributions across clients (e.g., tissue-specific data). Ideally, however, a universal embedder could be used. Nevertheless, the client-specific embedder primarily affects zero-shot learning on downstream tasks, and it improves tissue-specific prediction performance. For fine-tuning downstream tasks, the choice of embedder has minimal impact. Currently, Tabula has been trained on only 15 million single cells, focusing solely on transcriptomics data. Scaling up training to include more single cells, spatial transcriptomics and incorporating multi-omics data, e.g. chromatin accessibility and gene expression co-assays, is an important next step. For instance, ideas from multimodal single-cell models, including transformer-based multimodal prediction frameworks[55] and foundation models such as GET[56] and cross-modality translation frameworks such as scLinguist[57] could help extend Tabula from a transcriptomics-only model toward a multi-modal foundation model spanning gene expression, chromatin accessibility, and protein abundance. Although Tabula has shown promise in gaining mechanistic biological insights, such as high-order perturbation predictions and rejuvenation factor nomination, future work must further expand the prediction capabilities to reveal novel biological insights, such as the optimal gene combination that converts one cell fate to another for regenerative medicine. While single cell foundation models excel in various tasks, integration with more specialized models may be complementary in the short time.

Finally, with the rapid advancements of AI agents in recent years, combining foundation models with agent systems to automate biological discovery represents an exciting frontier in single cell biology[58,59]. This endeavor could pave the way not only for virtual cells[54], virtual embryos[60], but also for virtual labs[61] or even virtual scientists[62] in years to come.

# ACKNOWLEDGEMENTS

This work is supported by Laude Moonshot seed grant (X. Q.), Impetus Longevity grant (X. Q.), Pantas And Ting Sutardja Foundation (X. Q.), Wu Tsai Neurosciences Institute Big Ideas in Neuroscience Program (X. Q.), the Milky Way Research Foundation (J.S.W.), and the SciLifeLab & Wallenberg Data Driven Life Science Program (grant: KAW 2020.0239) (W.O.). We appreciate valuable brainstorming, technical discussion, feedback from Y.M., W.T., and K.P. and other members in the Qiu lab (Stanford) and DSE lab (MSU) during the preparation and execution of this work as well as the writing of the manuscript. We thank Daniel Zhu for his help with the fibroblast single cell data preparation. We thank Alexander Tyshkovskiy for his guidance on using the mSALT website and dataset. We also acknowledge the use of ChatGPT for proofreading the final version manuscript.

# AUTHOR CONTRIBUTIONS

J.D., and Y.W. conceived the study, integrating tabular learning for single-cell data modeling with federated learning to ensure data privacy protection. J.D. and X.Q. led the development of the entire project. X.Q. proposed strategies for validating Tabula on downstream tasks, including building the Chiron platform, pairwise and combinatorial gene regulatory prediction, aging-related applications. J.L. and S.J. contributed to model pretraining and fine-tuning across standard downstream benchmarks, including cell-type annotation, perturbation response prediction, gene imputation, and other related tasks. They also contributed to algorithm implementation and analysis, with guidance from J.D., X.Q., M.L., and J.T.. Z.M. developed the computational method for aging rejuvenation with guidance from J.D., X.Q., W.O. and Y.R.L.. Y.R.L. designed and conducted the wet-lab experiments for generating the skin fibroblast aging and Perturb-seq datasets, with guidance from R.S., W.C., and J.S.W.. X.Q. performed early analyses of the aging single cell dataset. N.M. and W.O. developed the Chiron platform. J.D. and N.M. designed and generated the initial version of all figures in the manuscript with the help of J.L., S.J., W.O., N.M., and Z.M.; X.Q. provided additional guidance on refining the revised figures, from initial concept design to final organization. All authors contributed to the development and/or writing of the manuscript.

# COMPETING INTERESTS

J.S.W. declares outside interest in 5AM Ventures, Amgen, nChroma, DEM Biosciences, KSQ Therapeutics, Maze Therapeutics, Tenaya Therapeutics, Tessera Therapeutics, Thermo Fisher, Xaira, and TRV. Y.R.L., has patent applications licensed to Life Biosciences Inc. J.D. is an employee of Hippocratic AI. W.O. is a co-founder of Amun AI AB, a commercial company that builds, delivers, supports and integrates AI systems for academic, biotech and pharmaceutical industries.

# CODE AVAILABILITY

**Tabula** (version: 1.0) is implemented with Pytorch Lightning and freely accessible at https://github.com/aristoteleo/tabula. Chiron is freely accessible at http://github.com/aristoteleo/chiron and can be accessed at https://chiron.aicell.io.

# DATA AVAILABILITY

All data utilized in this study are publicly accessible, with detailed usage descriptions provided in the Methods section. The data sources are as follows: The pretraining datasets were obtained from the CELLxGENE census (https://cellxgene.cziscience.com). For the imputation task, both PBMC5K and Jurkat datasets, generated by 10x Genomics, are available from their official website (https://www.10xgenomics.com/datasets). The Melanoma dataset is publicly accessible in the GEO database under accession number GSE99330. For the genetic perturbation prediction task, Norman and Adamson datasets can be accessed via (https://dataverse.harvard.edu/api/access/datafile/6154020) and (https://dataverse.harvard.edu/api/access/datafile/6154417), respectively. The Replogle dataset was retrieved from (https://gwps.wi.mit.edu). For the cell type annotation task, the Human Pancreas dataset is available at (https://figshare.com/projects/TOSICA_demo/158489). The Myeloid dataset can be accessed from (https://drive.google.com/drive/folders/1VbpApQufZq8efFGakW3y8QDDpY9MBoDS). The 293t & Jurkat cells and DC datasets are available at (https://hub.docker.com/r/jinmiaochenlab/batch-effect-removal-benchmarking). The corresponding GEO database accession numbers for the liver reference and query datasets are GSE115469 and GSE124395, respectively. For the multi-omics integration task, the 10x Multiome PBMC dataset was obtained from (https://scglue.readthedocs.io/en/latest/data.html). The BMMC dataset is available in the GEO database under accession number GSE194122. The Perirhinal Cortex dataset can be accessed through (https://cellxgene.cziscience.com/collections/283d65eb-dd53-496d-adb7-7570c7caa443). For the in silico gene regulatory network inference task: the hematopoiesis dataset is available at (https://dynamo-release.readthedocs.io/en/latest/notebooks/tutorial_hsc_velocity.html); the cardiogenesis dataset is accessible in the GEO database under accession number GSE126128; the neurogenesis dataset can be retrieved from (mousebrain.org); and the pancreatic endogenous dataset is available in the GEO database under accession number GSE101099.

# REFERENCES

1. Qiu, X. *et al.* Mapping transcriptomic vector fields of single cells. *Cell* **185**, 690–711.e45 (2022).
2. Deep Learning in Single-cell Analysis. *ACM Transactions on Intelligent Systems and Technology* (2024) doi:10.1145/3641284.
3. Rozenblatt-Rosen, O., Stubbington, M. J. T., Regev, A. & Teichmann, S. A. The Human Cell Atlas: from vision to reality. *Nature* **550**, 451–453 (2017).
4. HuBMAP Consortium. The human body at cellular resolution: the NIH Human Biomolecular Atlas Program. *Nature* **574**, 187–192 (2019).
5. Devlin, J., Chang, M.-W., Lee, K. & Toutanova, K. BERT: Pre-training of Deep Bidirectional Transformers for Language Understanding. in *Proceedings of the 2019 Conference of the North American Chapter of the Association for Computational Linguistics: Human Language Technologies, Volume 1 (Long and Short Papers)* 4171–4186 (2019).
6. Touvron, H. *et al.* LLaMA: Open and Efficient Foundation Language Models. (2023).
7. OpenAI *et al.* GPT-4 Technical Report. (2023).
8. Ranftl, R., Bochkovskiy, A. & Koltun, V. Vision Transformers for Dense Prediction. https://ieeexplore.ieee.org/document/9711226.
9. He, K., Zhang, X., Ren, S. & Sun, J. Deep Residual Learning for Image Recognition. https://ieeexplore.ieee.org/document/7780459.
10. Abramson, J. *et al.* Accurate structure prediction of biomolecular interactions with AlphaFold 3. *Nature* **630**, 493–500 (2024).
11. Lin, Z. *et al.* Evolutionary-scale prediction of atomic-level protein structure with a language model. *Science* (2023) doi:10.1126/science.ade2574.

12. Rives, A. *et al.* Biological structure and function emerge from scaling unsupervised learning to 250 million protein sequences. *Proc Natl Acad Sci U S A* **118**, (2021).
13. Vaswani, A. *et al.* Attention Is All You Need. (2017).
14. Theodoris, C. V. *et al.* Transfer learning enables predictions in network biology. *Nature* **618**, 616–624 (2023).
15. Hao, M. *et al.* Large-scale foundation model on single-cell transcriptomics. *Nat Methods* **21**, 1481–1491 (2024).
16. Wen, H. *et al.* CellPLM: Pre-training of Cell Language Model Beyond Single Cells. *bioRxiv* 2023.10.03.560734 (2023) doi:10.1101/2023.10.03.560734.
17. Cui, H. *et al.* scGPT: toward building a foundation model for single-cell multi-omics using generative AI. *Nat Methods* **21**, 1470–1480 (2024).
18. Rosen, Y. *et al.* Universal Cell Embeddings: A Foundation Model for Cell Biology. *Cell Biology* (2023).
19. Schaar, A. C. *et al.* Nicheformer: a foundation model for single-cell and spatial omics. *bioRxiv* 2024.04.15.589472 (2024) doi:10.1101/2024.04.15.589472.
20. Ferguson, D. ’nae. A Privacy Concern: Bioinformatics and Storing Biodata. *The ADMI 2021 Symposium* (2021).
21. Walker, C. R. *et al.* Private information leakage from single-cell count matrices. *Cell* **187**, 6537–6549.e10 (2024).
22. Kairouz, P. *et al.* Advances and Open Problems in Federated Learning. *MAL* **14**, 1–210 (2021).
23. Yang, F. *et al.* scBERT as a large-scale pretrained deep language model for cell type annotation of single-cell RNA-seq data. *Nature Machine Intelligence* **4**, 852–866 (2022).
24. Ding, J. *et al.* Tabula: A Tabular Self-Supervised Foundation Model for Single-Cell Transcriptomics. (2025).

25. Schaar, A. C. *et al.* Nicheformer: a foundation model for single-cell and spatial omics. *Bioinformatics* (2024).
26. Hsiao, M. & Hung, M. Construction of an Artificial Intelligence Writing Model for English Based on Fusion Neural Network Model. *Comput Intell Neurosci* **2022**, 1779131 (2022).
27. Generative pretraining from large-scale transcriptomes for single-cell deciphering. *iScience* **26**, 106536 (2023).
28. Gorishniy, Y., Rubachev, I., Khrulkov, V. & Babenko, A. Revisiting Deep Learning Models for Tabular Data. *Adv. Neural Inf. Process. Syst.* **34**, 18932–18943 (2021).
29. Mechtel, N. *et al.* BioEngine: scalable execution and adaptation of bioimage AI through agent-readable interfaces. *bioRxiv* (2026) doi:10.64898/2026.04.19.719496.
30. Merkel, D. Docker: Lightweight Linux Containers for Consistent Development and Deployment. *Linux Journal* **2014**, 2 (2014).
31. Gantikow, H., Walter, S. & Reich, C. Rootless Containers with Podman for HPC. in *Lecture Notes in Computer Science* 343–354 (Springer International Publishing, Cham, 2020).
32. Kurtzer, G. M., Sochat, V. & Bauer, M. W. Singularity: Scientific containers for mobility of compute. *PLoS One* **12**, e0177459 (2017).
33. McMahan, B., Moore, E., Ramage, D., Hampson, S. & Aguera y Arcas, B. Communication-Efficient Learning of Deep Networks from Decentralized Data. in *Artificial Intelligence and Statistics* 1273–1282 (PMLR, 2017).
34. Beutel, D. J. *et al.* Flower: A Friendly Federated Learning Research Framework. (2020).
35. Krumsiek, J., Marr, C., Schroeder, T. & Theis, F. J. Hierarchical differentiation of myeloid progenitors is encoded in the transcription factor network. *PLoS One* **6**, e22649 (2011).
36. de Soysa, T. Y. *et al.* Single-cell analysis of cardiogenesis reveals basis for organ-level developmental defects. *Nature* **572**, 120–124 (2019).
37. Nakanishi, T. *et al. Etiology and Morphogenesis of Congenital Heart Disease: From Gene Function and Cellular Interaction to Morphology*. (Springer, 2016).

38. La Manno, G. *et al.* Molecular architecture of the developing mouse brain. *Nature* **596**, 92–96 (2021).
39. Qiu, X., Ding, S. & Shi, T. From Understanding the Development Landscape of the Canonical Fate-Switch Pair to Constructing a Dynamic Landscape for Two-Step Neural Differentiation. *PLOS ONE* **7**, e49271 (2012).
40. Wang, C. *et al.* Dysregulated lung stroma drives emphysema exacerbation by potentiating resident lymphocytes to suppress an epithelial stem cell reservoir. *Immunity* **56**, 576–591.e10 (2023).
41. Wang, J., Yuan, R., Zhu, X. & Ao, P. Adaptive landscape shaped by core endogenous network coordinates complex early progenitor fate commitments in embryonic pancreas. *Sci. Rep.* **10**, 1112 (2020).
42. Lu, Y. *et al.* Reprogramming to recover youthful epigenetic information and restore vision. *Nature* **588**, 124–129 (2020).
43. Olova, N., Simpson, D. J., Marioni, R. E. & Chandra, T. Partial reprogramming induces a steady decline in epigenetic age before loss of somatic identity. *Aging Cell* **18**, e12877 (2019).
44. Yang, J.-H. *et al.* Loss of epigenetic information as a cause of mammalian aging. *Cell* **186**, 305–326.e27 (2023).
45. Tyshkovskiy, A. *et al.* Distinct longevity mechanisms across and within species and their association with aging. *Cell* **186**, 2929–2949.e20 (2023).
46. Peng, X. *et al.* Integrated bioinformatics analysis and experimental validation identifies CPE as a potential biomarker and therapeutic target for skin aging. *Exp. Dermatol.* **33**, e15120 (2024).
47. Ham, S. M. *et al.* SPARC Is Highly Expressed in Young Skin and Promotes Extracellular Matrix Integrity in Fibroblasts via the TGF-β Signaling Pathway. *Int J Mol Sci* **24**, (2023).

48. Park, H. N. *et al.* LRG1 Promotes ECM Integrity by Activating the TGF-β Signaling Pathway in Fibroblasts. *Int J Mol Sci* **24**, (2023).

49. Egbert, M. *et al.* The matricellular protein periostin contributes to proper collagen function and is downregulated during skin aging. *J Dermatol Sci* **73**, 40–48 (2014).

50. Drake, S. S. *et al.* Cellular rejuvenation protects neurons from inflammation-mediated cell death. *Cell Rep.* **44**, 115298 (2025).

51. Lu, Y. R. *et al.* Reprogramming factors activate a non-canonical oxidative resilience pathway that can rejuvenate RPEs and restore vision. *bioRxivorg* (2025) doi:10.1101/2025.08.30.673239.

52. Roux, A. E. *et al.* Diverse partial reprogramming strategies restore youthful gene expression and transiently suppress cell identity. *Cell Syst.* **13**, 574–587.e11 (2022).

53. Adamson, B. *et al.* A Multiplexed Single-Cell CRISPR Screening Platform Enables Systematic Dissection of the Unfolded Protein Response. *Cell* **167**, 1867–1882.e21 (2016).

54. Bunne, C. *et al.* How to build the virtual cell with artificial intelligence: Priorities and opportunities. *Cell* **187**, 7045–7063 (2024).

55. Tang, W. *et al.* Single-cell multimodal prediction via transformers. in *Proceedings of the 32nd ACM International Conference on Information and Knowledge Management* 2422–2431 (ACM, New York, NY, USA, 2023).

56. Fu, X. *et al.* GET: a foundation model of transcription across human cell types. *bioRxiv : the preprint server for biology* (2024) doi:10.1101/2023.09.24.559168.

57. Fang, Z. *et al.* scLinguist: A pre-trained hyena-based foundation model for cross-modality translation in single-cell multi-omics. *bioRxiv* 2025.09. 30.679123 (2025) doi:10.1101/2025.09.30.679123.

58. Huang, K. *et al.* Biomni: A General-Purpose Biomedical AI Agent. *bioRxiv* (2025) doi:10.1101/2025.05.30.656746.

59. Xu, W. *et al.* PantheonOS: An Evolvable Multi-Agent Framework for Automatic Genomics Discovery. *bioRxiv* (2026) doi:10.64898/2026.02.26.707870.

60. Cao, N., Lu, Y. & Qiu, X. Towards predictive virtual embryos with genomics and AI. *Nat Methods* (2026) doi:10.1038/s41592-026-03055-4.

61. Swanson, K., Wu, W., Bulaong, N. L., Pak, J. E. & Zou, J. The Virtual Lab: AI Agents Design New SARS-CoV-2 Nanobodies with Experimental Validation. *bioRxiv* 2024.11.11.623004 (2024) doi:10.1101/2024.11.11.623004.

62. Lu, C. *et al.* The AI Scientist: Towards Fully Automated Open-Ended Scientific Discovery. (2024).

63. Zhu, B. *et al.* XTab: Cross-table Pretraining for Tabular Transformers. (2023).

64. Marginal distribution. https://en.wikipedia.org/wiki/Marginal_distribution (2004).

65. Dao, T. FlashAttention-2: Faster attention with better parallelism and work partitioning. *arXiv [cs.LG]* (2023).

66. Sarkar, A. & Stephens, M. Separating measurement and expression models clarifies confusion in single-cell RNA sequencing analysis. *Nat. Genet.* **53**, 770–777 (2021).

67. Svensson, V. *et al.* Power Analysis of Single Cell RNA-Sequencing Experiments. *Nat. Methods* **14**, 381 (2017).

68. Jiang, R., Sun, T., Song, D. & Li, J. J. Statistics or biology: the zero-inflation controversy about scRNA-seq data. *Genome Biol.* **23**, (2022).

69. Taud, H. & Mas, J. F. Multilayer Perceptron (MLP). *Geomatic Approaches for Modeling Land Change Scenarios* 451–455 (2018).

70. Chen, T., Kornblith, S., Norouzi, M. & Hinton, G. A simple framework for contrastive learning of visual representations. *arXiv [cs.LG]* (2020).

71. Li, X., Huang, K., Yang, W., Wang, S. & Zhang, Z. On the Convergence of FedAvg on Non-IID Data. (2019).

72. Ding, J. *et al.* DANCE: a deep learning library and benchmark platform for single-cell analysis. *Genome Biology* **25**, 72 (2024).
73. Ding, J. *et al.* DANCE 2.0: Transforming single-cell analysis from black box to transparent workflow. *bioRxiv* 2025.07. 17.665427 (2025) doi:10.1101/2025.07.17.665427.
74. Luecken, M. D. *et al.* Benchmarking atlas-level data integration in single-cell genomics. *Nat. Methods* **19**, 41–50 (2022).
75. Wen, H. *et al.* Graph neural networks for multimodal single-cell data integration. in *Proceedings of the 28th ACM SIGKDD Conference on Knowledge Discovery and Data Mining* vol. 12 4153–4163 (ACM, New York, NY, USA, 2022).
76. Roohani, Y., Huang, K. & Leskovec, J. Predicting transcriptional outcomes of novel multigene perturbations with GEARS. *Nature Biotechnology* **42**, 927–935 (2023).
77. Arisdakessian, C., Poirion, O., Yunits, B., Zhu, X. & Garmire, L. X. DeepImpute: an accurate, fast, and scalable deep neural network method to impute single-cell RNA-seq data. *Genome Biol.* **20**, 211 (2019).
78. Virshup, I., Rybakov, S., Theis, F. J., Angerer, P. & Alexander Wolf, F. anndata: Access and store annotated data matrices. *Journal of Open Source Software* **9**, 4371 (2024).
79. Moore, J. & Kunis, S. Zarr: A Cloud-Optimized Storage for Interactive Access of Large Arrays. *Proc Conf Res Data Infrastr* **1**, (2023).
80. Brennecke, P. *et al.* Accounting for technical noise in single-cell RNA-seq experiments. *Nature Methods* **10**, 1093–1095 (2013).
81. McInnes, L., Healy, J., Saul, N. & Großberger, L. UMAP: Uniform Manifold Approximation and Projection. *Journal of Open Source Software* **3**, 861 (2018).
82. Falcon, W. *et al.* PyTorchLightning/pytorch-lightning: 0.7.6 release. doi:10.5281/zenodo.3828935.
83. Nvidia, N. [引用] NVIDIA Collective Communications Library (NCCL). (2017).
84. Loshchilov, I. & Hutter, F. Decoupled Weight Decay Regularization. (2017).

85. Torre, E. *et al.* Rare cell detection by single-cell RNA sequencing as guided by single-molecule RNA FISH. *Cell Syst.* **6**, 171–179.e5 (2018).

86. Norman, T. M. *et al.* Exploring genetic interaction manifolds constructed from rich single-cell phenotypes. *Science* **365**, 786–793 (2019).

87. Replogle, J. M. *et al.* Mapping information-rich genotype-phenotype landscapes with genome-scale Perturb-seq. *Cell* **185**, 2559–2575.e28 (2022).

88. Chen, J. *et al.* Transformer for one stop interpretable cell type annotation. *Nat Commun* **14**, 223 (2023).

89. Baron, M. *et al.* A single-cell transcriptomic map of the human and mouse pancreas reveals inter- and intra-cell population structure. *Cell Syst.* **3**, 346–360.e4 (2016).

90. Tran, H. T. N. *et al.* A benchmark of batch-effect correction methods for single-cell RNA sequencing data. *Genome Biol* **21**, (2020).

91. MacParland, S. A. *et al.* Single cell RNA sequencing of human liver reveals distinct intrahepatic macrophage populations. *Nat. Commun.* **9**, 4383 (2018).

92. Aizarani, N. *et al.* A human liver cell atlas reveals heterogeneity and epithelial progenitors. *Nature* **572**, 199–204 (2019).

93. Cao, Z.-J. & Gao, G. Multi-omics single-cell data integration and regulatory inference with graph-linked embedding. *Nat. Biotechnol.* **40**, 1458–1466 (2022).

94. Luecken, M. D., Burkhardt, D. B. & Cannoodt, R. A sandbox for prediction and integration of DNA, RNA, and proteins in single cells. *Thirty-fifth conference* (2021).

95. Siletti, K. *et al.* Transcriptomic diversity of cell types across the adult human brain. *Science* **382**, eadd7046 (2023).

96. Villani, A.-C. *et al.* Single-cell RNA-seq reveals new types of human blood dendritic cells, monocytes, and progenitors. *Science* **356**, (2017).

97. Herrmann, F., Groß, A., Zhou, D., Kestler, H. A. & Kühl, M. A boolean model of the cardiac gene regulatory network determining first and second heart field identity. *PLoS One* **7**, e46798 (2012).

# ONLINE METHODS

## Pretraining data collection and preprocessing

**Data collection**. To establish a robust foundation for Tabula self-supervised pretraining, we assemble an extensive single cell RNA sequencing (scRNA-seq) dataset comprising transcriptomic data from approximately 15 million human cells, sourced from the CELLxGENE portal **(**[**https://cellxgene.cziscience.com**](https://cellxgene.cziscience.com/)**/).**

Given that the CELLxGENE portal is regularly updated, we utilized the release from July 25, 2023. The dataset encompasses cells from a range of major human tissues, including the intestine, pancreas, lung, heart, blood, kidney, and brain. Furthermore, cells from a variety of less-represented human tissues like the spinal cord and spleen were aggregated into a separate category called “others”. The compilation resulted in a total of 8 categories, namely “intestine”, “pancreas”, “lung”, “heart”, “blood”, “kidney”, “brain” and “others”, comprising 37.99 million human cells. In our federated learning framework, we set the number of clients to be eight, each corresponding to a distinct tissue category aforementioned in our setting. To maintain data balance across categories and guarantee effective pretraining, a selection process was implemented, limiting each category to a maximum of 3 million cells.

We regrouped each category according to its original dataset identifiers. We then performed quality control and data preprocessing procedures for cells belonging to the same source dataset separately (detailed methodology described in the next subsection). Within each category, we randomly selected 3 million cells from the preprocessed datasets. For categories with fewer than 3 million total available cells, all cells were used. This process ultimately yields a pretraining dataset of 15 million cells. The number of pretraining cells for each tissue client is as follows: Intestine: 80,060 cells; Pancreas: 220,436 cells; Lung: 3,013,840 cells ; Heart: 1,768,184 cells; Blood: 3,057,298 cells; Kidney: 816,538 cells ; Brain: 3,024,233 cells; Others: 3,046,184 cells.

**Quality control & data preprocessing for federated learning**. We implemented a dataset-specific strategy for quality control and data preprocessing. This strategy diverges from previous approaches of consolidating data prior to quality control and data preprocessing, allowing us to preserve the distinctive characteristics of each original dataset and reduce the impact of the batch effect. Specifically, we filtered cells with fewer than 250 detected genes to exclude potential empty droplets, non-viable cells, or severely degraded cells. Concurrently, we also removed genes detected in fewer than 250 cells to mitigate the influence of lowly expressed or rare transcripts. To capture biologically relevant variations, we identified 1,200 highly variable genes (HVGs) within each dataset after quality control. It is worth noting that HVGs were identified prior to the aggregation of datasets for each tissue. Consequently, HVGs may differ among datasets even within the same tissue. We chose this strategy because selecting HVGs from individual datasets or distinct contexts can improve the model’s ability to discern cell-specific characteristics while also enhances the model's capacity to learn generalizable features from varied biological signals. To standardize the data for pretraining and further mitigate batch effects, we implemented a value binning technique, discretizing the expression values into 50 bins for each cell. This transformation normalizes the scale of gene expression across batches, creating a more uniform input space for the pretraining process. The choice of 50 bins strikes a balance between preserving gene expression differences and reducing batch-specific technical variations.

**Data preprocessing for tabular modeling**. Our model employs a tabular learning approach for self-supervised pretraining, with learning objectives that include contrastive loss and reconstruction loss. The contrastive loss focuses on enhancing feature discrimination between cells, while the reconstruction loss aims to capture complex relationships among genes within each cell. To achieve these objectives, we need to corrupt the input samples to construct corrupted views from the original views. We followed Xtab [63] to construct corrupted views through random feature resampling. Specifically, we randomly selected a subset of genes in cells in a training batch and then resampled their values from the empirical marginal distribution [64] of these genes within the same training batch. We set the corrupted ratio at 60%. This means that for each cell sample's original view $x$ and its corrupted view $\tilde{x}$, 60% of the gene values are resampled while 40% remain unchanged. This resampling strategy will provide sufficient perturbation to enhance the model's robustness while retaining enough original information to maintain biological relevance.

## Tabula model architecture

Tabula is a privacy-preserving foundational model for single cell omics developed within a federated learning framework. Tabula also innovatively treats single cell data (represented as a cell-by-gene matrix) similarly to tabular data and incorporates a tabular transformer with attention mechanisms to facilitate the unsupervised learning of gene and cell embeddings. Additionally, Tabula handles cross-tissue variations by partitioning the models into different clients with tissue-specific embedders and projection heads that capture the characteristics of each tissue, and a shared tabular transformer that stores the common knowledge across tissues. Note that this cross-tissue specific client system is implemented to demonstrate the federated learning, however, in general federated learning can be used for cross-institute learning to address privacy concerns by avoiding data sharing across institutes while benefiting the knowledge learned from the data produced from other institutes during pretraining.

In the following, we will describe the four main modules related to the Tabula model architecture, namely the embedder module, the tabular transformer module, the pretraining objectives and the collaborative federated learning framework. The embedder module transforms the original cell-by-gene table into embeddings, which are then used as inputs for the tabular transformer. Then the tabular transformer, powered with FlashAttention-2 [65] takes the embeddings to learn contextual gene and cell representation efficiently and effectively. The output of the tabular transformer is further processed by projection heads to derive the pretraining losses. Tabular pretraining objective is based on self-supervised loss functions tailored for cell-by-gene table learning through cell-wise row learning and gene-wise column learning. the collaborative federated learning framework module details the process of updating between the local and global tabular transformer models.

**Embedder module**. scRNA-seq data can be represented as a cell-by-gene matrix, $\mathbf{X} \in \mathbb{R}^{N \times M}$, where each entry $\mathbf{X}_{i,j} \in \mathbb{R}^{+}$ denotes the read count of RNA molecules for gene $j \in \{0, 1, \ldots, M\}$ in cell $i \in \{0, 1, \ldots, N\}$. Each row of the cell-by-gene table, $\mathbf{X}$, is considered as an input cell sample in a tabular transformer, and each column is a gene feature token. The function of the embedder module is to convert each cell sample to feature embeddings $\mathbf{E} \in \mathbb{R}^{(M+1) \times d}$. Here, $M$ denotes the number of columns, $+1$ indicates the special [CLS] token (see below) and $d$ is the embedding dimension. Tabula processes each row of the table as an unordered sequence of genes, where the gene in each column is represented by its column feature name embedding (gene token embedding) and its corresponding feature value embedding (gene expression value embedding). It is important to note that even though the cell-by-gene table we are encoding in Tabula solely consists of gene tokens and does not contain metadata column features, such as cell type, tissue type, or sequencing technology, it is entirely feasible to encode metadata as well and we leave it for future work.

- **Column feature name embedding**. The feature column in the cell-by-gene table for each client corresponds to individual genes. Our pretraining dataset includes 196 distinct studies, from which we select 1,200 HVGs per study, yielding a total of 23,156 HVGs across all datasets. Consequently, we have 23,156 unique feature columns to learn during the pretraining phase. We assign each column feature name, $feature_j$, a unique integer identifier, $id(feature_j)$. These identifiers collectively form the column feature name vocabulary utilized in Tabula. Additionally, we include a special token, [CLS], in the vocabulary, which is used to aggregate all column features into a cell-level representation. The input column feature tokens for cell $i$ are therefore represented as a vector $\mathbf{f}_{\text{token}}^{(i)} \in \mathbb{R}^{M}$, where $M$ is a predetermined maximum sequence length in the tabular transformer:

  $$\mathbf{f}_{\text{token}}^{(i)} = \left[\text{id}\left(\text{feature}_1^{(i)}\right), \text{id}\left(\text{feature}_2^{(i)}\right), \ldots, \text{id}\left(\text{feature}_M^{(i)}\right)\right]$$

  We then utilize the standard Embedding layer, which functions as a lookup table that maps discrete indices (e.g., unique id of column feature names) to continuous dense vectors, or embeddings. The embedding representation of input column feature tokens for cell $i$ is denoted as $\mathbf{E}_{\text{token}}^{(i)} \in \mathbb{R}^{M \times d}$ where $d$ represents the embedding dimension:

  $$\mathbf{E}_{\text{token}}^{(i)} = \text{emb}_{\text{token}}\left(\mathbf{f}_{\text{token}}^{(i)}\right)$$

  where $\text{emb}_{\text{token}}$ represents an embedding layer to learn feature token embeddings.

- **Column feature value embedding**. Different sequencing technologies, with variable sensitivity, sequencing depth, and ability to detect lowly expressed genes, resulting in different data distributions and scales, introduce complexities into scRNA-seq data modeling [66]. For instance, some platforms capture more genes but at lower depth, while others focus on a subset of genes at higher resolution [67]. These differences lead to inconsistent gene expression measurements, making it challenging to integrate data from diverse platforms for accurate modeling. To tackle this issue, we adopt scGPT's value binning technique [17] to convert expression counts into relative values. Unlike scGPT, which only focuses on non-zero expression counts in each cell, we select 1,200 HVGs for modeling in each dataset, including those with zero expression. By incorporating zero-expressed genes, our method Tabula provides a more complete view of gene expression variability, capturing both active and inactive genes. This broader perspective allows for the detection of important biological patterns that could be overlooked when considering only non-zero genes, as zero expression can have biological significance, such as in regulatory processes [68]. For each HVG expression count in a cell, we compute the raw absolute values and partition them into $B$ consecutive intervals $[Bin_t, Bin_{t+1}]$, where $t \in \{1, 2, \dots, B\}$. The binned expression value $x_j^{(i)}$ for cell $i$ is designated as:

$$x_j^{(i)} = \begin{cases} t, & \text{if } X_{i,j} > 0 \text{ and } X_{i,j} \in [Bin_t, Bin_{t+1}], \\ 0, & \text{if } X_{i,j} = 0. \end{cases}$$

Of note, for certain fine-tuning tasks, if we initially select more than 1,200 HVGs, we also filter out zero-expressed genes from within those HVGs.

The column feature values for cell $i$ are therefore represented as a vector $\mathbf{f}_{\text{val}}^{(i)} \in \mathbb{R}^M$, where $M$ is a predefined maximum sequence length in the tabular transformer:

$$\mathbf{f}_{\text{val}}^{(i)} = \left[x_1^{(i)}, x_2^{(i)}, \dots, x_M^{(i)}\right]$$

The embedding representation of input column feature values for cell $i$ is denoted as:

$$\mathbf{E}_{\text{val}}^{(i)} = \text{emb}_{\text{val}}\left(\mathbf{f}_{\text{val}}^{(i)}\right)$$

where $\mathbf{E}_{\text{val}}^{(i)} \in \mathbb{R}^{M \times d}$ , and $\text{emb}_{\text{val}}$ represents an embedding layer to learn feature value embeddings.

Consequently, the embedding $\mathbf{E}^{(i)} \in \mathbb{R}^{M \times d}$ for cell $i$ is represented as:

$$\mathbf{E}^{(i)} = \mathbf{E}_{\text{token}}^{(i)} + \mathbf{E}_{\text{val}}^{(i)}$$

Finally, the embedding of the [CLS] token is appended to the feature embedding, resulting in a final feature embedding $\mathbf{E}^{(i)} \in \mathbb{R}^{(M+1) \times d}$. The final hidden state corresponding to the [CLS] token is used as row (cell) representation aggregating from column feature embeddings. It is important to note that each client's embedder module is distinct and specialized for learning tissue-specific features.

**Tabular transformer.** The backbone of the tabular transformer is based on self-attention transformer [13]. The local tabular transformers and global tabular transformers share the same structure among clients. The tabular transformer on each client is designed to learn specific knowledge for different tissues during the pretraining stage. Our tabular transformer treats each gene column as an input token and aims to capture column gene dependency for each cell. The tabular transformer takes $\mathbf{E}^{(i)} \in \mathbb{R}^{(M+1)\times d}$ as input a set of $M$ gene column embeddings along with a [CLS] token embedding. The input to the first layer of the stacked tabular transformer is denoted as $\mathbf{E}_0^{(i)} = \mathbf{E}^{(i)}$. The output of the stacked transformer blocks is defined as follows:

$$\mathbf{E}_l^{(i)} = \text{transformer_block}\left(\mathbf{E}_{l-1}^{(i)}\right), \quad \forall l \in [1, k].$$

where $l$ represents the number of stacked transformer layers. The core function in $\text{transformer_block}$ is the attention mechanism that can be formulated as:

$$\text{Attention}(Q, K, V) = \text{Softmax}\left(\frac{QK^T}{\sqrt{d_k}}\right)V$$

where $Q = EW_q$, $K = EW_k$ and $V = EW_v$ denote linear transformations of the input $E$. Specifically, $Q$, $K$, and $V$ correspond to the query matrix, key matrix, and value matrix, respectively. $W_q$, $W_k$ and $W_v$ are training parameters. $d_k$ is the dimension of the key vectors. The softmax function is applied to the scaled dot-products of the queries and keys to produce attention weights, which are then used to weigh the values. This mechanism allows the model to focus on different column genes of the cell input dynamically, enabling it to capture complex gene dependencies effectively. To optimize the efficiency of self-attention mechanisms, we employ FlashAttention-2 [65] accelerated implementation. This advanced approach enhances its ability to handle large input dimension $M$ effectively. It is important to note that the current transformer in Tabula can be replaced with any other efficient transformer.

The output $\mathbf{E}_l^{(i)} \in \mathbb{R}^{(M+1)\times d}$ incorporates updated $M$ gene embeddings and a [CLS] embedding, which represents the cell embedding. This representation is automatically aggregated from all column gene embeddings through an attention mechanism. For pretraining and fine-tuning, the transformer output is processed by a projection head, typically a Multilayer Perceptron (MLP) [69]. This MLP handles the embeddings from the transformer, facilitating pretraining tasks like predicting masked gene columns and fine-tuning for specific tasks such as cell type classification. The [CLS] embedding can then be extracted to evaluate and visualize the effectiveness of the learned cell representations.

**Tabular pretraining objective.** Current foundation models for single cell analysis draw parallels between single cells and natural language, where texts are composed of words, and cells are defined by genes. Whether employing autoregressive self-supervised learning as in scGPT [17] or mask-reconstruction pretraining used in Geneformer[14], scFoundation[15], and scBERT[23], these methods treat scRNA-seq data as ordered gene sequences. They either use attention maps to artificially impose an order, or rank genes to create ordered sequences. Both modeling approaches contrast with the inherent unordered nature of single cell omics. Additionally, natural language consists solely of discrete word token IDs, whereas scRNA-seq data includes both discrete gene token IDs and continuous numerical gene values. Instead, we represent scRNA-seq data as tabular data, with columns corresponding to gene features and rows representing cells. Swapping any two gene columns does not alter the properties of the cells. Feature values can be either numerical or categorical; however, this work focuses exclusively on numerical values of genes while leaving metadata information for future study. To effectively model the cell-by-gene table, we design two specialized pretraining losses for tabular learning: reconstruction loss and contrastive learning to learn gene column features and row cell features respectively.

To calculate the reconstruction and contrastive losses, a corrupted view must be generated for each cell. The corrupted view is created through random feature resampling. For example, for cell $i$'s original view $\mathbf{c}^{(i)} \in \mathbb{R}^M$ where $M$ represents the number of gene columns and the gene values in $\mathbf{c}^{(i)}$ have been already binned, we randomly choose a subset of $M$ genes from cell $i$ as genes whose expression will be corrupted and resample their values from the empirical marginal distribution [64] of these genes within the training batch. We set the corrupted ratio at 60%. It is worth noting that the corrupted genes for each cell differ from one another. The corrupted cell-wise view for cell $i$ is denoted as $\tilde{\mathbf{c}}^{(i)} \in \mathbb{R}^M$.

- **Reconstruction loss as gene-wise column learning**. Reconstruction loss is a self-supervised training objective. It aims to recover the original gene-wise view from a corrupted view of that gene. Taking gene $j$ as an example, we take the corrupted view, $\tilde{\mathbf{g}}_j \in \mathbb{R}^N$, for gene $j$ as input and aim to reconstruct its original binned gene expression $\mathbf{g}_j \in \mathbb{R}^N$ where $\mathbf{g}_j = \left[\mathbf{c}_j^{(1)}, \mathbf{c}_j^{(2)}, \ldots, \mathbf{c}_j^{(N)}\right]$. The reconstructed view for gene $j$ can be represented as $\hat{\mathbf{g}}_{(j)} \in \mathbb{R}^N$. In this study, we use Mean Squared Error (MSE) to measure the distance between the original view and the reconstructed view. The reconstruction loss is denoted as:

$$\mathcal{L}_{\text{rec}} = \frac{1}{M \cdot N} \sum_{j=1}^{M} \sum_{i=1}^{N} \left(\hat{\mathbf{c}}_j^{(i)} - \mathbf{c}_j^{(i)}\right)^2$$

  It is crucial to highlight that the reconstruction loss $\mathcal{L}_{\text{rec}}$ focuses on reconstructing all $M$ gene values at the column (gene) level, not just the corrupted ones. By training the model to restore gene expressions, the model develops a deep understanding of each gene's distribution and its relationship with other genes. This leads to more accurate gene-specific feature representations. It is important to highlight that existing foundation models [14,15,23] typically assign a mask or corrupted value of -1 or 0, which does not align with the conditions encountered during the inference stage, where the gene values are not set to -1 or 0. Instead, we replace the corrupted values by resampling from the empirical marginal distribution of all corrupted genes, ensuring consistency with the inference process.

- **Contrastive loss as cell-wise row learning**. Similar to the reconstruction objective, we generate $\tilde{\mathbf{c}}^{(i)}$ as a corrupted cell-wise view of the cell $i$. $\mathbf{c}^{(i)}$ and its corresponding corrupted view $\tilde{\mathbf{c}}^{(i)}$ are treated as a positive cell pair, while $\mathbf{c}^{(i)}$and other cells $\mathbf{c}^{(j)}$ within the batch are treated as negative cell pairs. In general, contrastive loss aims to minimize the distance between positive pairs of sample cells and maximize the distance for negative pairs. In this study, we employ the SimCLR [70] loss for contrastive pretraining, which is denoted as :

$$\mathcal{L}_{\text{contrast}} = -\frac{1}{2N}\sum_{i=1}^{2N}\log\frac{\exp\left(\text{sim}\left(\mathbf{e}^{(i)}, \tilde{\mathbf{e}}^{(i)}\right)/\tau\right)}{\sum_{k=1}^{2N}\mathcal{I}_{[k\neq i]}\exp\left(\text{sim}\left(\mathbf{e}^{(i)}, \mathbf{e}^{(k)}\right)/\tau\right)}$$

where $\mathbf{e}^{(i)}$ and $\tilde{\mathbf{e}}^{(i)}$ represent the cell embedding of cell $i$ from the original view and corrupted view respectively after being processed through the transformer block. $\mathcal{I}_{[k\neq i]} \in \{0,1\}$ is an indicator function that takes the value of 1 if $k \neq i$; otherwise, it is 0. The similarity function $sim(\cdot,\cdot)$ represents cosine similarity here to measure the similarity between two cell embeddings, $\tau$ denotes the temperature parameter and $N$ is the number of cells in a batch.

The final pretraining objective in Tabula is a combination of the two loss functions:

$$\mathcal{L} = \mathcal{L}_{rec} + \mathcal{L}_{contrast}$$

The overall loss combining reconstruction and contrastive losses offers a holistic approach to learning both gene-level (column) and cell-level (row) features from the cell-by-gene table.

**Collaborative federated learning framework.** Tabula is based on the federated learning framework. Unlike existing foundation models based on centralized learning, which require aggregating all data in a central location and thus raise privacy and ethical concerns, Tabula allows data to remain distributed across multiple clients. Each client trains a local model on its own client data, and only model parameters are shared with the central server for global model updates. To be specific, the key steps are as follows:

- **Local training**. Each client $k$ (e.g., institution or hospital) trains a local copy of the model using its own data $D_k$. During this step, the model weights $w_k$ are updated by minimizing the local loss function $\mathcal{L}_k(w)$ based on the client's local dataset . The local training process can be denoted as:

  $$w_k^r = w_k^{r-1} - \eta\nabla\mathcal{L}_k(w_k^{r-1})$$

  where $r = 1, \ldots, R$ is the current round, $\eta$ is the learning rate and $\nabla\mathcal{L}_k(w_k^{r-1})$ is the gradient of the loss function at the last round weights $w_k^{r-1}$. The model weights contain two components: $w_k^{(t)}$ is the model weights for tabular transformer and $w_k^{(o)}$ is the model weights for the rest of architecture including embedders and projection heads. $w_k^r$ can be represented as $w_k^r = w_k^{r,(t)} + w_k^{r,(o)}$.

- **Sending local updates**. After every epoch of local training, each client $k$ sends its updated tabular transformer weights $w_k^{r,(t)} - w_k^{r-1,(t)}$ to a central server. These model weights represent the local knowledge learned from the individual datasets. This can be summarized as:

  $$\text{Client } k \rightarrow \text{Server} : w_k^{r,(t)} - w_k^{r-1,(t)}$$

  Tissue-specific embedder parameters and projection head weights remain strictly local, preserving both data privacy and tissue-specific representational capacity.

- **Global aggregation**. The central server aggregates the local tabular transformer updates from all clients to create a new global tabular transformer. In Tabula, we adopt a common aggregation method which is Federated Averaging (FedAvg) [71]. The global tabular transformer weights $w_g^{r,(t)}$ are updated by computing a weighted average of the local tabular transformer updates, where the number of training samples reflects the weight significance of each client. The importance of each client is quantified by a corresponding weight $p_k$.It can be summarized as:

$$P = \sum_{k=1}^{K} p_k$$

$$w_g^{r,(t)} = w_g^{r-1,(t)} + \frac{1}{P} \sum_{k=1}^{K} p_k \left( w_k^{r,(t)} - w_k^{r-1,(t)} \right)$$

- **Broadcast global update**. After aggregation, the global tabular transformer is updated with the newly averaged weights $w_g^r$.This updated global model is then broadcasted back to the clients, which subsequently update their local models based on the received global weights. This process can be expressed as:

$$\text{Server} \rightarrow \text{Clients } k : w_g^{r,(t)}$$

$$w_k^r = w_g^{r,(t)} + w_k^{r,(o)}$$

Those processes are repeated for several rounds, with clients continuing to train the model locally, send updates, and the server performing global aggregation, until the model converges or reaches the desired performance. In Tabula, we update the global model once every epoch and treat each client with the same importance weight.

Our federated pretraining framework offers three key advantages. First, it ensures data privacy, as sensitive information remains localized on the client side, preventing unnecessary data transfers. Second, it distributes the computational burden across multiple clients, alleviating the demand for resources on any single entity and enhancing scalability. Third, the federated architecture naturally enables the development of tissue-specific embedders and projection heads, which are designed to capture tissue-specific features. In contrast, conventional foundation models rely on a single, generalized embedder, making them less effective at capturing the nuances of tissue-specific variations. It is important to note that in our federated training framework, both the local model and the tissue-specific embedder, along with the projection head, can be shared. This allows for greater flexibility and collaboration across clients while preserving the benefits of tissue-specific feature learning.

**Similarity and innovation compared to existing foundation models.** Compared to existing foundation models for single cell data, Tabula presents four key innovations. First, Tabula utilizes federated learning instead of centralized training, ensuring enhanced data privacy by allowing institutions to keep sensitive data local while only sharing model updates. Second, Tabula introduces an innovative approach by treating single cell data as tabular data and designing two distinct pretraining losses specifically for tabular modeling. Reconstruction loss is designed for column-level learning to capture gene-related features, while contrastive loss focuses on row-level learning to highlight distinctions among individual cells. This dual mechanism allows the model to have a holistic view to understand tabular data. Third, we are the first to propose tissue-specific embedders and projection heads in the field of foundation models for single cells. This novel approach allows the model to tailor its representations to the unique characteristics of each tissue type, enabling more precise feature extraction compared to traditional models that rely on a single, generalized embedder. Finally, Tabula is a lightweight foundation model, achieving comparable performance to larger models like scGPT with significantly fewer parameters, making it more efficient and scalable for real-world applications.

## Fine-tuning on downstream tasks and analysis

Following the growing practice of evaluating single-cell deep learning models across standardized downstream tasks, as exemplified by benchmark platforms such as DANCE[72,73], we assessed Tabula on a broad suite of cell-wise and gene-wise tasks, including cell type annotation, multi-omics integration, multi-batch integration, perturbation prediction, reverse perturbation prediction, and imputation.

- Cell type annotation
  The performance of cell type classification was evaluated through metrics including accuracy, precision, recall, and F1 score (See **SUPPLEMENTARY NOTES S1**), offering a well-rounded view of Tabula's predictive accuracy against other models. Additionally, the confusion matrix was built to reveal patterns of predictions across cell types.

  To fine-tune the pretrained Tabula for the cell type annotation task, an MLP classifier is introduced. This classifier takes cell embeddings generated by the Tabular Transformer as input and outputs categorical predictions for cell types. The entire model is optimized using cross-entropy loss $\mathcal{L}_{AN}$, as depicted below:

$$a_j^{(i)} = MLP(\mathbf{E}_{cls}^{(i)})$$

$$\mathcal{L}_{AN} = -\frac{1}{N}\sum_{i=1}^{N}\sum_{c=1}^{C} y_{i,c}\log(a_{j,c}^{(i)})$$

  Here, the MLP layer outputs the predicted probabilities of cell type labels $a_j^{(i)}$ taking the cell representation $\mathbf{E}_{cls}^{(i)}$ as input. The cross-entropy loss, $\mathcal{L}_{AN}$, is computed where $N$ denotes the total number of samples in a batch, $C$ is the number of cell type classes, $y_{i,c}$ represents the true label for $i$-th sample for class $c$, encoded in one-hot format, and $a_{j,c}^{(i)}$ is the predicted probability for the $i$-th sample that belongs to cell type class $c$.

  We assessed our model using four datasets including Human pancreas (hPancreas), Myeloid, Cell Lines, and Liver datasets (See **Downstream Task Datasets**). Benchmark comparisons were made against scGPT, Geneformer, and scFoundation. The fine-tuning settings of these benchmark models follow their default settings.

- Multi-omics integration
  To assess Tabula's performance in integrating cell embeddings of multi-omics data after fine-tuning, we employ biological conservation metrics[74,75] including Normalized Mutual Information ($\mathrm{NMI}_{\mathrm{cell}}$), Adjusted Rand Index ($\mathrm{ARI}_{\mathrm{cell}}$), and Average Silhouette Width ( $\mathrm{ASW}_{\mathrm{cell}}$) (See **SUPPLEMENTARY NOTES S2**). These metrics quantify the alignment between cluster formations derived from the embeddings and the validated ground truth cell type annotations. Consistent with methodologies used in previous benchmarking studies [17], we calculated the average of these three metrics ($\mathrm{AvgBIO}$) to provide a composite score reflecting overall performance in the task.

  During the fine-tuning phase, the model weights are refined using multiple loss functions to enhance the learning of aligning different omics modalities. Specifically, the Tabula employs three fine-tuning losses: the reconstruction loss $\mathcal{L}_{\mathrm{rec}}$ from the pretraining stage (See **Tabular pretraining objective**), masked gene modeling (MGM) by applying masking to random-selected genes, and cell context-aware masked gene modeling (CMGM), which integrates cell-level information, as detailed below:

  - Masked gene modeling
    To elucidate gene interrelationships, Tabula employs MGM loss as a fine-tuning objective. This approach parallels the reconstruction loss employed during the pretraining but modifies the handling of genes at these masked positions. Specifically, whereas the reconstruction loss utilizes a corrupted view for masking, MGM involves the random masking of a subset of gene expression values $e^i$ in each input cell. Tabula is then optimized to predict these masked expression values accurately. This setup enables the model to infer gene expression patterns from other genes within cells, thereby enhancing inter-gene awareness. The training objective is quantified using the mean squared error (MSE) at the masked positions, denoted as $\mathcal{M}_{\mathrm{mask}}$. The formula is depicted below:
    $$\hat{x}^{(i)} = MLP\left(concat(\mathbf{E}_k^{(i)}, emb_m(t_m^{(i)}))\right)$$
    $$\mathcal{L}_{\mathrm{MGM}} = \frac{1}{|\mathcal{M}_{\mathrm{mask}}|} \sum_{j \in \mathcal{M}_{\mathrm{mask}}} (\hat{x}_j^{(i)} - x_j^{(i)})^2$$
    Here, $\mathbf{E}_k^{(i)}$ denotes the gene embeddings from the Tabular Transformer layers. By concatenating the embedding of modality information $emb_m(t_m^{(i)})$ generated by a modality encoder given a modality label $t_m^{(i)}$, the integrated embedding will be passed to an MLP to predict the gene expression value $\hat{x}^{(i)}$ for cell $i$.
  - Cell context-aware masked gene modeling
    Following the previous work[17], the CMGM loss advances model optimization by predicting gene expression values through incorporating cell representation,

which promotes contextual awareness between cells and genes and enhances the efficacy of cell representation learning. Specifically, a gene-specific query vector $q_j$ is created given the gene token embedding $\mathbf{E}_{token}^{(i)}$ and calculates the predicted expression value using the parameterized inner product between $q_j$ and the cell representation $\mathbf{E}_{cls}^{(i)}$:

$$q_j = MLP(\mathbf{E}_{token}^{(i)})$$

$$\hat{x}_j^{(i)} = q_j \cdot W\mathbf{E}_{cls}^{(i)}$$

$$\mathcal{L}_{\mathrm{CMGM}} = \frac{1}{|\mathcal{M}_{\mathrm{mask}}|} \sum_{j \in \mathcal{M}_{\mathrm{mask}}} (\hat{x}_j^{(i)} - x_j^{(i)})^2$$

Here, $\mathbf{E}_{cls}^{(i)}$ denotes the cell-level embedding extracted from the [CLS] token. $\mathcal{L}_{\mathrm{CMGM}}$ is computed as MSE loss between the predicted expression value and true value, as similar to $\mathcal{L}_{MGM}$.

- Multi-batch integration

  To access modeling performance on multi-batch integration, biological conservation metrics [74], especially the inverse of the average silhouette width for batch clustering ( $ASW_{batch}$) and graph connectivity ($Graph_Conn$) are employed. The overall score, $Avg_{batch}$ is computed by averaging the two metrics (See **SUPPLEMENTARY NOTES S2**).

  During the fine-tuning phase, a domain adaptation (DA) loss is utilized to enable the model to optimize batch correction by reversing the backpropagation of learned batch labels. Following the methodology used by scGPT, we initiate an MLP classifier that predicts the batch label based on the cell representation $\mathbf{E}_{cls}^{(i)}$. Through a reverse back-propagation procedure, the model will aggregate batch information by integrating the reversed gradients, thereby enhancing the robustness of the batch correction process. Cross-entropy loss is employed for model finetuning. The formula is illustrated as follows:

$$b_j^{(i)} = MLP(\mathbf{E}_{cls}^{(i)}),$$

$$\mathcal{L}_{DA} = -\frac{1}{N} \sum_{i=1}^{N} \sum_{c=1}^{C} y_{i,c} \log(b_{j,c}^{(i)})$$

  Here, the MLP layer outputs the predicted probabilities of batch labels $b_j^{(i)}$ using the cell representation $\mathbf{E}_{cls}^{(i)}$ as the input. The domain adaptation loss, $\mathcal{L}_{DA}$, is computed as the negative categorical cross-entropy loss, where $N$ is the total number of samples in a batch, $C$ is the total number of batch classes, $y_{i,c}$ is the true label for $i$-th sample for class $c$, in one-hot encoded form, and $b_{j,c}^{(i)}$ is the predicted probability for the $i$-th sample belongs to batch class $c$.

We conducted a comprehensive evaluation of multi-omics and multi-batch integration tasks within a unified framework, with four losses: $\mathcal{L}_{\text{rec}}$, $\mathcal{L}_{MGM}$, $\mathcal{L}_{\text{CMGM}}$, and $\mathcal{L}_{DA}$, optimized together according to the batch and omic features in the dataset. For these evaluations, we use datasets including 10x Multiome PBMC, BMMC, Perirhinal Cortex, Human dendritic cells (See **Downstream Task Datasets**). Benchmark comparisons were made against scGPT [17], Geneformer [14], and scBERT [23]. The fine-tuning protocols for these benchmarking models adhered to the methods described for scGPT.

- Genetic perturbation prediction

  We assessed the performance on genetic perturbation prediction using two key metrics: $Pearson_{delta}$ that quantifies the correlation between predicted and actual expression changes after perturbation, and $\text{DEG Pearson}_{delta}$ that specifically evaluates this correlation for the top 20 most differentially expressed genes (See **SUPPLEMENTARY NOTES S3**). During the model testing, there are four distinct scenarios pertaining to gene perturbation visibility as defined in GEARS [76]: (i) 1/1 Unseen: single-gene perturbations are not previously observed during the model training; (ii) 2/2 Unseen: two-gene perturbations that neither gene are not previously observed during the model training; (iii) 1/2 Unseen: two-gene perturbations where one gene is previously unobserved during the model training; and (iv) 0/2 Unseen: two-gene perturbations with both genes having been previously observed during the model training. These evaluations help elucidate the model's predictive robustness across varying levels of experimental uncertainty.

  During the fine-tuning stage, a perturbation encoder (MLP) is initialized for incorporating perturbation information through adding a perturbation embedding $emb_p(t_p^{(i)})$ to the original input. Then, $\hat{x}^{(i)}$ is computed by passing the integrated embedding to the transformer layers and a reconstruction head. The $\mathcal{L}_{pert}$ is calculated as MSE loss across all the positions. The process is outlined in the formula below:

$$\hat{x}^{(i)} = MLP\left(\text{transformer_block}(addition(\mathbf{E}_{token}^{(i)}, \mathbf{E}_{val}^{(i)}, emb_p(t_p^{(i)})))\right)$$

$$\mathcal{L}_{pert} = \frac{1}{M}\sum_{j=1}^{M}(\hat{x}_j^{(i)} - x_j^{(i)})^2$$

  Here, $\mathbf{E}_{token}^{(i)}$ and $\mathbf{E}_{val}^{(i)}$ are representation embeddings of gene token and gene expression value, respectively.

  We evaluated our model using three Perturb-seq datasets derived from leukemia cell lines, including Adamson, Norman, and Replogle (See **Downstream Task Datasets**). Adhering to preprocessing protocols established by the recent perturbation prediction methodology, GEARS [76], we benchmarked Tabula against scGPT, scBERT, and Geneformer. This comparison was conducted in a consistent fine-tuning framework, utilizing gene-level embedding and perturbation information to predict perturbed expression profiles.

- Reverse perturbation prediction
  For the reverse perturbation prediction task, we aimed to identify perturbation conditions that best replicate observed gene expression profiles from a query set. We followed the setting of scGPT [17] to select a subset of 20 genes from the Norman dataset, which consisted of 39 cases for training, 3 for validation, and 7 for testing. These 20 genes include a perturbation space comprising 210 combinations, encompassing both single-gene and double-gene perturbations. For each test case, we treated the ground truth gene expression profiles under the test perturbations as queries and all 210 perturbation conditions as references for retrieval. Following the methodology outlined in scGPT, we formulated the retrieval process as a top-K retrieval task with a two-round selection strategy involving Euclidean distance-based ranking and ensemble voting to identify the most-likely perturbation conditions.

  To construct the reference database, we generated 30 distinct gene expression profiles for each of the 210 perturbation conditions by randomly sampling control cells, resulting in 6,300 post-perturbation profiles:

  $$\mathbf{D}_{\mathrm{ref}} = \{\hat{y}_{\mathrm{pert}}^{(j)} \mid j = 1, 2, \ldots, 210\} \times 30$$

  This approach allowed for increased diversity and robustness in representing each perturbation condition. For each test perturbation $X + Y$, the ground truth gene expression profiles from cells undergoing $X + Y$ perturbation were treated as the query set:

  $$\mathbf{Q}_{X+Y} = \{\mathbf{y}_{\mathrm{true}}^{(X+Y)} \mid \text{perturbation conditions with } X + Y \text{ genes}\}$$

  The retrieval strategy involved two rounds. In the first round, each cell within the query set, $Q_{\mathrm{X+Y}}$, identified its top K closest profiles from the reference database, $D_{\mathrm{ref}}$, by minimizing the Euclidean distance:

  $$\mathbf{K}_i = \mathrm{argmin}_{\hat{y}_{\mathrm{pert}}^{(j)} \in \mathbf{D}_{\mathrm{ref}}} \left( \sum_{m=1}^{M} \left( \mathbf{y}_{\mathrm{true},m}^{(X+Y)} - \hat{y}_{\mathrm{pert},m}^{(j)} \right)^2 \right),$$

  where $K_i$ denotes the set of top K most similar expression profiles for each query cell from $D_{\mathrm{ref}}$. In the second round, we implemented an ensemble voting scheme: each query cell cast a vote for its top K matches within $D_{\mathrm{ref}}$. Perturbation conditions were then ranked by the total number of votes received from all cells in the query set:

  $$\mathbf{R}_K = \mathrm{top_vote}(\{\mathbf{K}_i \mid i = 1, 2, \ldots, N\})$$

  The conditions with the highest votes were identified as the predicted perturbation conditions, which we reported in descending order based on vote count.

  For evaluation, we employed a modified top-K accuracy metric to assess retrieval performance. This metric accounted for two criteria: (1) exact matches, where the predicted perturbation condition precisely matched the ground truth condition, and (2) partial matches, where at least one gene in the predicted condition overlapped with the actual perturbation combination:

$$\text{Accuracy}_K = \frac{1}{|Q|} \sum_{i \in Q} \delta(\mathbf{R}_K^{(i)}, \mathbf{y}_{\text{true}}^{(i)})$$

Here, $\delta$ denotes the evaluation function that returns a score of 1 for an exact or partial match, and $Q$ represents the query set.

To ensure robustness, we averaged retrieval performances across five random seeds. We benchmarked Tabula against scGPT on both exact match and partial match settings.

- Imputation
  We assessed the imputation performance using several quantitative metrics (See **SUPPLEMENTARY NOTES S4**). The primary evaluation involves the Pearson correlation coefficient (Pearson), Root Mean Squared Error (RMSE), and Mean Absolute Error (MAE) to measure the accuracy of imputed values compared to the true values. To demonstrate the model's robustness, we applied different masking ratios (10%, 20% and 30%) to the dataset, with Pearson correlation as one of the metrics. Additionally, the model's performance is analyzed through Mean Squared Error (MSE) at both the cell-wise and gene-wise levels. To assess the quality of imputation, cosine similarity is calculated between different cell types. This helps evaluate how well the imputed data preserves the original structure and relationships of the data. Specifically, it provides insights into the accuracy of the imputation and ensures that the natural diversity and variability between cell types remain consistent before and after the imputation process.

  Since the true dropout values are unknown, we evaluated the method's performance by randomly masking the expression matrix in the scRNA-seq dataset, replacing these values with zeros. To more accurately simulate the distribution of dropout values, we adopted the settings from DeepImpute[77], and utilized an exponential kernel[16]. For each cell, we identify genes with non-zero expression values, excluding cells with few expressed genes. For these non-zero genes, we compute sampling probabilities using an exponential distribution and normalize them to sum to 1. Using these probabilities as weights, we randomly sample a subset of the non-zero genes without replacement. The expression values of the selected genes are then masked to zero to simulate dropout genes, while the remaining genes retain their original expression values. The model is then used to reconstruct the true expression values of these masked genes. During the fine-tuning process, the masked data is divided into three sets: the training set, the validation set, and the test set. For each set, predictions are made exclusively for the positions within that specific set. Subsequently, an imputation decoder (MLP) is introduced for each gene, with the objective of predicting the expression levels at the respective positions. $\hat{x}^{(i)}$ is computed by passing the integrated embedding to the transformer layers and imputation decoder. We calculate the imputation loss $\mathcal{L}_{imp}$ in masked positions. The process is outlined in the formula below:

$$\hat{x}^{(i)} = MLP\left(\text{transformer_block}(addition(\mathbf{E}^{(i)}_{token}, \mathbf{E}^{(i)}_{val}))\right)$$

$$\mathcal{L}_{imp} = \frac{1}{|\mathcal{M}_{\text{mask}}|} \sum_{j \in \mathcal{M}_{\text{mask}}} \left(\hat{x}^{(i)}_j - x^{(i)}_j\right)^2$$

Here, $\mathbf{E}^{(i)}_{token}$ and $\mathbf{E}^{(i)}_{val}$ are representation embeddings of gene token and gene expression value, respectively.

We assessed our model using four datasets derived including PBMC5K, Jurkat, Melanoma, and hPancreas (See **Downstream Task Datasets**). We benchmarked Tabula against scGPT, scBERT, and Geneformer. This comparison was conducted in a consistent fine-tuning framework, utilizing gene-level embeddings for the prediction of dropout expression profiles.

- In silico gene regulatory network inference

  For the in silico inference on gene regulatory networks, we performed an iterative process on our pretrained Tabula model to explore the impact of incremental increases in gene A expression on gene B. Initially, the expression levels for all genes in a given cell are captured, with specific focus on genes A and B, denoted as $y_0$ and $x_0$, respectively. At the initial conditions, gene expressions of a cell are illustrated as $\mathbf{G}_0 = \{g_{1,0}, g_{2,0}, \ldots, g_{n,0}\}$, where $g_{i,0}$ represents the initial expression level of the $i$-th gene, with $g_{A,0} = y_0$ and $g_{B,0} = x_0$. Then, each iteration involves modifying the expression level of gene A by incrementing it by one, while maintaining the context of all other genes from the last iteration's output: $g_{A,t+1} = g_{A,t} + 1$. The entire gene expression profile, including the updated gene A, is input into the Tabula model, the output is then obtained from the pretrained reconstruction head. The iterative process is illustrated below:

$$\mathbf{G}_{t+1} = f(\mathbf{G}_t, g_{A,t+1}; \theta)$$

  Here, $\mathbf{G}_t$ represents the gene expression profile at time $t$, and $f$ is the function modeled by the pretrained Tabula, encapsulating the parameter $\theta$. The continuous output expression value of gene B is recorded at each iteration step. This pseudo time-series data of gene B's expression levels provides a quantitative measure of the dynamic response to the systematic perturbation of gene A. Let $X = \{x_1, x_2, \ldots, x_T\}$ denotes the sequence of gene B's expression levels over $T$ iteration steps. Based on the continuous output sequence for gene B, one can plot the trajectory of gene B's expression levels across the iteration steps. By observing the trend in these values, it is possible to infer the regulatory effect of gene A on gene B. Should the expression levels of gene B consistently increase, this suggests that gene A acts as an activator. Conversely, a decline in gene B's expression levels would imply a suppressive influence by gene A.

  Building upon the in silico perturbation prediction of pair-wise gene regulation e, we extended our approach to combinatorial regulation predictions. In this more complicated scenario, we conducted in silico perturbations on multiple genes simultaneously and observed the continuous output expression of the target gene, following the logic established above. We accessed gene regulatory rules across four biological systems including hematopoiesis, pancreatic endogenesis, neurogenesis, and cardiogenesis (See **Downstream Task Datasets**). For the evaluation of combinatorial regulation logic, our analysis is concentrated on the hematopoiesis and cardiogenesis systems, the only two systems with very known combinatorial gene regulation logic.

## Tabula-guided computational discovery of rejuvenation factors

**Post-training of Tabula on skin fibroblasts aging data.** To further adapt the pretrained Tabula model to aging-related and fibroblast identity-related biological context in the longitudinal skin aging dataset, we performed an additional post-training stage in which reconstruction was constrained to preserve the predefined age and identity scores of the input cells. In each mini-batch, cells were grouped by file identity corresponding to different biological samples (`fid`), and score targets were computed separately within each subgroup using the corresponding dataset-specific age, and fibroblast marker gene sets. For each subgroup, age and identity scores were first computed from the input expression matrix and then from the reconstructed expression matrix, and score-consistency losses were defined as the mean squared error between the reconstructed and input scores. During the main post-training stage, these score-consistency terms were optimized jointly with the original contrastive and reconstruction objectives, with the total loss defined as

$$\mathcal{L}_{\mathrm{post}} = \mathcal{L}_{\mathrm{con}} + \mathcal{L}_{\mathrm{rec}} + 10\,(7\,\mathcal{L}_{\mathrm{age}} + 63\,\mathcal{L}_{\mathrm{id}})\,.$$

Here, $\mathcal{L}_{\mathrm{con}}$ denotes the contrastive loss, $\mathcal{L}_{\mathrm{rec}}$ denotes the reconstruction loss, and $\mathcal{L}_{\mathrm{age}}$ and $\mathcal{L}_{\mathrm{id}}$ denote the score-consistency losses for the normalized age and identity scores, respectively. The coefficients 7 and 63 were chosen according to the ratio between the numbers of genes contributing to the corresponding score-related gene sets, among the 1,200-gene input dimension. We further chose 10 for the whole reconstructive objective term to ensure the faithful reconstruction of the aging and identity score.

**Age and identity score construction in longitudinal skin aging dataset**. For each fibroblast cell, we computed two gene set-based scores, namely an age score and an identity score, using a cohort-mean fold-change scheme, revealing the relative expression level with respect to the global baseline. Let $x_{ig}$ denote the normalized expression of gene $g$ in cell $i$, and let

$$\bar{x}_g = \frac{1}{N}\sum_{i=1}^{N} x_{ig}$$

denote the mean expression of gene $g$ across all fibroblast cells in the dataset. For any predefined gene set $G$, we defined the corresponding per-cell fold-change score as

$$\mathrm{FC}_G(i) = \frac{1}{|G'|}\sum_{g\in G'} \frac{x_{ig}}{\max(\bar{x}_g, 10^{-8})},$$

where $G'$ denotes the subset of genes in $G$ that were present in the expression matrix. The age score was defined as

$$\mathrm{AgeScore}_i = \mathrm{FC}_{\mathrm{AGE}}(i) - \mathrm{FC}_{\mathrm{YOUNG}}(i),$$

where $\mathrm{AGE}$ and $\mathrm{YOUNG}$ denote the predefined age-associated and young-associated gene sets, respectively. The identity score was defined as

$$\mathrm{IdentityScore}_i = \mathrm{FC}_{\mathrm{FB19}}(i),$$

where $\mathrm{FB19}$ denotes the curated fibroblast identity gene set of 19 genes. Thus, the age score increases when age-associated genes are elevated relative to the cohort mean and/or young-associated genes are reduced, whereas the identity score reflects the average relative expression of fibroblast marker genes.

**Discriminator MLP training**. To provide a differentiable surrogate objective for downstream optimization, we trained an independent discriminator multilayer perceptron (MLP) for each fibroblast lineage pair. In all settings, the MLP for a given lineage was trained only on cells from that lineage. The difference between the Patient specific HVGs and Patient union HVGs settings lay only in how the input gene list was defined: in the Patient specific HVGs setting, the input features were the lineage-specific HVGs for that lineage, whereas in the Patient union HVGs setting, the input features were selected from HVGs identified from all cells across the three cell lines, but the MLP was still trained separately for each lineage.

For a cell $i$, let $x_i \in \mathbb{R}^p$ denote its expression vector restricted to the selected input gene list, where $p$ is the number of input genes in the current setting. The corresponding lineage-specific discriminator $f_{\theta^{(\ell)}}$ was defined as

$$f_{\theta^{(\ell)}}(x_i) = W_2 \tanh(W_1 x_i + b_1) + b_2,$$

with a hidden dimension of 64 and a two-dimensional output

$$f_{\theta^{(\ell)}}(x_i) = (\hat{a}_i, \hat{d}_i),$$

where $\hat{a}_i$ and $\hat{d}_i$ denote the predicted age score and identity score, respectively. In implementation, this corresponds to a feedforward network with a linear layer from the input dimension to a 64-dimensional hidden layer, followed by a Tanh activation and a final linear layer mapping to two outputs.

For each lineage-specific model, parameters were optimized by minimizing the mean squared error between predicted and target scores:

$$\mathcal{L}^{(\ell)}_{\mathrm{MLP}} = \frac{1}{N_\ell} \sum_{i=1}^{N_\ell} \left[ (\hat{a}_i - a_i)^2 + (\hat{d}_i - d_i)^2 \right],$$

where $a_i$ and $d_i$ are the normalized age and identity scores of cell $i$, and $N_\ell$ is the number of cells in lineage $\ell$. After training, the parameters of each lineage-specific MLP were frozen and used only for score evaluation during the subsequent in silico optimization procedure.

**Definition of target score of the rejuvenation trajectory prediction**. Aged fibroblasts were optimized through a frozen Tabula reconstruction model rather than by directly scoring the observed aged-cell matrix. Let $R_\phi$ denote the pretrained reconstruction module with fixed parameters $\phi$. For an input matrix $X \in \mathbb{R}^{n\times p}$, where $n$ is the number of aged cells and $p$ is the number of input genes, the reconstruction model outputs

$$\widetilde{X} = R_\phi(X).$$

To define the optimization targets, we first evaluated the matched young and old fibroblast groups using the frozen lineage-specific MLP. Let $c(\cdot)$ denote the population-level summary statistic used to define the center of a score distribution; in the current implementation, $c(\cdot)$ was taken to be the median across cells. Here, both input space and output space refer to Tabula-processed expression matrices: **input space** refers to the Tabula-guided input expression matrix that is directly optimized, whereas **output space** refers to the matrix reconstructed by Tabula from that input. Based on the evaluation space we choose, we computed the target scores as:

$$\begin{aligned} a_{\text{young}} &= c\Big(f_{\theta^{(\ell)}}(Z_{\text{young}})_{[:,0]}\Big)\,, \\ a_{\text{old}} &= c\Big(f_{\theta^{(\ell)}}(Z_{\text{old}})_{[:,0]}\Big)\,, \\ d_{\text{young}} &= c\Big(f_{\theta^{(\ell)}}(Z_{\text{young}})_{[:,1]}\Big)\,, \end{aligned}$$

where $Z_{\text{young}}$ and $Z_{\text{old}}$ denote the matrices used for calibration. In the current implementation, calibration was performed in the output space by default, that is, $Z_{\text{young}} = R_\phi(X_{\text{young}}), \qquad Z_{\text{old}} = R_\phi(X_{\text{old}})$.

The lower-bound age target was then defined by linear interpolation or extrapolation along chronological age. If $\tau_{\text{young}}$ and $\tau_{\text{old}}$ are the chronological ages of the matched young and old samples, and $\tau_{\text{LB}}$ is the target age used for rejuvenation calibration, then the target age score was defined as

$$a^\star = a_{\text{young}} + \frac{a_{\text{old}} - a_{\text{young}}}{\tau_{\text{old}} - \tau_{\text{young}}}\left(\tau_{\text{LB}} - \tau_{\text{young}}\right).$$

In the current setting, $\tau_{\text{LB}} = 17$. The identity target was defined as the young-group center, $d^\star = d_{\text{young}}$.

**Tabula-guided in silico prediction of rejuvenation trajectory**. Let $X^{(0)}$ denote the observed aged-cell expression matrix for one lineage and one input-gene setting. We treated $X$ itself as the optimization variable and iteratively updated it while keeping both the reconstruction model $R_\phi$ and the discriminator $f_{\theta^{(\ell)}}$ frozen. At optimization step $t$, the current matrix $X^{(t)}$ was first passed through the reconstruction model:

$$\widetilde{X}^{(t)} = R_\phi\big(X^{(t)}\big)\,.$$

The reconstructed matrix $\widetilde{X}^{(t)}$ was then evaluated by the lineage-specific MLP:

$$(\widehat{A}^{(t)}, \widehat{D}^{(t)}) = f_{\theta^{(\ell)}}\Big(\widetilde{X}^{(t)}\Big)\,,$$

where $\widehat{A}^{(t)} \in \mathbb{R}^n$ and $\widehat{D}^{(t)} \in \mathbb{R}^n$ are the cell-level predicted age and identity scores.

The optimization loss combined age and identity objectives:

$$\mathcal{L}_{\text{age}}^{(t)} = \frac{1}{n}\sum_{i=1}^{n}\left(\widehat{A}_i^{(t)} - a^{\star}\right)^2,$$

$$\mathcal{L}_{\text{id}}^{(t)} = \frac{1}{n}\sum_{i=1}^{n}\left(\widehat{D}_i^{(t)} - d^{\star}\right)^2,$$

and the total loss was defined as

$$\mathcal{L}^{(t)} = \mathcal{L}_{\text{age}}^{(t)} + \alpha\,\mathcal{L}_{\text{id}}^{(t)},$$

where $\alpha$ controls the weight assigned to the identity-preservation term. In the current implementation, $\alpha = 5$. Importantly, the loss was computed in output space, so gradients propagated through the frozen reconstruction model back to the optimized aged-cell matrix.

**Trajectory-derived gene statistics**. To summarize gene behavior along the predicted rejuvenation trajectory, we recorded gene-wise changes in both input space and output space at every predicted rejuvenation step. For each gene $g$, we computed its mean stepwise change across cells between two consecutive iterations, denoted by $\delta_g^X(t)$ in input space and $\delta_g^R(t)$ in output space. These quantities capture whether the expression of a gene tends to increase or decrease, on average, as optimization proceeds.

Based on these stepwise changes, we further derived cumulative path statistics for each gene. Specifically, $\text{pos_path}$ represents the accumulated magnitude of positive stepwise changes across iterations, whereas $\text{neg_path}$ represents the accumulated magnitude of negative stepwise changes. We computed these quantities separately in input space and output space. In parallel, we also counted how many optimization steps each gene increased or decreased, recorded as $n_{\text{up}}$ and $n_{\text{down}}$, respectively. Intuitively, $n_{\text{up}}$ measures how consistently a gene moves upward during optimization, while $\text{pos_path}$ reflects the total extent of that upward movement.

In addition to these trajectory-based statistics, we calculated the final net change of each gene as the mean difference between the optimized endpoint and the initial aged-cell matrix, again in both input space and output space. Together, these quantities summarize the direction, consistency, and magnitude of gene-level changes along the optimization process, and were exported for downstream candidate ranking.

**Rejuvenation factor prioritization and overlap analysis**. Candidate rejuvenation factors were ranked from the trajectory-derived statistics within each lineage and under each input-gene setting. Genes were ordered first by $n_{\text{up}}$ in descending order and then, when ties occurred, by $\text{pos_path}$ in descending order. For overlap analysis, the top 500 genes from each lineage-specific ranking were compared across groups using UpSet-based set comparison.

For summary analyses across the three biological groups, a consensus ranking was obtained by averaging lineage-specific ranks. Let $r_g^{(m)}$ denote the rank of gene $g$ in lineage $m \in \{A, B, H\}$. The consensus rank was defined as

$$\bar{r}_g = \frac{1}{3} \sum_{m \in \{A,B,H\}} r_g^{(m)}.$$

Genes with smaller $\bar{r}_g$ were considered higher-priority candidates in the cross-lineage summary ranking.

**Gene ranking enrichment analysis using mSALT**. To quantify how well each ranking strategy prioritized biologically relevant genes, we compared the top-ranked genes from each setting against an external mSALT-derived reference set. The mSALT-positive set $R$ was defined from the mSALT summary table as genes satisfying *Slope* > 0 and *p-value* < 0.05. We evaluated two types of HVG settings: Patient specific HVGs and Patient union HVGs. For Patient specific HVGs, each group (A, B, and H) had its own lineage-specific HVG set. To ensure a common candidate universe across groups, we first took the intersection of genes shared by all three groups, $A \cap B \cap H$. This shared lineage-HVG set was then intersected with the mSALT gene universe to define the baseline candidate pool, $L = (A \cap B \cap H) \cap \mathrm{mSALT}.$

For Patient union HVGs, the candidate universe was defined by the 1,200-gene HVG set identified from cells across all three groups and then intersected with the mSALT universe, $U = \mathrm{SharedHVG} \cap \mathrm{mSALT}.$ Within each setting, genes were first ranked within each lineage using the current rule, and then aggregated across lineages by average rank.

For a given cutoff $k$, let $Top_k$ denote the top-$k$ genes in the aggregated ranking. The number of mSALT-positive genes $R$ among the top-$k$ candidates was defined as $P_k = |Top_k \cap R|,$ so that the observed positive fraction was $\frac{P_k}{k}.$

For the matched baseline pools, we defined $N_L = |L|,\ Q_L = |L \cap R|$ for Patient specific HVGs, and $N_U = |U|,\ Q_U = |U \cap R|$ for Patient union HVGs. The corresponding baseline positive fractions were therefore $\frac{Q_L}{N_L}$ and $\frac{Q_U}{N_U}.$

The enrichment score shown in the figure was calculated as the fold change of the observed top-$k$ positive fraction over the matched baseline fraction:

$$\mathrm{Fold}_{\mathrm{Lineage}} = \frac{P_k/k}{Q_L/N_L},$$

$$\mathrm{Fold}_{\mathrm{Union}} = \frac{P_k/k}{Q_U/N_U}.$$

A fold value greater than 1 indicates that the ranked candidates are enriched for mSALT-positive genes beyond baseline expectation. In the current figure setting, enrichment was evaluated at $k = 10$. In addition to the HVG-based analyses, we also evaluated the inferred young gene set obtained from the whole-transcriptome analysis (**Supplementary Table 1**). Let $Y$ denote the inferred young gene set and let $W$ denote the whole-transcriptome background after restriction to genes present in the mSALT table. The corresponding fold enrichment was defined as

$$\mathrm{Fold}_{\mathrm{Young}} = \frac{|Y \cap R|/|Y|}{|W \cap R|/|W|}.$$

This whole-transcriptome comparison provided a reference for assessing whether the inferred young gene set showed preferential support from independent longevity intervention data.

## Chiron federated training platform

The Tabula federated training infrastructure is deployed as Chiron, a decentralized training platform built on BioEngine[29] v0.11. Geographically distributed workers connect through secure Hypha services, and the training workflow is organized as three BioEngine applications constituting the management, coordination, and local training tiers.

**Worker node deployment.** Each BioEngine Worker node hosts a pair of isolated containers launched via the standard BioEngine deployment. The data server exposes local single-cell datasets to the co-located training container via authenticated HTTP streaming with per-dataset access control and short-lived URL tokens. The data server is accessible to the training container exclusively over the container-internal network, and no raw data traverses external channels. To lower the barrier to participation, the Chiron platform provides a browser-based setup wizard (https://chiron.aicell.io/#/worker) that generates a ready-to-run launch command from a short form. The operator selects a container runtime (Docker Compose, Podman, Singularity, or Apptainer), enters a worker name and the path to their local dataset directory, and sets hardware parameters (CPU cores, GPU count, and memory). The wizard outputs either a Docker Compose file or a shell command tailored to the chosen runtime, which the operator runs locally to connect their site to the federated network without any changes to central infrastructure.

**Dataset onboarding.** Each BioEngine Worker exposes a host-mounted directory in which the site operator lays out one subfolder per dataset, with each subfolder containing the AnnData[78] files for that dataset and a manifest.yaml declaring a unique dataset identifier, an authorized-users list, and optional bibliographic metadata. The data server reads .h5ad files directly and converts them to the Zarr v3 format[79] representation on first access, rescanning the directory every 30 s so newly-added datasets become available without restarting the worker. On the same first-read pass the data server applies the same preprocessing pipeline performed for Tabula pretraining. Every gene is scored by its per-dataset over-dispersion[80] (variance over mean) and ranked. The 1,200 most variable genes are selected as the model-input subset, and expression values within that subset are discretized into 50 bins per cell. The original .h5ad files stay unchanged. Applying the same protocol at every site means that every Trainer Application across the federated network operates on data prepared identically, ensuring quality consistency regardless of where the original raw dataset was uploaded or who prepared it. The data server additionally computes a 2-D UMAP[81] embedding of the binned cells-by-HVG matrix by fitting a streaming IncrementalPCA on its rows and projecting a deterministic random subsample of up to 50,000 cells through the fitted axes. The resulting embedding and a compact histogram of the over-dispersion score distribution are surfaced in the Chiron worker dashboard, allowing operators to inspect dataset structure and confirm that the platform's gene selection will retain the most biologically variable transcripts available to the model, all without raw data ever leaving the worker. To minimise the operational burden on data providers, the Chiron worker configuration interface ships an inline manifest.yaml form together with a single AI agent skill URL (https://chiron.aicell.io/skills/chiron-platform/SKILL.md) that documents the platform's data-preparation contract along with the worker launch and training workflows. Any compatible AI agent can use this skill to guide an operator through dataset preparation, worker launch, and training control without leaving the Chiron interface. Once a correctly formatted dataset is placed in the worker's data directory, the data server discovers it automatically and exposes it to the federation.

The training workflow is implemented as three cooperating BioEngine applications (a Manager, an Orchestrator, and a Trainer), each registered as a Hypha service on the worker node.

**Manager Application**. The Manager application is the control-plane entry point for a given BioEngine Worker. Deployed with zero CPU and GPU, it functions as a lightweight orchestration relay. It exposes three groups of endpoints: real-time reporting of cluster resources, geographic location, and available datasets; launching and stopping Orchestrator and Trainer applications via the BioEngine Worker API; and lifecycle tracking of all child applications through periodic status polling. State-reporting endpoints are publicly accessible to any authenticated BioEngine user, while launching or terminating applications requires admin access to the BioEngine Worker, enforced via BioEngine's access-control mechanism.

**Orchestrator Application.** Each Orchestrator application manages exactly one federated training session (1 CPU, 0 GPU, 8 GB RAM). It implements the server side of the Flower framework[34] using the standard FedAvg strategy[33]. Trainer nodes register by their Hypha service ID, and upon registration the orchestrator instantiates a client proxy that issues fit and evaluate calls over BioEngine's RPC layer, with the FedAvg strategy handling parameter aggregation. Running multiple concurrent sessions requires deploying independent orchestrator instances.

Each round, the FedAvg strategy sends the current global parameter vector and a fit configuration to all registered trainers. Each trainer returns its updated transformer weight tensors and its training sample count, and the orchestrator computes a sample-weighted average to form the new global model. Validation follows immediately using the same trainers, with losses and metrics aggregated by the same weighting scheme. A per-round timeout (default 300 s) bounds each fit/evaluate phase, and incomplete rounds are excluded from the training history. The orchestrator supports resumable training and, when pretrained weights are specified, downloads them from BioEngine's versioned artifact store and distributes them to all trainers before round 1.

**Trainer Application.** Each Trainer application implements the Flower client interface[34] and trains on a fixed local dataset partition (1 CPU, 1 GPU, 16 GB RAM). The dataset scope is fixed at deployment time, and multiple concurrent training sessions on the same worker require independent trainer instances.

During the fit phase, the trainer streams the pre-cut binned cells-by-HVG matrix from the co-located data server one row at a time over the container-internal authenticated channel, without ever materializing the raw expression matrix in the trainer container, and runs one epoch of the Tabula pretraining objective using PyTorch Lightning[82]. Per-batch progress is reported to the orchestrator by polling over BioEngine's RPC layer. Only the updated transformer weight tensors and sample count are returned, and tissue-specific embedder weights and projection head weights remain local at all times. During the evaluation phase, the trainer scores the current global parameters on its local validation split and returns loss and metrics. Training can be interrupted gracefully between batches without losing completed-round state.

## Human fibroblasts culturing

Human skin fibroblast samples (information below) were obtained from Baltimore Longitudinal Study of Aging (BLSA) and cultured at 37 °C in DMEM (Invitrogen) supplemented with non-essential amino acids, 10% tetracycline-free fetal bovine serum (TaKaRa Bio, 631106), and 1% penicillin-streptomycin (ThermoFisher Scientific, 15140122). Fibroblasts were expanded and used within 4 passages upon arrival.

| # | Coriell ID | Age at Biopsy | Gender/age gap | *passage* | *cPDL* |
|---|---|---|---|---|---|
| HF A1 | AG04054 | 29 | Male | 4 | 18.99 |
| HF A1.5 | AG11557 | 36 | Same person, 8 | 4 | 13.94 |
| HF A2 | AG13442 | 46 | Same person, 17 | *4* | *9* |
| | | | | | |
| HF B1 | AG08790 | 31 | Male | 4 | 18 |
| HF B2 | AG14245 | 48 | Same person, 17 | 4 | 5 |
| | | | | | |
| HF H1 | AG06234 | 17 | Male | *4* | *6* |
| HF I 2 | AG13151 | 54 | Male, father of H1, 37 | *5* | *6* |

**Lentiviral infection.** Doxycycline-inducible lentiviral constructs were generated by Gateway LR recombination between pDonor-ORF and a home-made pLIX_403_mCherry destination vector. Young and aged fibroblasts were plated on day 0 and transduced on day 1 with lentivirus in the presence of 8 μg/mL polybrene. On day 2, the medium was replaced with fresh medium lacking polybrene but containing 2 μg/mL doxycycline, and cells were maintained under these conditions through day 6. On day 7, the medium was replaced with a fresh medium containing 0.2 μg/mL doxycycline. Cells were then harvested by trypsinization on day 8. Single-cell RNA-seq was performed using the Chromium Next GEM Single Cell 5′ Reagent Kits v2.

**Perturb-seq vector design and perturbation capture strategy.** Because the pooled TF library lacks unique 3′ barcodes, individual overexpressed TFs could not be distinguished using standard 3′ single-cell RNA-seq. In addition, all transgene-derived mRNAs share a common rtTA sequence at the 3′ end. We therefore used the 10x Genomics 5′ capture platform to sequence into the variable TF coding region from the 5′ end of each transcript. To selectively enrich transcripts derived from the TF library, targeted PCR amplification was performed using one primer against the shared V5 tag and a second primer against the conserved sequence introduced during 10x library preparation from the Illumina gel bead oligonucleotides. This approach enabled identification of the exogenous TF expressed in each transduced cell.

## Implementation details

Tabula was implemented in Python and utilized PyTorch Lightning to construct its overall framework. Within the federated learning setup, message passing interface (MPI) was employed in the cluster to allocate computing nodes to each client. The NCCL framework [83] was adopted for efficient communication among clients over the InfiniBand (IB) network. Additionally, *torch.distributed* was implemented to realize the message queues of the parameter sending, receiving, broadcasting and managing process.

- Tabula
  Tabula employs FlashAttention-2 [65] architecture, comprising three transformer blocks each containing 8 attention heads, with an output embedding size of 192. The maximum input gene length was set to 1,200, corresponding to 1,200 highly variable genes (HVGs) selected from each dataset. The pretraining dataset consisted of 15 million cells, which were randomly sampled from tissue-specific subsets within the CELLxGENE whole-human dataset. During the pretraining stage, an AdamW optimizer [84] was utilized with a 0.0001 initial learning rate, which was decreased by a factor of 0.95 after each epoch. The model was trained for 8 epochs with a batch size of 32. For the tissue-specific client models, 90% of the data was used for training and the remaining 10% for validation. The corruption rate was set to 0.6 and the temperature coefficient of $\mathcal{L}_{\text{contrast}}$ to 0.07. To achieve a relative balance between the two learning objectives, the weight of $\mathcal{L}_{\text{contrast}}$ and $\mathcal{L}_{\text{rec}}$ was set to 1:0.03. For the federated learning settings, local tabular transformer weight aggregation and update were performed at the end of every training epoch for all clients. Since all clients are considered equally important, the global tabular transformer weight was averaged across all clients during aggregation. After pretraining, we extract the client model weights following 8 epochs as the final tissue-specific embedders and projection heads weights, and we use the aggregated weight of the transformer as the final global tabular transformer weight.

- Decentralized training platform
  We package the BioEngine Worker with preinstalled dependencies for the Manager, Orchestrator and Trainer applications as a Docker image (ghcr.io/aicell-lab/tabula) deployable via Docker Compose or Apptainer. Selected runtime dependencies for the trainer include: torch v1.13.1, pytorch-lightning v2.2.0, flwr v1.22.0, scikit-learn v1.3.0, zarr v3.1.3 (used to stream single-cell datasets stored in Zarr v3 format from the co-located Data Server), and numpy v1.26.0. FlashAttention-2 (flash-attn v2.3.5) can be enabled via an environment variable for compatible CUDA devices. BioEngine Workers operate under Python 3.11 with Ray v2.55.1. All Hypha RPC communication is provided by the hypha-rpc package (v0.21.40). Control-plane interactions between the Manager application and the BioEngine Worker service use the WebSocket transport, while interactions between the orchestrator and trainer nodes during federated training use WebRTC-based peer-to-peer channels.

- Benchmark federated vs. centralized pretraining
  We benchmarked two pretraining schemes: federated learning and centralized learning. The federated learning scheme followed the same training methodology in Tabula. In contrast, the centralized learning scheme adopted a traditional approach, where all data was consolidated, and model pretraining was performed on a single centralized device using all acquired data. This approach avoids maintaining multiple models for individual clients and performing model weight aggregation and updates across clients.

- Benchmark tabular learning vs. MLM
  We benchmarked two single-cell modeling strategies: tabular learning and masked language modeling (MLM). The tabular learning strategy was using the same learning objectives in Tabula, while the MLM strategy aimed to reconstruct these masked expression values through contextual inference. Specifically, 15% of the non-zero gene expression values in the input sequence were randomly masked by setting their expression levels to -1, following scBERT [23]. The mean squared error (MSE) loss is computed exclusively for the masked positions.

- Cell type annotation
  For the cell type annotation task, we adopted an initial learning rate of 0.0001 and a weight decay of 0.1. The batch size was set to 32, while finetuning was performed with an early stopping strategy. For the cell line dataset, we randomly splitted 33% of the data as the test set, with the remaining 67% used for training where 10% was randomly chosen as the validation set. In other datasets, the training and test sets were pre-defined (see **Downstream Task Datasets**), and 10% of the training set was similarly reserved as the validation set. The optimal weights were selected based on the lowest loss on the validation set and used for evaluating model's performance on the test set. The data preprocessing steps included transcripts-per-million normalization, log1p transformation, and value binning. In addition, we selected 3,000 highly variable genes and prioritized the genes with non-zero expression in each cell during data loading until the maximum length requirement of 1,200 genes was reached. If fewer than 1,200 genes were selected, genes with zero expression will be filled. This approach ensures that the model senses more genes during the finetuning process while meeting the maximum length requirement.

- Multi-omics and multi-batch integration
  For multi-omics and multi-batch integration tasks, we selected 2,000 HVGs from the dataset and randomly chose 1,200 genes, which is the maximum input length of the pretrained Tabula, for each training epoch. The data preprocess steps included TPM normalization, log1p transformation, HVG selection and value binning. During finetuning, we used an initial learning rate of 0.0001 and a weight decay of 0.95. For datasets with multi-omics, the model was extended to accommodate new tokens for new omics modality similar to gene embeddings, the mask ratio of $\mathcal{L}_{\mathrm{MGM}}$ and $\mathcal{L}_{\mathrm{CMGM}}$ was set to 0.4, and the corruption rate of $\mathcal{L}_{\mathrm{reconstruction}}$ was set to 0.6. For datasets with multi-batch, $\mathcal{L}_{\mathrm{DA}}$ was adopted with the weight of 1. The four losses were applied to both tasks if the dataset has multiple batches and multiple omics. Gradient clipping with a value of 0.5 was enabled to avoid exploding gradients. The model was trained with an early stopping strategy. The best performance of each dataset is reported based on the best-combined validation loss.

- Genetic perturbation prediction
  For the gene perturbation prediction task, in order to reconstruct all genes to enable comprehensive understanding of inter-gene relationship, for the datasets exceeding the maximum input length of 1,200 genes, we randomly initialized the parts that could not be accommodated by the pretrained model weights. The model optimization adopted a learning rate of 0.0001 and a weight decay of 0.95. The batch size was set to 64, while the training was conducted with an early stopping strategy with patience of 5 epochs.

- Imputation
  For the imputation task, the initial learning rate was set to 0.0001 with a weight decay of 0.95. The batch size was set to 32, and fine-tuning was performed over a fixed 10 epochs. The maximum sequence length was set to 2,000. We masked the data to simulate dropout values. This process masked non-zero gene expression data for each cell, splitting it into training, validation, and test sets. For each cell, non-zero values were identified. If there were fewer than 5 non-zero values, no masking was applied. Otherwise, based on an exponential probability distribution (scale 20), 10% of the data was masked for validation and another 10% for testing. The masked values were set to 0, and the remaining data formed the training set. The data preprocessing steps included TPM normalization and log1p transformation. We only conducted a value binning preprocess for the gene expressions in the training and validation sets, and the test sets were used with log1p values. The model weights exhibiting the lowest loss on the validation set were selected as the optimal weights for evaluating performance on the test set.

- In silico gene regulatory network inference

For the task of in silico gene regulatory network inference, tissue-specific embedders and projection heads weights were applied across various biological systems. These included using blood-specific weight for hematopoiesis, pancreas-specific weight for pancreatic endogenesis, brain-specific weight for neurogenesis, and heart-specific weight for cardiogenesis. The data preprocess steps included TPM normalization, log1p transformation, HVG selection and value binning before the model inference. To capture the gene-gene relationships inherent in regulatory networks, an input length of 1,200 is insufficient. Consequently, we initialized Tabula with an extended input length and loaded pretrained weights that have compatible dimensions, specifically: 1,900 for pancreatic endogenesis dataset, 1,775 for hematopoiesis dataset, and 2,400 for both neurogenesis and cardiogenesis datasets.

## Downstream Task Datasets

- CELLxGENE
  The CELLxGENE dataset is a publicly accessible resource developed and maintained by the Chan Zuckerberg Initiative, designed for storing, exploring, and analyzing scRNA-seq data. It aggregates data from research projects across the globe. Through the CELLxGENE portal, we collect scRNA-seq data from approximately 40 million human cells, and the release version from July 25, 2023. After quality control, preprocessing and filtering, about 15 million human scRNA-seq data are obtained for Tabula pretraining. This comprehensive dataset includes a diverse range of 23,156 genes and 422 cell types from 133 tissues and 196 studies.

- PBMC 5K
  We use a high-throughput scRNA-seq dataset PBMC 5K from 10x Genomics containing samples of peripheral blood mononuclear cells (PBMCs) from healthy donors. These cells are tagged with a panel of TotalSeq-B antibodies (v3 chemistry). The dataset comprises a total of 5,247 cells and includes 33,538 genes.

- Jurkat
  The Jurkat cells are human T-lymphocyte cell lines derived from patients with acute T-cell leukemia and are widely used in immunology research. This dataset provides scRNA-seq data for Jurkat cells generated by the 10x Genomics. The dataset comprises a total of 3,258 cells and includes 32,738 genes.

- Melanoma
  The Melanoma dataset centres on scRNA-seq of WM989-A6-G3 melanoma cells using two technologies: DropSeq and Fluidigm's C1 mRNA Seq HT chip[85]. These methods allow a detailed analysis of the transcriptomic characteristics of melanoma cells. We select data sequenced using DropSeq technology. The dataset comprises a total of 8,640 cells and includes 32,287 genes.

- Adamson

  The Adamson perturbation is a scRNA-seq dataset derived from K562 human cell lines [53], designed to study the effects of CRISPR interference on gene expression. The dataset contains RNA expression profiles for 65,337 raw cells, with 43,550 cells retained after quality filtering. It includes measurements for 32,738 genes across 87 genetic perturbations, following filtering from 90 initial perturbations. The dataset is preprocessed by GEARS data processing function [76] for further finetuning.

- Norman

  The Norman perturbation dataset is also a scRNA-seq dataset generated from K562 human cell lines [86], designed to explore genetic interactions through CRISPR activation. The dataset captures RNA expression profiles from 111,668 raw cells, with 82,081 cells retained after quality filtering. It contains measurements for 33,694 genes across 105 single-gene perturbations and 131 dual-gene perturbations. The dataset is preprocessed by GEARS data processing function [76] for further finetuning.

- Replogle

  The Replogle perturbation dataset uses a multiplexed CRISPR interference with single-cell transcriptome sequencing to analyze the effects of gene knockdowns across cell types, creating a comprehensive human cell genotype-phenotype atlas [87]. The dataset has been meticulously processed based on the curated protocols of scGPT [17], which includes a total of 171,542 samples, derived from 1,823 distinct one-gene perturbations, of which 99 target transcription factors. Additionally, the dataset features a test set consisting of 456 perturbations, with 25 specifically affecting transcription factors. The dataset is preprocessed by GEARS data processing function [76] for further finetuning.

- Human Pancreas

  The Human pancreas dataset comprises data from five scRNA-seq studies on human pancreas cells, reprocessed by [88] for cell type annotation. In total, there are 14 cell types including alpha, beta, ductal, acinar, delta, pancreatic stellate, pancreatic polypeptide, endothelial, macrophage, mast, epsilon, Schwann, T cell and MHC class III. The reference set includes data from two sources containing 10,600 cells across 13 cell types, while the query data covers the remaining three sources consisting of 4,218 cells spanning 11 cell types. The query set includes MHC class II, absent in the reference set, but lacks macrophage, Schwann, pancreatic polypeptide and T cell, which are present in the reference set.

- Myeloid

  The Myeloid dataset provides a comprehensive analysis of subpopulation profiling of tumour-infiltrating myeloid cells at the pan-cancer level, by integrating single cell transcriptomic data from nine different cancer types [89]. The dataset was randomly subsampled by [17]. The reference set contains 9,748 cells from six cancer types including UCEC, PAAD, THCA, LYM, cDC2 and kidney, while the query set contains 3,430 cells from four cancer types including MYE, OV-FTC and ESCA.

- 293t & Jurkat cells
  The 293t & Jurkat cells dataset, generated using the 10x Genomics platform, profiles gene expression across 16,602 genes in two distinct cell lines: 293T cells, originating from human embryonic kidney and commonly used in gene expression and viral studies, and Jurkat cells, a T lymphocyte line derived from a leukemia patient and frequently employed in T cell signaling research. As reported by [90], the dataset comprises three experimental batches: Batch 1 consisting of 2,885 293T cells, Batch 2 containing 3,258 Jurkat cells, and Batch 3 featuring a balanced mixture of both cell types with 3,388 cells total distributed equally between 293T and Jurkat cell lines.

- Liver
  The Liver dataset contains a variety of human liver cell types from two primary sources. The reference dataset, processed by[91], contains 8,444 parenchymal and non-parenchymal cells from five human livers. It includes 14 different cell types, including central venous sinusoidal endothelial cells, erythroid cells, hepatocytes, inflammatory and non-inflammatory macrophages, mature B cells, natural killer (NK) cells, periportal sinusoidal endothelial cells, plasma cells, portal endothelial cells, stellate cells, alpha-beta T cells, cholangiocytes and gamma-delta T cells. While the query dataset provided by[92]. consists of 9,162 cells from normal human liver tissue, it overlaps with 7 of the cell types found in the reference dataset.

- 10x Multiome PBMC
  The 10x Multiome PBMC dataset captures dual-modality single cell profiles from human peripheral blood mononuclear cells through paired RNA and ATAC sequencing analysis. Processed by [93], this dataset contains 9,631 cells derived from a healthy 25-year-old female donor, with granulocytes removed via cell sorting before nuclear isolation and sequencing. The data spans 29,095 genes for expression analysis and 107,194 regions for chromatin accessibility measurements. The cells are classified into 19 distinct immune populations, encompassing various T cell subtypes (CD4+ naive, CD4+ TCM, CD4+ TEM, CD8+ naive, CD8+ TEM 1, CD8+ TEM 2, MAIT, Treg, gdT), B cell subtypes (naive, intermediate, memory, plasma), myeloid cells (CD14+ monocytes, CD16+ monocytes, cDC, pDC), as well as NK cells and HSPCs.

- BMMC
  The BMMC dataset is a comprehensive multimodal single-cell profiling of bone marrow mononuclear cells from 12 healthy human donors [94], originally designed for the Multimodal Single-Cell Data Integration Challenge at NeurIPS 2021. It includes paired single-cell RNA and protein abundance measurements across 12 batches. Processed by [17], the final dataset consists of 90,261 cells, with data on 13,953 genes and 134 surface proteins, along with detailed annotations for 45 immune cell subtypes.

- Perirhinal Cortex

The Perirhinal Cortex dataset is a subset of a larger study by [95], which created a comprehensive transcriptomic atlas of the human brain using over three million nuclei from around 100 dissections across various brain regions. From this dataset, two separate batches from the perirhinal cortex were selected for further analysis, following the methodology outlined by scGPT [17]. The first batch contains 8,465 cells, while the second batch includes 9,070 cells. Together, the datasets comprise a total of 59,357 genes, with the ten unique cell types identified in the original study being used for analysis.

- DC
  This DC dataset contains scRNA-seq data from human blood dendritic cells (DCs), originally collected by [96] and processed by [90]. It includes two batches, each with four distinct cell types: CD1C DC, CD141 DC, plasmacytoid DC (pDC), and double negative cells. To ensure non-identical cell compositions across batches, CD1C DCs are excluded from batch 1, and CD141 DCs are excluded from batch 2. Each batch contains 288 cells and 16,594 genes, with 96 pDCs and 96 double negative cells shared between the batches. Batch 1 includes 96 CD141 cells, while batch 2 includes 96 CD1C cells.

- Hematopoiesis
  The Hematopoiesis dataset is derived from [1]. The dataset consists of 1,947 cells representing 8 cell types across five developmental nodes. These cell types include: hematopoietic stem cells (HSC), megakaryocyte-erythrocyte progenitors (MEP), granulocyte-monocyte progenitors (GMP), monocytes (MON), neutrophils (Neu), megakaryocytes (Meg), erythrocytes (Ery), and basophils (Bas). Based on the known cell differentiation[35], HSC first differentiates into MEP and GMP, with GMP further developing into Mon and Neu, while MEP diverts into Meg, Ery, and Bas.

- Cardiogenesis
  The Cardiogenesis dataset is obtained from the cardiogenic region of mouse embryos [36]. The cardiogenic mesoderm differentiates from a shared population of cardiovascular progenitor cells into two regions with distinct gene expression: the first heart field (FHF) and the second heart field (SHF). We focus on the gene regulatory relationships within the FHF and SHF. A total of 1,165 cells are selected from this dataset for analysis, comprising cells from both the posterior second heart field (pSHF) and FHF.

- Neurogenesis
  The Neurogenesis dataset is a comprehensive single cell transcriptomic atlas of the embryonic mouse brain between gastrulation and birth [38]. We focus on the transformation of glial cells into astrocytic and oligodendrocytic lineages within the central nervous system. We select a total of 4,704 cells from the atlas, which include 2,516 mixed region astrocytes, 1,990 oligodendrocyte precursor cells, 162 committed oligodendrocyte precursor cells, and 36 mature oligodendrocytes.

- Pancreatic Endogenous

The Pancreatic Endogenous dataset is based on the work of [41]. We aim to investigate the master transcription factors that govern pancreatic cell fate during the development of the pancreas. A total of 14,445 cells are selected from the dataset across three time points: 4,631 cells at E12, 5,179 cells at E14, and 4,635 cells at E17.

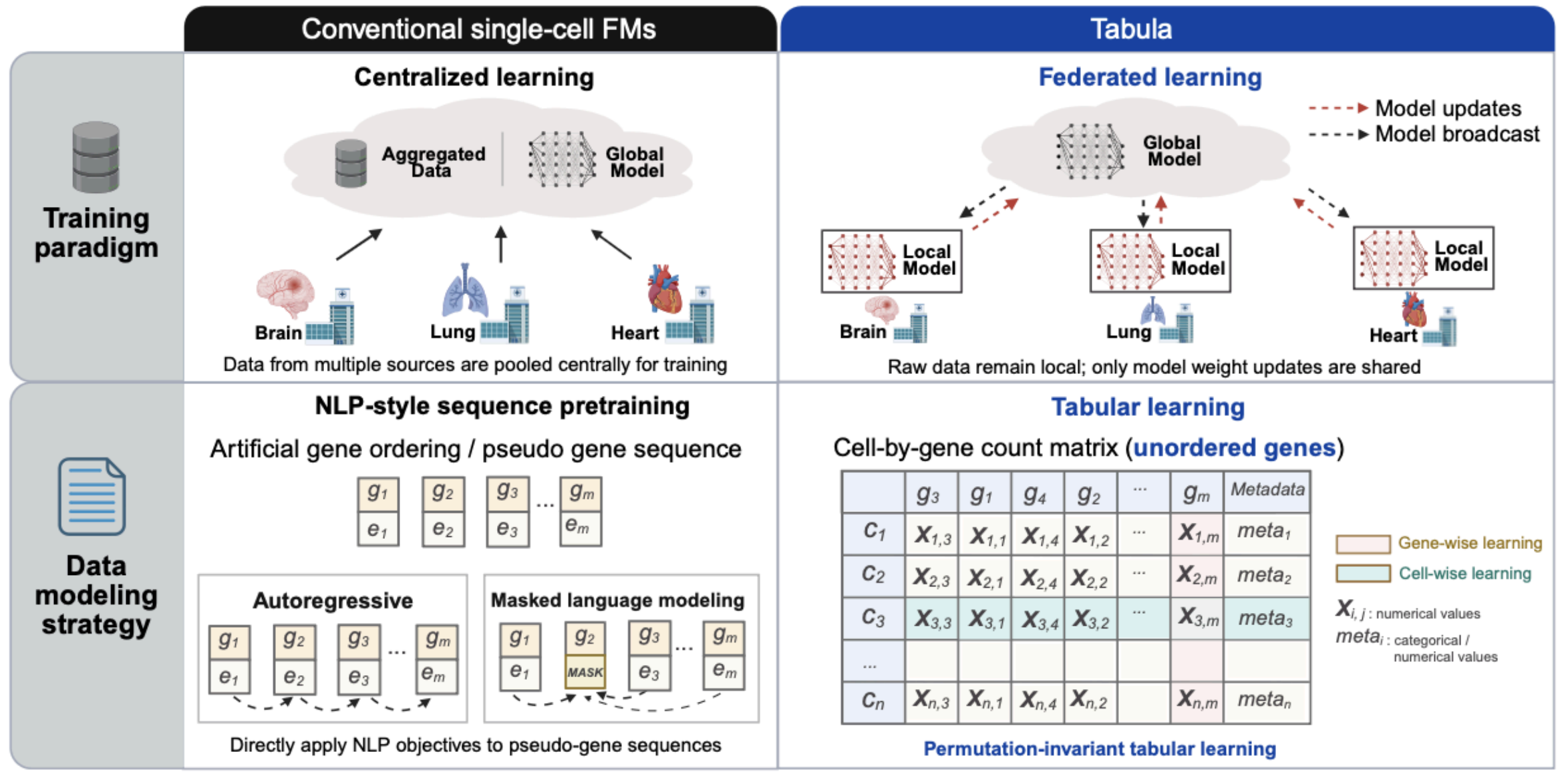
a
Conventional Single-Cell Foundation Models vs. Tabula
Conventional single-cell FMs
Tabula
Training paradigm
Centralized learning
Aggregated Data
Global Model
Brain
Lung
Heart
Data from multiple sources are pooled centrally for training
Federated learning
Model updates
Model broadcast
Global Model
Local Model
Raw data remain local; only model weight updates are shared
Data modeling strategy
NLP-style sequence pretraining
Artificial gene ordering / pseudo gene sequence
Autoregressive
Masked language modeling
MASK
Directly apply NLP objectives to pseudo-gene sequences
Tabular learning
Cell-by-gene count matrix (unordered genes)
Metadata
Gene-wise learning
Cell-wise learning
$X_{i,j}$: numerical values
$meta_i$: categorical / numerical values
Permutation-invariant tabular learning

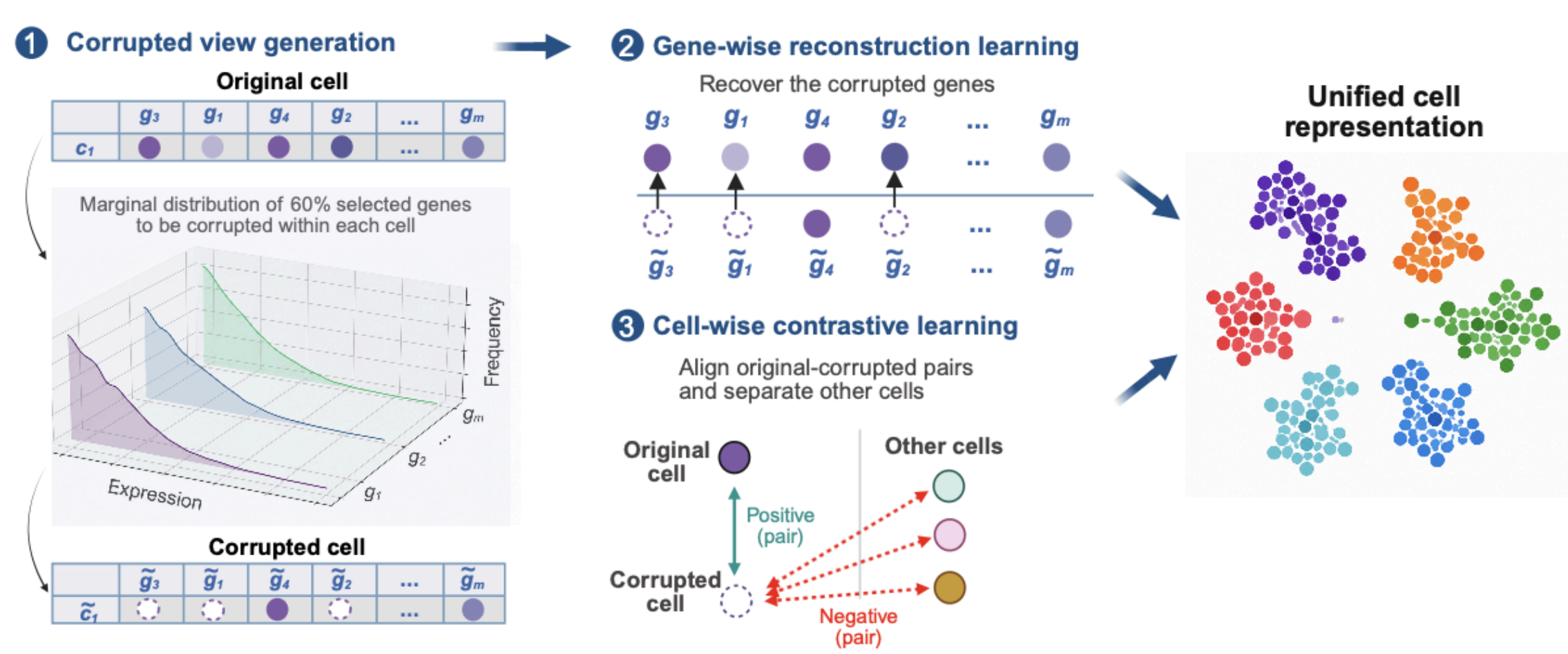
b
Self-Supervised Tabular Learning
1 Corrupted view generation
Original cell
Marginal distribution of 60% selected genes to be corrupted within each cell
Frequency
Expression
Corrupted cell
2 Gene-wise reconstruction learning
Recover the corrupted genes
3 Cell-wise contrastive learning
Align original-corrupted pairs and separate other cells
Original cell
Other cells
Positive (pair)
Corrupted cell
Negative (pair)
Unified cell representation

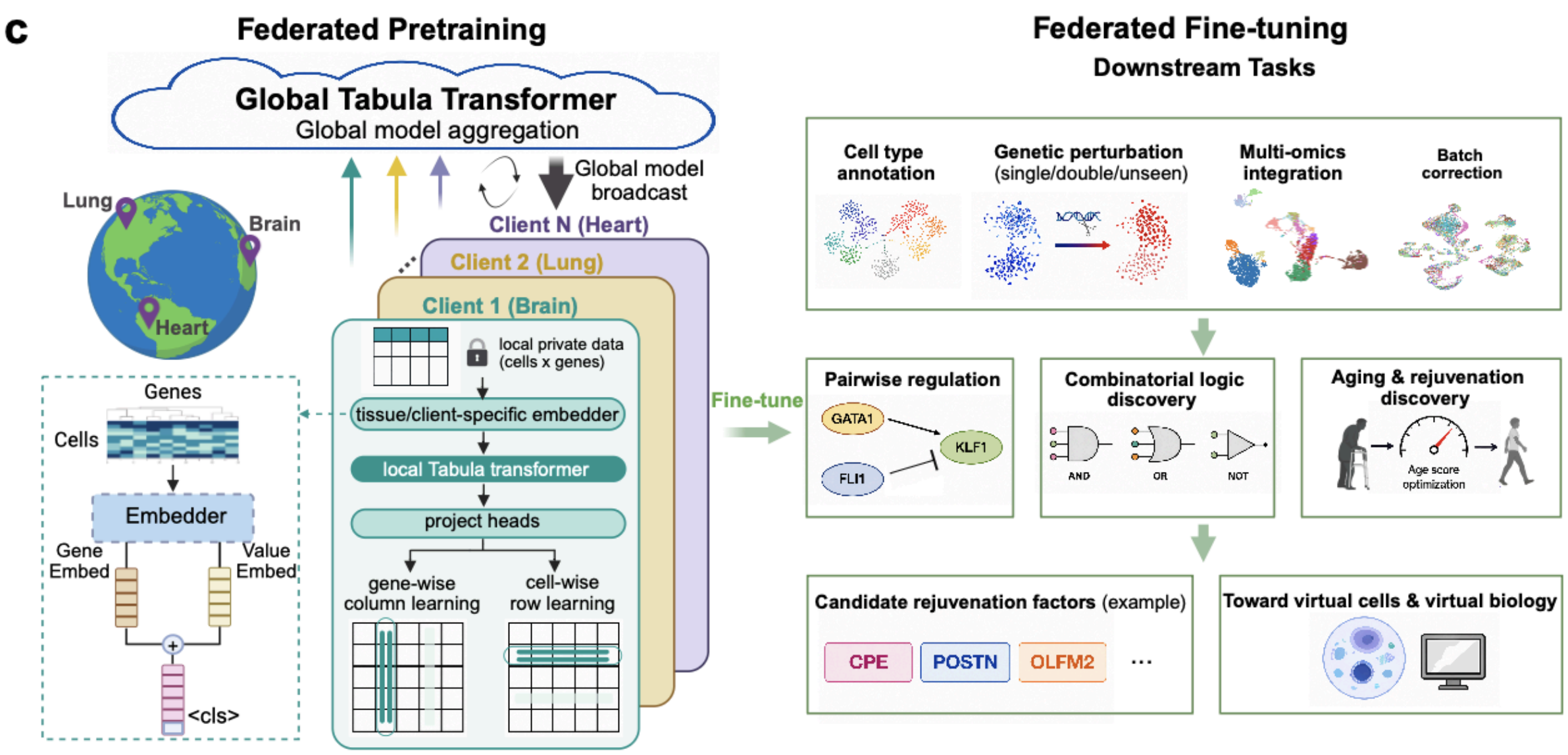
c
Federated Pretraining
Global Tabula Transformer
Global model aggregation
Global model broadcast
Lung
Brain
Heart
Client N (Heart)
Client 2 (Lung)
Client 1 (Brain)
local private data (cells x genes)
tissue/client-specific embedder
local Tabula transformer
project heads
gene-wise column learning
cell-wise row learning
Genes
Cells
Embedder
Gene Embed
Value Embed
<cls>
Fine-tune
Federated Fine-tuning
Downstream Tasks
Cell type annotation
Genetic perturbation (single/double/unseen)
Multi-omics integration
Batch correction
Pairwise regulation
GATA1
KLF1
FLI1
Combinatorial logic discovery
AND
OR
NOT
Aging & rejuvenation discovery
Age score optimization
Candidate rejuvenation factors (example)
CPE
POSTN
OLFM2
Toward virtual cells & virtual biology

## Figure 1: The schematic of Tabula, a novel single cell foundation model that integrates tabular learning and federated learning.

a. **Key conceptual innovations of Tabula over existing single cell foundation models.** Existing single-cell foundation models differ from Tabula along two key dimensions. In terms of training paradigm, existing models such as Geneformer[14] and scGPT[17] rely on centralized learning where data from multiple sources are pooled centrally for training, whereas Tabula employs federated learning[22] where raw data remain local and only model weight updates are shared. In terms of data modeling strategy, existing models treat gene expression within each cell as artificially ordered gene sequences and apply NLP objectives such as autoregressive or masked language modeling, despite the fact that gene expression has no natural order. Instead, Tabula explicitly considers the tabular structure of scRNA-seq data and uses tabular learning to build the foundation model. In Tabula, each cell in the gene expression matrix is represented by a permutation-invariant row of genes (each gene corresponds to a column) containing numerical expression values or metadata categories, respecting the unordered structure of gene features.
b. **Tabula relies on self-supervised tabular modeling of corrupted expression data along both gene and cell dimensions**. Tabula's self-supervised tabular learning proceeds in three steps. First, a corrupted view of the scRNA-seq data is generated by resampling 60% of genes within each cell and replacing their expression values based on each gene's individual marginal distribution. For instance, $g_3, g_1, g_2$ are randomly selected and replaced with the samples values $\tilde{g}_3, \tilde{g}_1, \tilde{g}_2$ from their corresponding marginal distributions in cell $c_1$. Second, gene-wise reconstruction learning recovers the original gene expression from its corrupted view, ensuring robust gene-level representations. Third, cell-wise contrastive learning aligns positive cell pairs (the original and corrupted views of the same cell) while separating negative pairs (cells from distinct origins) in the latent space, ensuring both intra-cellular consistency and inter-cell variability. Together, these two objectives capture the structure of single-cell data along both the gene and cell axes, yielding a unified cell representation.
c. **Model architecture, federated training scheme, and downstream applications of Tabula**. **Left**: Tabula is built on a shared global transformer architecture, while each client incorporates tissue- or client-specific embedders and projection heads to capture client-dependent characteristics. During federated pretraining, each client updates its local tabular transformer using its private dataset for one epoch, computing both gene-wise column learning loss and cell-wise row learning loss. The updated transformer weights are then transmitted to a central orchestrator for global model aggregation, without sharing raw data. The aggregated transformer weights are redistributed to all clients for the next training round, and this iterative process continues until convergence. **Right**: Following pretraining, the global Tabula transformer can be fine-tuned for a broad suite of downstream tasks. Classical tasks include cell type annotation, genetic perturbation prediction (single, double, and unseen perturbations), multi-omics integration, and batch correction. Beyond these, Tabula enables novel biological discovery tasks including pairwise regulation inference, combinatorial logic discovery, and aging and rejuvenation discovery, exemplified by the nomination of candidate rejuvenation factors such as *CPE*, *POSTN*, and *OLFM2*, paving the way toward virtual cells and virtual biology.

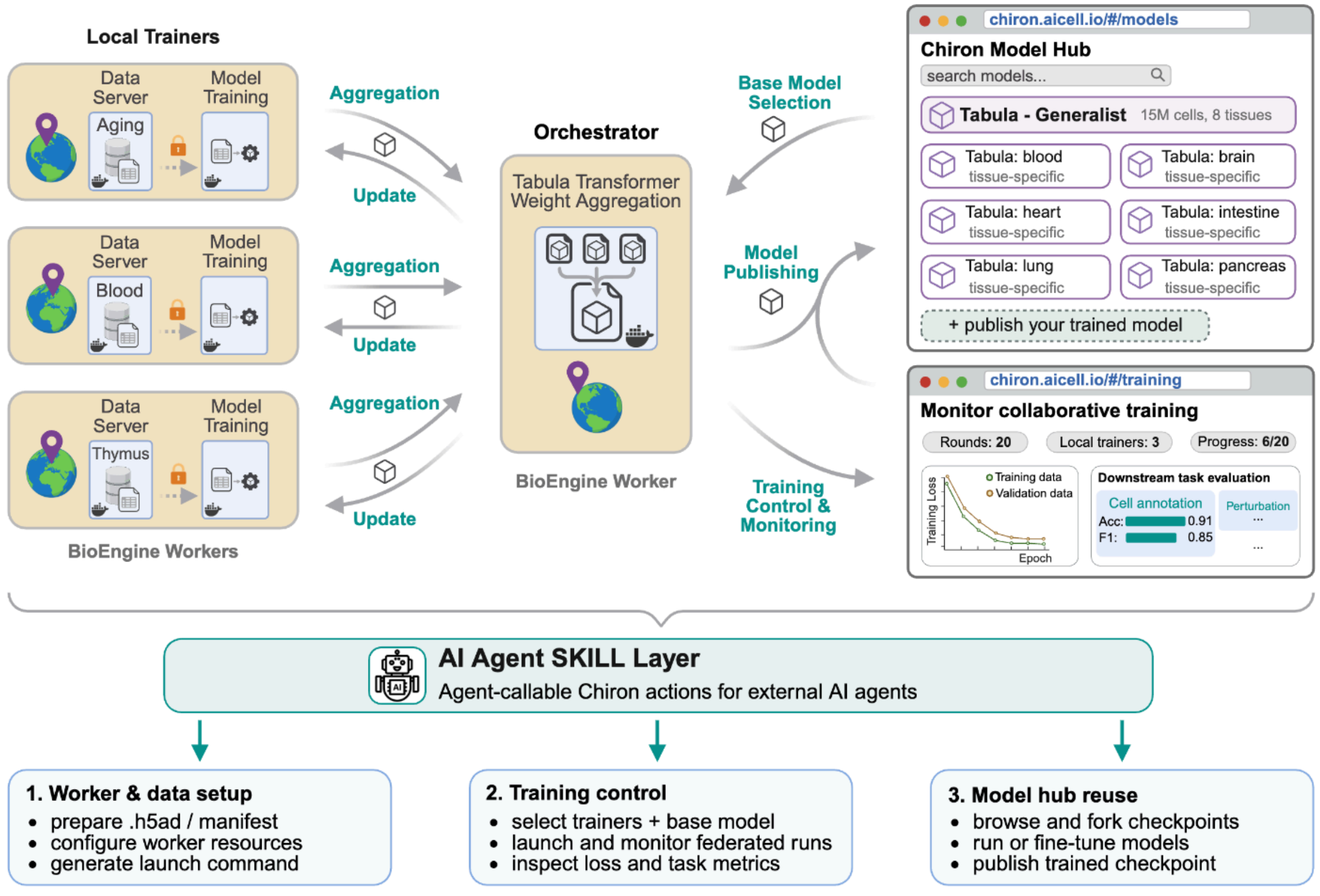
a
Chiron: A decentralized platform for collaborative single-cell foundation models
Local Trainers
Data Server
Model Training
Aging
Blood
Thymus
Aggregation
Update
Orchestrator
Tabula Transformer Weight Aggregation
BioEngine Worker
BioEngine Workers
Base Model Selection
Model Publishing
Training Control & Monitoring
chiron.aicell.io/#/models
Chiron Model Hub
search models...
Tabula - Generalist 15M cells, 8 tissues
Tabula: blood tissue-specific
Tabula: brain tissue-specific
Tabula: heart tissue-specific
Tabula: intestine tissue-specific
Tabula: lung tissue-specific
Tabula: pancreas tissue-specific
+ publish your trained model
chiron.aicell.io/#/training
Monitor collaborative training
Rounds: 20
Local trainers: 3
Progress: 6/20
Training data
Validation data
Training Loss
Epoch
Downstream task evaluation
Cell annotation
Perturbation
Acc: 0.91
F1: 0.85
AI Agent SKILL Layer
Agent-callable Chiron actions for external AI agents
1. Worker & data setup
prepare .h5ad / manifest
configure worker resources
generate launch command
2. Training control
select trainers + base model
launch and monitor federated runs
inspect loss and task metrics
3. Model hub reuse
browse and fork checkpoints
run or fine-tune models
publish trained checkpoint

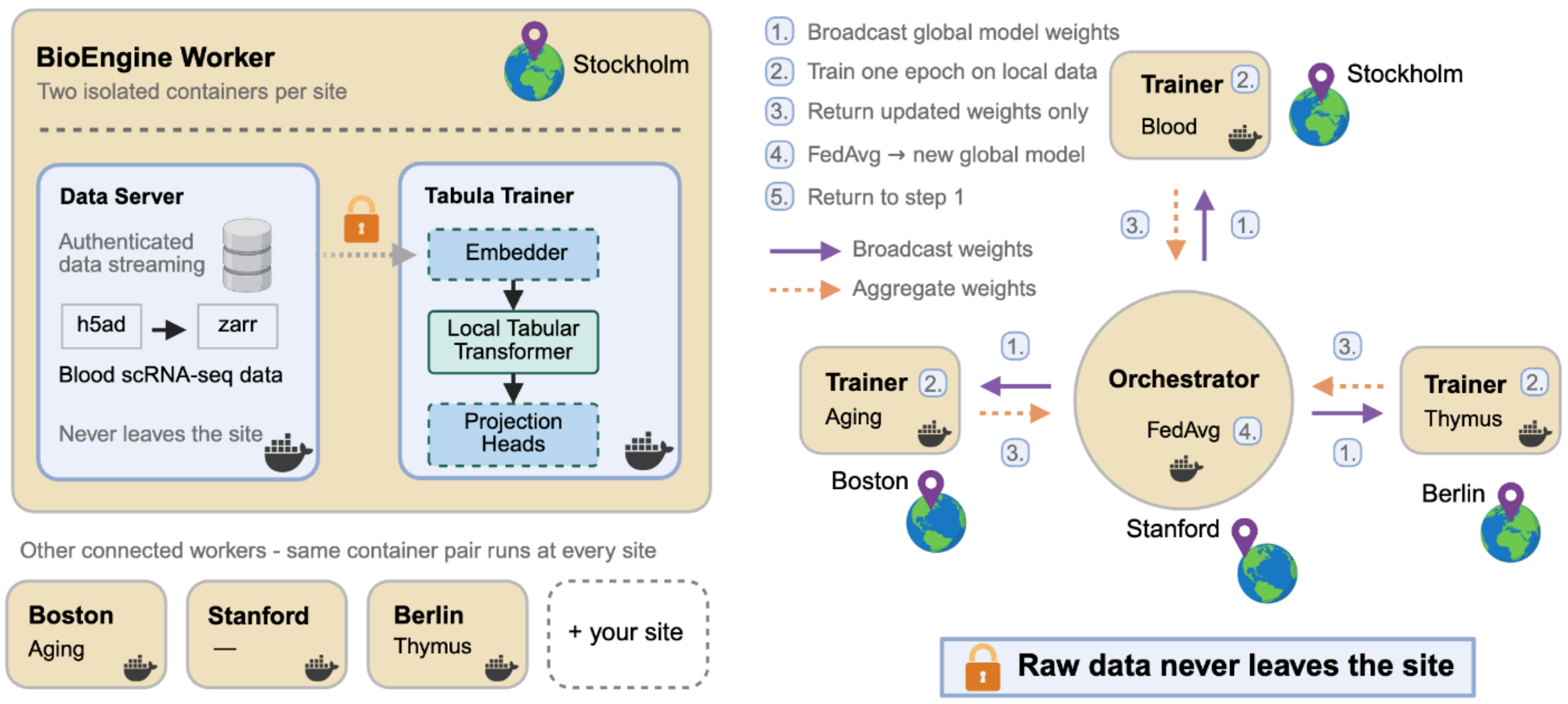
b BioEngine Worker: what runs at each site
BioEngine Worker
Two isolated containers per site
Stockholm
Data Server
Authenticated data streaming
h5ad
zarr
Blood scRNA-seq data
Never leaves the site
Tabula Trainer
Embedder
Local Tabular Transformer
Projection Heads
Other connected workers - same container pair runs at every site
Boston
Aging
Stanford
—
Berlin
Thymus
+ your site
c Chiron Orchestrator: one federated round
1. Broadcast global model weights
2. Train one epoch on local data
3. Return updated weights only
4. FedAvg → new global model
5. Return to step 1
Broadcast weights
Aggregate weights
Trainer
Blood
Stockholm
Aging
Boston
Orchestrator
FedAvg
Stanford
Thymus
Berlin
Raw data never leaves the site

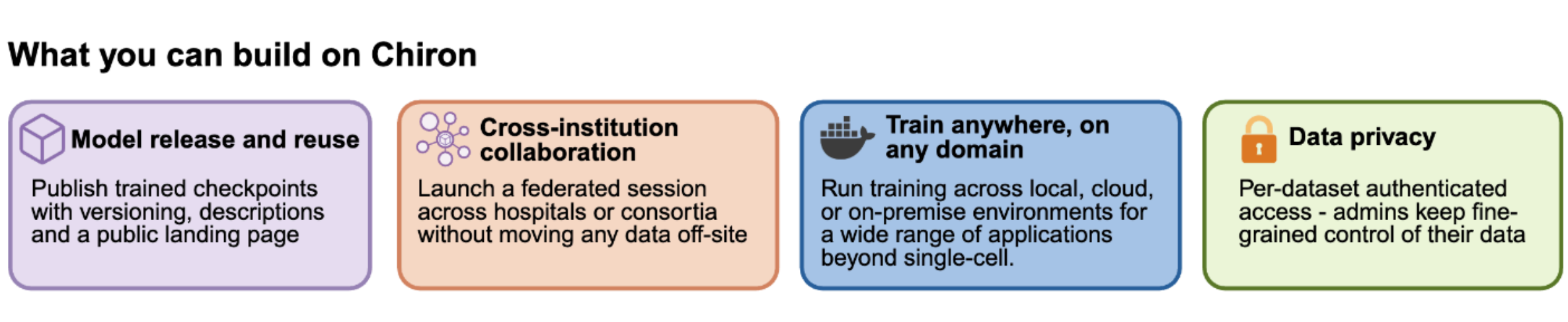
d What you can build on Chiron
Model release and reuse
Publish trained checkpoints with versioning, descriptions and a public landing page
Cross-institution collaboration
Launch a federated session across hospitals or consortia without moving any data off-site
Train anywhere, on any domain
Run training across local, cloud, or on-premise environments for a wide range of applications beyond single-cell.
Data privacy
Per-dataset authenticated access - admins keep fine-grained control of their data

## Figure 2: Chiron, a decentralized platform for collaborative Tabula and other single-cell foundation models.

a. **End-to-end workflow of a federated training session on Chiron.** Three illustrative BioEngine workers[29] act as geographically distributed local trainers (**left**), each pairing a local data server that connects the institution's private single-cell dataset with a model-training container that updates the tabular model for one local epoch per round. After every round, each trainer sends only the updated transformer weights to the central orchestrator, which runs a Tabula transformer weight aggregation step on its own BioEngine worker and broadcasts the new global weights back to every trainer for the next round. The orchestrator is wired into two browser-based interfaces in the Chiron platform. The Chiron model hub (**top right**) is the entry point for base model selection at the start of a session and for model publishing of the trained checkpoint at the end. The hub catalogues the centralized Tabula Generalist (15 million cells across eight tissues) alongside tissue-specific Tabula variants released by previous federated sessions, and any session participant can republish their own trained model back to the model hub. The training control and monitoring interface (**bottom right**) lets the operator launch, watch, and steer the live session through per-round training and validation loss curves and downstream-task evaluation metrics. A built-in AI Agent skill (**bottom**) exposes the same Chiron actions in natural language, with three entry points covering the workflow above: set up your worker and data, launch and monitor a training run, and browse and reuse published models. This system is demonstrated for the Tabula model but applicable to many single cell foundation models.
b. **The configuration of a BioEngine Worker of each participating institute.** The data server (left) ingests the institution's local single-cell data, converts AnnData .h5ad files to a streaming-friendly zarr representation upon its first access while leaving the originals on disk untouched, and exposes the resulting zarr stores to the co-located trainer over the container-internal network via authenticated streaming. Raw single-cell data never leaves the site. The Tabula Trainer (right) holds the GPU-bound replicas of the tissue-specific embedder, the local tabular transformer, and the projection heads, and exchanges only transformer weights with the central orchestrator at the end of each round. One worker is shown as the foreground example, running a trainer on local blood scRNA-seq data. The same container pair runs at every participating institution, so adding a new site is a configuration step rather than an infrastructure project.
c. **The cartoon of steps involved in a federated round on the Chiron orchestrator.** The orchestrator broadcasts the current global transformer weights to every registered trainer (purple arrows, step 1). Each trainer trains for one local epoch on its private dataset (step 2) and returns only the updated transformer weights to the orchestrator (orange arrows, step 3). The orchestrator then performs a FedAvg aggregation across the contributions to produce a new global model (step 4) and re-enters step 1 for the next round. The orchestrator instance can run on any registered worker regardless of whether that worker also hosts local data. In the illustration it runs on a worker that does not contribute training data. Across the entire run, the orchestrator only ever sees transformer weights and scalar metrics, and raw single-cell data never leaves the site that owns it.

d. **What can be built on Chiron?** The four cards summarize the use patterns the platform supports. (1) Model release and reuse: packages trained checkpoints with versioning, descriptions and a public landing page so any reader can pull and run the model behind a published study. (2) Cross-institution collaboration: launches a federated session across hospitals or consortia without moving any data off site. (3) Train anywhere, on any domain: runs the same container pair in local, cloud, or on-premise environments so the platform can support a wide range of applications beyond the Tabula single-cell foundation model. (4) Data privacy: imposes per-dataset authenticated access and per-user authorization lists, so site administrators retain fine-grained control of who can stream which dataset and under which training session.

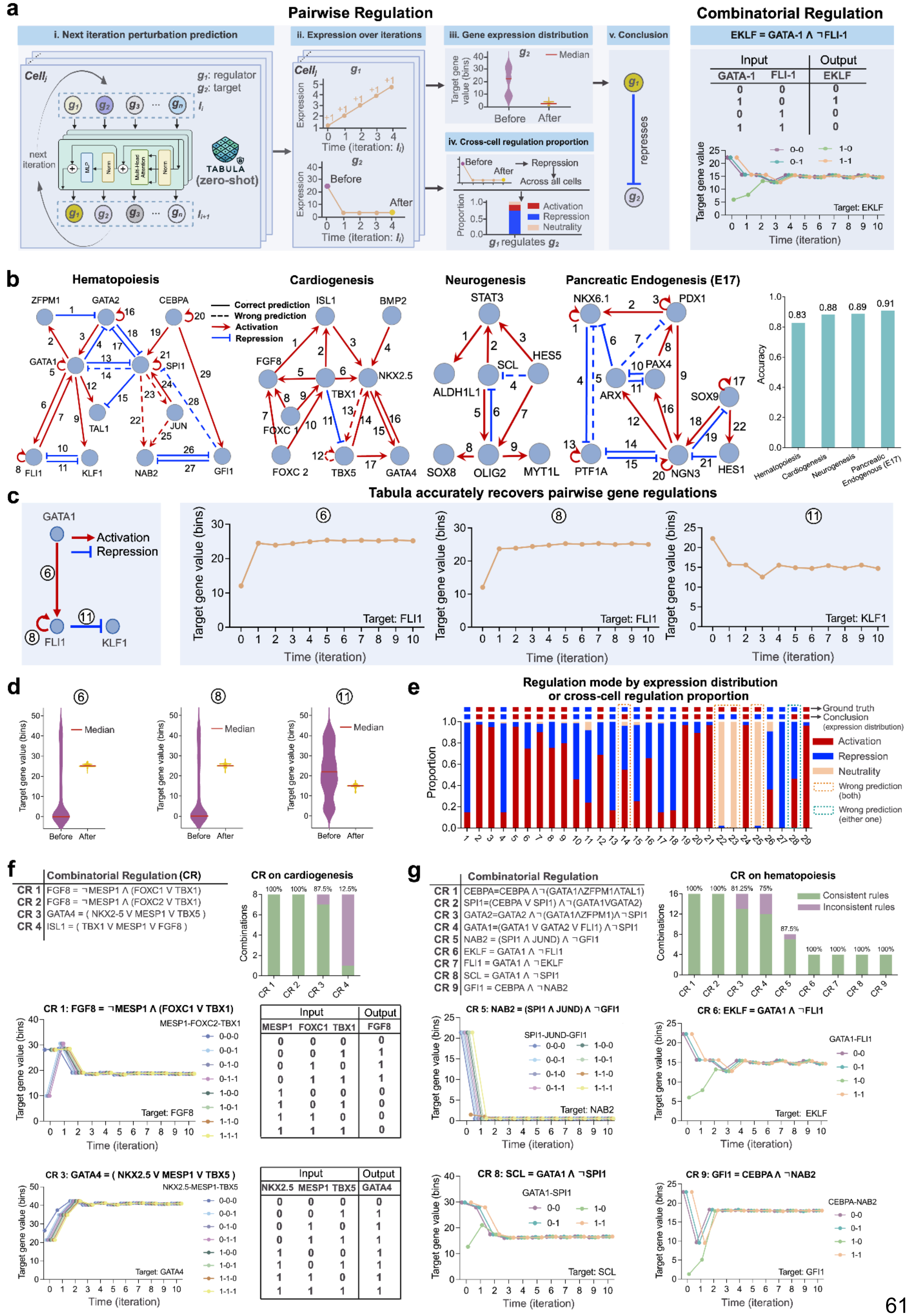
a
Pairwise Regulation
i. Next iteration perturbation prediction
ii. Expression over iterations
iii. Gene expression distribution
v. Conclusion
iv. Cross-cell regulation proportion
Combinatorial Regulation
EKLF = GATA-1 ∧ ¬FLI-1
Input
Output
GATA-1
FLI-1
EKLF
TABULA
(zero-shot)
next iteration
Before
After
Median
Activation
Repression
Neutrality
represses
Across all cells
Time (iteration)
Target gene value (bins)
Target: EKLF
b
Hematopoiesis
Cardiogenesis
Neurogenesis
Pancreatic Endogenesis (E17)
Correct prediction
Wrong prediction
Activation
Repression
Accuracy
0.83
0.88
0.89
0.91
c
Tabula accurately recovers pairwise gene regulations
GATA1
FLI1
KLF1
Target: FLI1
Target: KLF1
d
Median
Before
After
e
Regulation mode by expression distribution or cross-cell regulation proportion
Ground truth
Conclusion (expression distribution)
Wrong prediction (both)
Wrong prediction (either one)
Proportion
f
Combinatorial Regulation (CR)
CR 1 FGF8 = ¬MESP1 ∧ (FOXC1 V TBX1)
CR 2 FGF8 = ¬MESP1 ∧ (FOXC2 V TBX1)
CR 3 GATA4 = ( NKX2-5 V MESP1 V TBX5 )
CR 4 ISL1 = ( TBX1 V MESP1 V FGF8 )
CR on cardiogenesis
Combinations
CR 1: FGF8 = ¬MESP1 ∧ (FOXC1 V TBX1)
Target: FGF8
CR 3: GATA4 = ( NKX2.5 V MESP1 V TBX5 )
Target: GATA4
g
Combinatorial Regulation
CR 1 CEBPA=CEBPA ∧¬(GATA1∧ZFPM1∧TAL1)
CR 2 SPI1=(CEBPA V SPI1) ∧¬(GATA1VGATA2)
CR 3 GATA2=GATA2 ∧¬(GATA1∧ZFPM1)∧¬SPI1
CR 4 GATA1=(GATA1 V GATA2 V FLI1) ∧¬SPI1
CR 5 NAB2 = (SPI1 ∧ JUND) ∧ ¬GFI1
CR 6 EKLF = GATA1 ∧ ¬FLI1
CR 7 FLI1 = GATA1 ∧ ¬EKLF
CR 8 SCL = GATA1 ∧ ¬SPI1
CR 9 GFI1 = CEBPA ∧ ¬NAB2
CR on hematopoiesis
Consistent rules
Inconsistent rules
CR 5: NAB2 = (SPI1 ∧ JUND) ∧ ¬GFI1
Target: NAB2
CR 6: EKLF = GATA1 ∧ ¬FLI1
Target: EKLF
CR 8: SCL = GATA1 ∧ ¬SPI1
Target: SCL
CR 9: GFI1 = CEBPA ∧ ¬NAB2
Target: GFI1

## Figure 3: Tabula accurately predicts pairwise and combinatorial regulation across a variety of biological systems.

a. **The schematic of utilizing Tabula zero-shot prediction to characterize both pairwise and combinatorial regulations**. The left subpanel outlines Tabula's strategy for predicting pairwise regulation across cells between a regulator $g_1$ and its target $g_2$ through an iterative perturbation zero-shot prediction procedure. (i) This iterative update procedure involves consistently increasing or decreasing the rank of gene expression of the regulator by one across cells, followed by zero-shot predicting the expression of all genes after passing through the Tabula model. Then the updated gene expression is used as the input for the next iteration. (ii) The predicted trajectories of the expression dynamics across both the regulator and target under the above *in silico* perturbations can then be obtained and subsequently used to reveal $g_1$'s effect on $g_2$. (iii) Diagram of target gene expression distribution. Expression distribution (median values indicated by red lines) of the target before and after the iterative perturbation. By default, this approach is used to conclude the type of gene regulation, see (v). (iv) Diagram of cross-cell regulation proportion. Cross-cell regulation proportions depict the relative fraction of activatory, repressive, or neutral regulation modes from the regulator to the target across all cells. (v) Based on the median gene expression before and after iterative perturbation, we can conclude the gene regulation mode. In this case, we find the expression of $g_2$ decreases after the iterative perturbation and we thus conclude a repression regulation from $g_1$ to $g_2$. Both the median expression based regulatory mode and cross-cell regulation proportion can be used for determining the gene regulation mode in this study. The right subpanel illustrates how Tabula determines combinatorial regulation using the example of the combinatorial regulation from GATA1 and FLI-1 to EKLF. We first enumerate all possible boolean logic gates from GATA1 and FLI1 to EKLF and follow the procedure from the left panel to perform iterative perturbation prediction where the perturbation is executed on both the GATA1 and FLI1. Note that if the value is set to be zero (one) on the logic gate, the regulator will decrease (increase) the gene rank by one. Similarly, based on the target gene expression dynamics across iterations under different perturbation input, we conclude it matches up with the combinatorial logic from GATA1 and FLI1 to EKLF.

b. **Four biological systems with extensively validated regulatory networks that Tabula applied to and its overall performance on pairwise regulation prediction**. The core regulatory networks for hematopoiesis (Fig. 4b-e) [35], cardiogenesis (Fig. 4b) [37], neurogenesis (Fig. 4b) [39], and pancreatic endogenesis (E17) (Fig. 4b) [41] is shown on the left. In the network diagram, an activation regulation is denoted by red arrows, repression regulation by blue arrows, with solid arrows indicating correct predictions and dashed arrows incorrect ones. The bar plot on the right quantifies the global performance of the Tabula model across all systems.

c. Tabula recovers pairwise gene regulations within a three-gene example network motif in the hematopoiesis system [35]. The left diagram illustrates regulatory relationships between GATA1, FLI1, and KLF1. The line plots on the right display predicted expression levels of target genes (FLI1 and KLF1) over iterations after increasing the regulator, matching up with expected expression dynamics.

d. **Distribution of target gene expression pre- and post-perturbation under Tabula prediction in a three-gene example network motif from the hematopoiesis system** [35]. Violin plots illustrate expression levels of target genes for interactions 6, 8, and 11 before and after perturbation.
e. **Comparison of regulation mode identified by expression distribution shift and cross-cell regulation proportions**. Bar plot shows activation (red), repression (blue), and neutrality (orange) proportions for each interaction of the hematopoiesis system [35], based on cross-cell regulation proportions (see panel **a-iv**). The first row of the regulation mode shown on the top of each bar represents ground truth mode while the second row of the regulation mode indicates the prediction from the expression distribution. Highlighted boxes with dashed lines denote inconsistent prediction: orange dashed box for incorrect predictions from both expression distribution shift based and cross-cell regulation proportion based regulation modes, while blue dashed box for incorrect prediction in only one such prediction.
f. **Combinatorial regulation (CR) predictions for the cardiogenesis system** [97]. The table (**top left**) lists four CR rules, where multiple regulators work in concert to regulate the expression of FGF8, GATA4, and ISL1 through specific combinatorial regulatory logic. The bar plot (**top right**) illustrates prediction accuracy for each CR rule, with the x-axis indicating the index of CR. Line plots (left of the second and third rows) depict expression dynamics of FGF8 and GATA4 over *in silico* perturbation iterations under different configurations of the regulators, corresponding to CR1 and CR3. Boolean tables (bottom right) detail the groundtruth combinatorial logic for each rule.
g. Same as in panel **f** but for the hematopoiesis system [35].

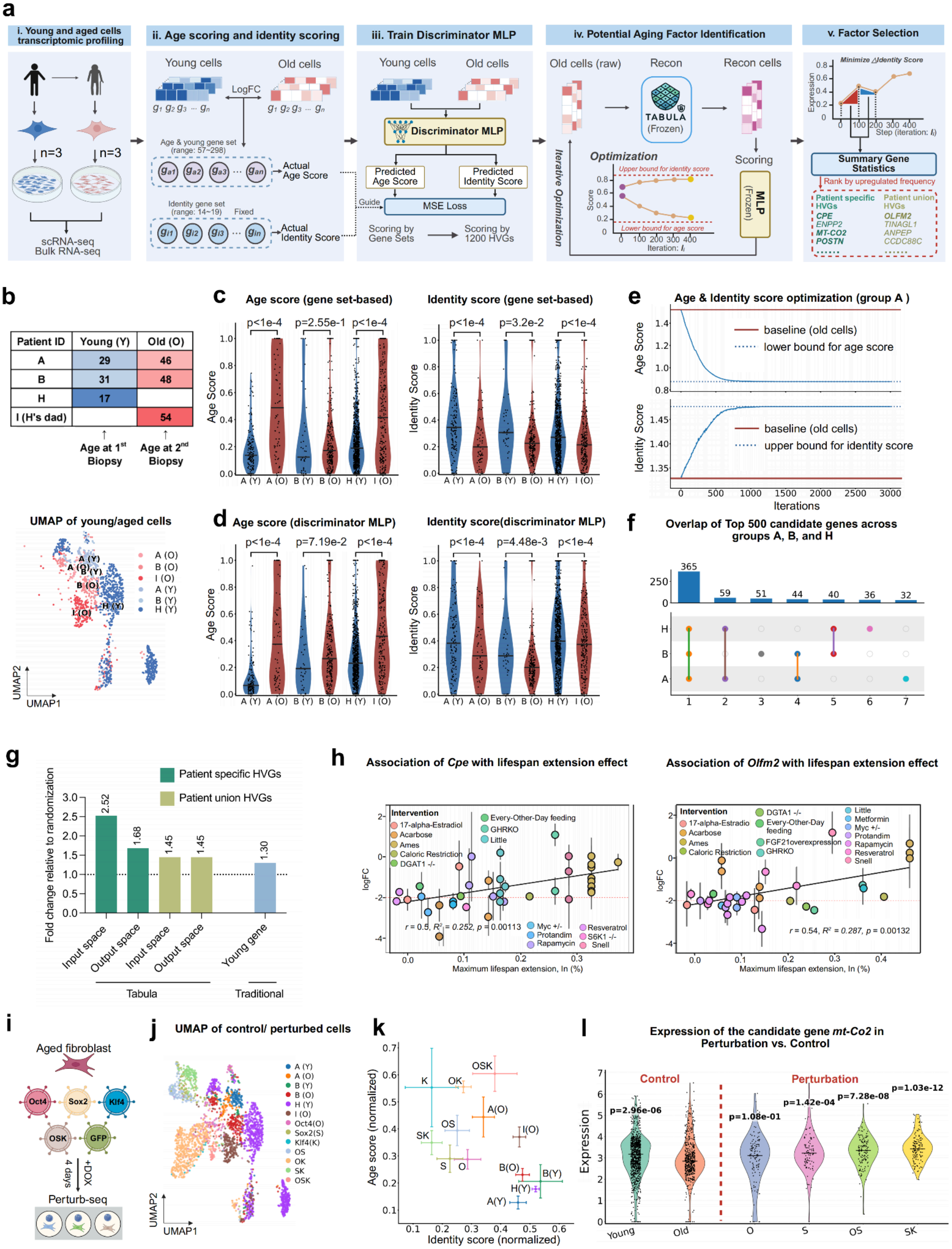
a
i. Young and aged cells transcriptomic profiling
n=3
n=3
scRNA-seq
Bulk RNA-seq
ii. Age scoring and identity scoring
Young cells
Old cells
LogFC
Age & young gene set (range: 57~298)
Actual Age Score
Identity gene set (range: 14~19)
Fixed
Actual Identity Score
iii. Train Discriminator MLP
Young cells
Old cells
Discriminator MLP
Predicted Age Score
Predicted Identity Score
Guide
MSE Loss
Scoring by Gene Sets
Scoring by 1200 HVGs
iv. Potential Aging Factor Identification
Old cells (raw)
Recon
Recon cells
TABULA (Frozen)
Iterative Optimization
Optimization
Upper bound for identity score
Lower bound for age score
Scoring
MLP (Frozen)
Iteration: $I_i$
v. Factor Selection
Minimize △Identity Score
Expression
Step (iteration: $I_i$)
Summary Gene Statistics
Rank by upregulated frequency
Patient specific HVGs
CPE
ENPP2
MT-CO2
POSTN
Patient union HVGs
OLFM2
TINAGL1
ANPEP
CCDC88C
b
Patient ID | Young (Y) | Old (O)
A | 29 | 46
B | 31 | 48
H | 17 |
I (H's dad) | | 54
Age at 1st Biopsy
Age at 2nd Biopsy
UMAP of young/aged cells
A (O)
B (O)
I (O)
A (Y)
B (Y)
H (Y)
UMAP1
UMAP2
c
Age score (gene set-based)
Identity score (gene set-based)
p<1e-4
p=2.55e-1
p<1e-4
p<1e-4
p=3.2e-2
p<1e-4
Age Score
Identity Score
d
Age score (discriminator MLP)
Identity score(discriminator MLP)
p<1e-4
p=7.19e-2
p<1e-4
p<1e-4
p=4.48e-3
p<1e-4
e
Age & Identity score optimization (group A )
baseline (old cells)
lower bound for age score
upper bound for identity score
Iterations
f
Overlap of Top 500 candidate genes across groups A, B, and H
365
59
51
44
40
36
32
g
Patient specific HVGs
Patient union HVGs
Fold change relative to randomization
2.52
1.68
1.45
1.45
1.30
Input space
Output space
Input space
Output space
Young gene
Tabula
Traditional
h
Association of Cpe with lifespan extension effect
Association of Olfm2 with lifespan extension effect
Intervention
17-alpha-Estradiol
Acarbose
Ames
Caloric Restriction
DGAT1 -/-
Every-Other-Day feeding
GHRKO
Little
Myc +/-
Protandim
Rapamycin
Resveratrol
S6K1 -/-
Snell
FGF21overexpression
Metformin
r = 0.5, $R^2$ = 0.252, p = 0.00113
r = 0.54, $R^2$ = 0.287, p = 0.00132
logFC
Maximum lifespan extension, ln (%)
i
Aged fibroblast
Oct4
Sox2
Klf4
OSK
GFP
4 days
+DOX
Perturb-seq
j
UMAP of control/ perturbed cells
Oct4(O)
Sox2(S)
Klf4(K)
OS
OK
SK
OSK
k
Age score (normalized)
Identity score (normalized)
l
Expression of the candidate gene mt-Co2 in Perturbation vs. Control
Control
Perturbation
p=2.96e-06
p=1.08e-01
p=1.42e-04
p=7.28e-08
p=1.03e-12
Expression
Young
Old
O
S
OS
SK

# Figure 4: Tabula predicts skin rejuvenation factors through age- and identity score-guided in silico optimization.

a. **The schematic of utilizing Tabula to identify candidate rejuvenation factors through score-guided in silico optimization.** The workflow begins with transcriptomic profiling of young and aged fibroblasts from matched biological groups using scRNA-seq: (i) Differential expression between young and aged cells is then used to derive old-associated and young-associated gene sets for age scoring, whereas the identity score is defined using a fixed set of known fibroblast identity genes [43]. (ii) These scores are next used to train a discriminator multilayer perceptron (MLP), which takes cellular gene expression as input and predicts age and identity scores. (iii) To identify potential rejuvenation factors, aged cells were iteratively optimized in silico by treating the input expression matrix as the optimization variable. At each step, the current cell state was passed through a frozen Tabula model fine-tuned for skin dataset (see Methods) and evaluated by a frozen scoring MLP, and gradients from the age and identity score objective were used to update the input toward a younger, identity-preserving state; the optimization aims to reduce the age score toward the lower boundary corresponding to the score of simulated 17-year-old cells, while increasing the identity score toward the upper boundary defined as the median identity score of young cells (iv). Finally, genes are ranked by their upregulated frequency during optimization, thereby prioritizing candidate rejuvenation factors from Patient specific and Patient union HVGs settings (v).

b. **Donor information and UMAP visualization of the young and aged fibroblast dataset used in this study.** The top table summarizes donor identities and ages at biopsy for the young (A, B and H) and aged (A, B and I) fibroblast samples. The UMAP below illustrates the low dimensional manifold of fibroblast transcriptomes across donors and age states.

c. **Age and identity score validation in fibroblast groups.** Violin plots illustrate age and identity scores calculated from normalized gene expression using predefined gene sets, with all scores further rescaled to the 0–1 range. Aged cells consistently show elevated age scores and reduced identity scores relative to young cells across groups, supporting the use of these two scores as optimization objectives for identifying candidate rejuvenation factors. *P* values are shown above each comparison.

d. **MLP predicts age and identity scores in fibroblast groups.** Same as in panel c but for age and identity scores predicted by the discriminator MLP.

e. **Optimization trajectories of age and identity scores during in silico optimization of aged fibroblast group A.** The top plot shows that the age score progressively decreases from the baseline of old cells toward the predefined lower boundary, whereas the bottom plot shows that the identity score correspondingly increases toward the upper boundary defined by young cells. These results indicate that the optimization procedure successfully drives reconstructed cells toward a more rejuvenated yet identity-preserving transcriptional state.

f. **Overlap of the top 500 candidate rejuvenation genes across fibroblast groups A, B, and H.** The UpSet plot summarizes shared and group-specific candidate genes identified from the accumulated path score ranking.
g. **Enrichment of top-ranked candidate rejuvenation genes in independent longevity intervention signatures.** Bar plots showing the fold enrichment of validated rejuvenation genes among the top 10 candidates nominated under different Tabula settings, compared with the young gene set derived from the whole transcriptome. Rejuvenation genes were defined using the mSALT-derived reference set[45] (slope > 0 and $P < 0.05$). The baseline was calculated as the expected fraction of mSALT-positive genes in randomly selected, size-matched gene sets drawn from the corresponding HVG pool for each Tabula setting, or from the whole transcriptome for the young gene set (see **Supplementary Fig. 12**). The Top10 genes from Patient specific HVGs Input space, Patient specific HVGs Output space, Patient union HVGs Input space, and Patient union HVGs Output space showed 2.52-fold, 1.68-fold, 1.45-fold, and 1.45-fold enrichment, respectively. The young gene set showed a 1.30-fold enrichment over baseline expectation.
h. **Association of the candidate gene *Cpe/Olfm2* with lifespan extension effects across independent mouse longevity interventions based on the mSALT database.** Each dot represents a single dataset and corresponds to the mean hepatic log fold change of *Cpe/Olfm2* under a specific intervention, with error bars indicating standard errors. The x-axis denotes the maximum lifespan extension achieved by the intervention. Both genes' expression change is positively associated with lifespan extension effect (*Cpe*: $r = 0.5$, $R^2 = 0.252$, $p = 0.00113$; *Olfm2*: $r = 0.54$, $R^2 = 0.287$, $p = 0.00132$).
i. **Schematic of the overexpression Perturb-seq experiment in aged fibroblasts.** Aged fibroblasts were subjected to perturbations of Yamanaka reprogramming factors to induce a shift toward a younger transcriptional state, followed by single-cell transcriptomic profiling for downstream analysis.
j. **UMAP visualization of control and perturbed cells.** The plot shows the transcriptional UMAP embedding of cells under different perturbation conditions.
k. **Summary of normalized age and identity scores across perturbation conditions.** Each perturbation is positioned in the two-dimensional space defined by identity score and age score, providing a compact overview of the transcriptional effects of different perturbation conditions.
l. **Expression of the candidate gene *mt-Co2* in control and perturbation conditions.** Violin plots compare *mt-Co2* expression between young and old cells in the control setting and across multiple perturbation conditions, with old cells used as the baseline for statistical comparison. Corresponding *P* values are shown above each comparison.

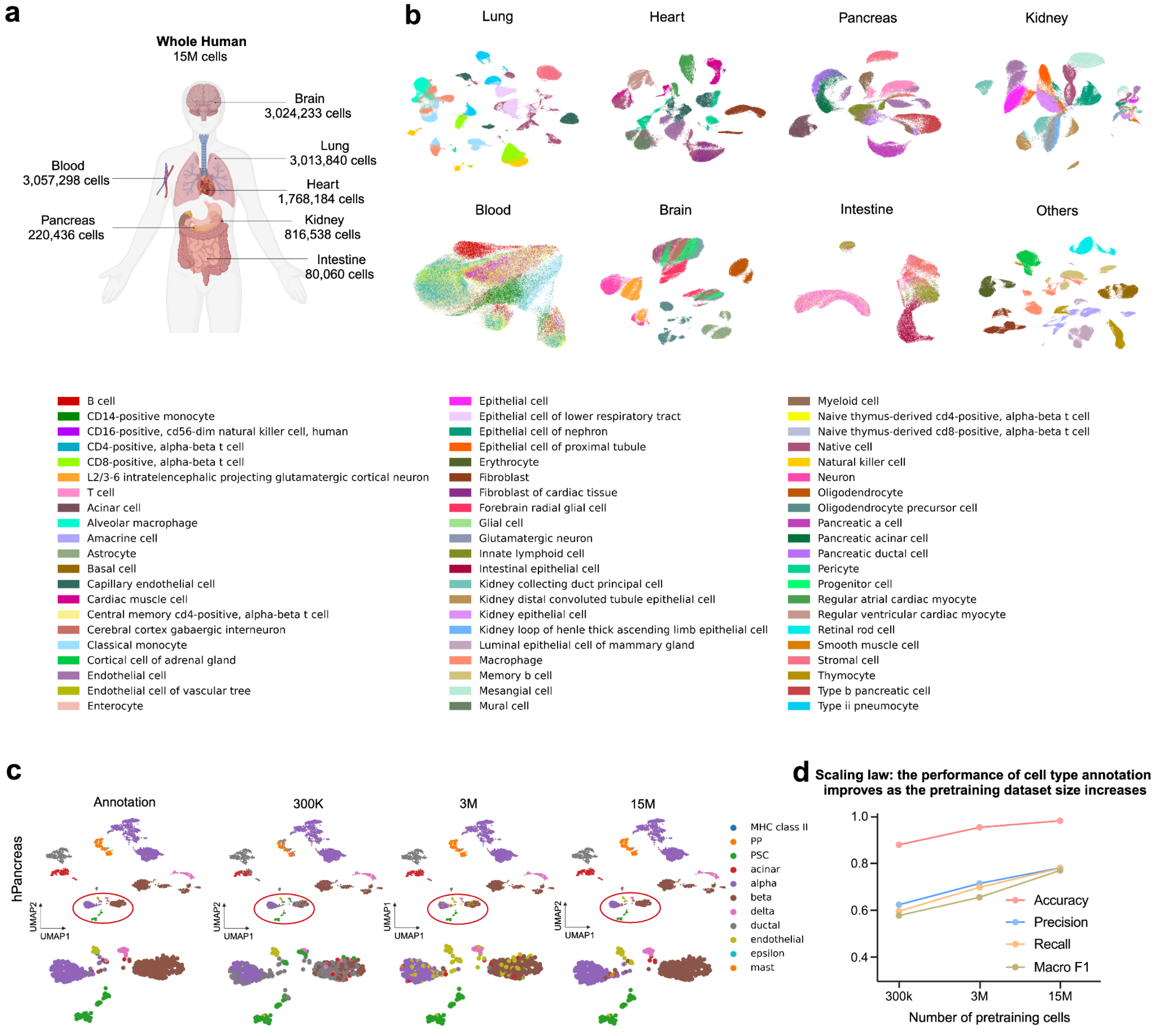

# Supplementary Figure 1: The overview of pretraining dataset and scaling law in Tabula.

a. Summary of Tabula's pretraining scRNA-seq data's tissue type distribution in the human. Tabula's pretraining dataset includes around 15 million (M) scRNA-seq data distributed across many human tissues. To demonstrate the Tabula federated learning framework, we assign each tissue to a distinct client.
b. UMAP visualization of cell embeddings from pretrained Tabula client models, with 10,000 cells randomly sampled from each client. The "Others" client aggregates all less-represented tissues.
c. Tabula based pancreas-specific cell type annotation on the hPancreas dataset [88], visualized in the UMAP embedding. The performance improves as the size of the pretraining dataset increases from 300k to 3M, and to 15M cells.

d. Scaling law: Impact of pretraining dataset size on the performance of cell type annotation in hPancreas dataset in panel **c** (evaluated by accuracy, precision, recall, and macro F1).

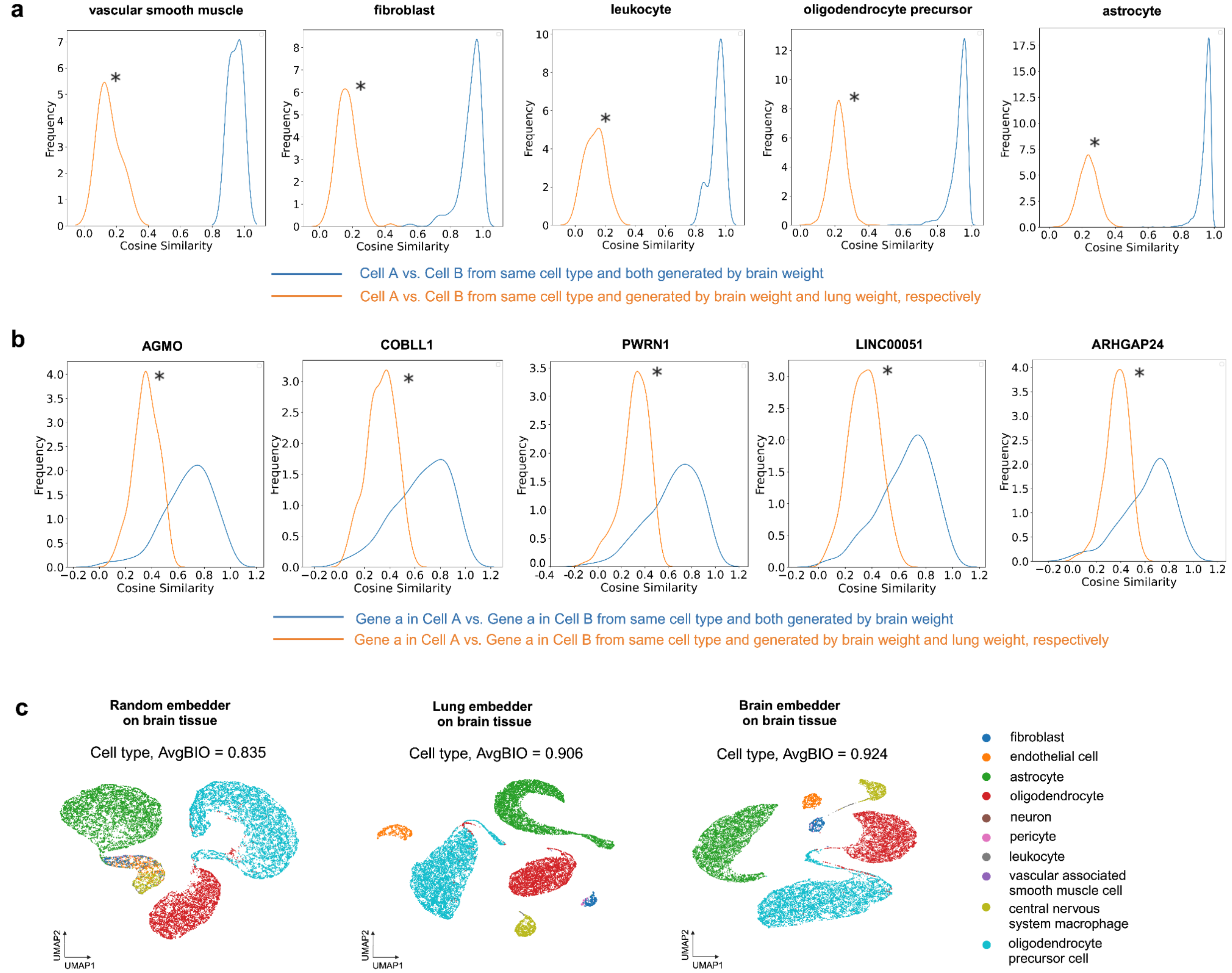

# Supplementary Figure 2: tissue-specific embedders encode distinctive tissue features.

a. Tissue-specific embedding captures different biological contexts across various cell types at cell level. Perirhinal cortex dataset from brain tissue is adopted as an example. Each plot compares the cosine similarity between every possible cell pair from the same type both generated by brain-specific Tabula (line in blue) and between cells of the same type but derived from different tissues, brain and lung-specific Tabulaweights, respectively (line in orange). The distribution indicates the degree of similarity, with peaks close to 1 suggesting high intra-group consistency. The distribution from the blue lines is significantly shifted to the right (* indicates $p < 0.05$ by Wilcoxon test, FDR-corrected) highlights the tissue-specific Tabula models learn unique cellular molecular profiles of tissues.
b. Same as in panel **a** but for comparing the cosine similarity of gene representations.
c. Tissue-specific Tabula model demonstrates enhanced performance on corresponding datasets. By fine-tuning the Tabula model using diverse embedders on the Perirhinal cortex dataset derived from brain tissue, the brain-specific Tabula embedder notably outperforms embedders initialized with weights from other tissues and those with random initialization. This finding highlights the benefits of employing tissue-specific embedders in improving downstream tasks tailored to specific biological contexts.

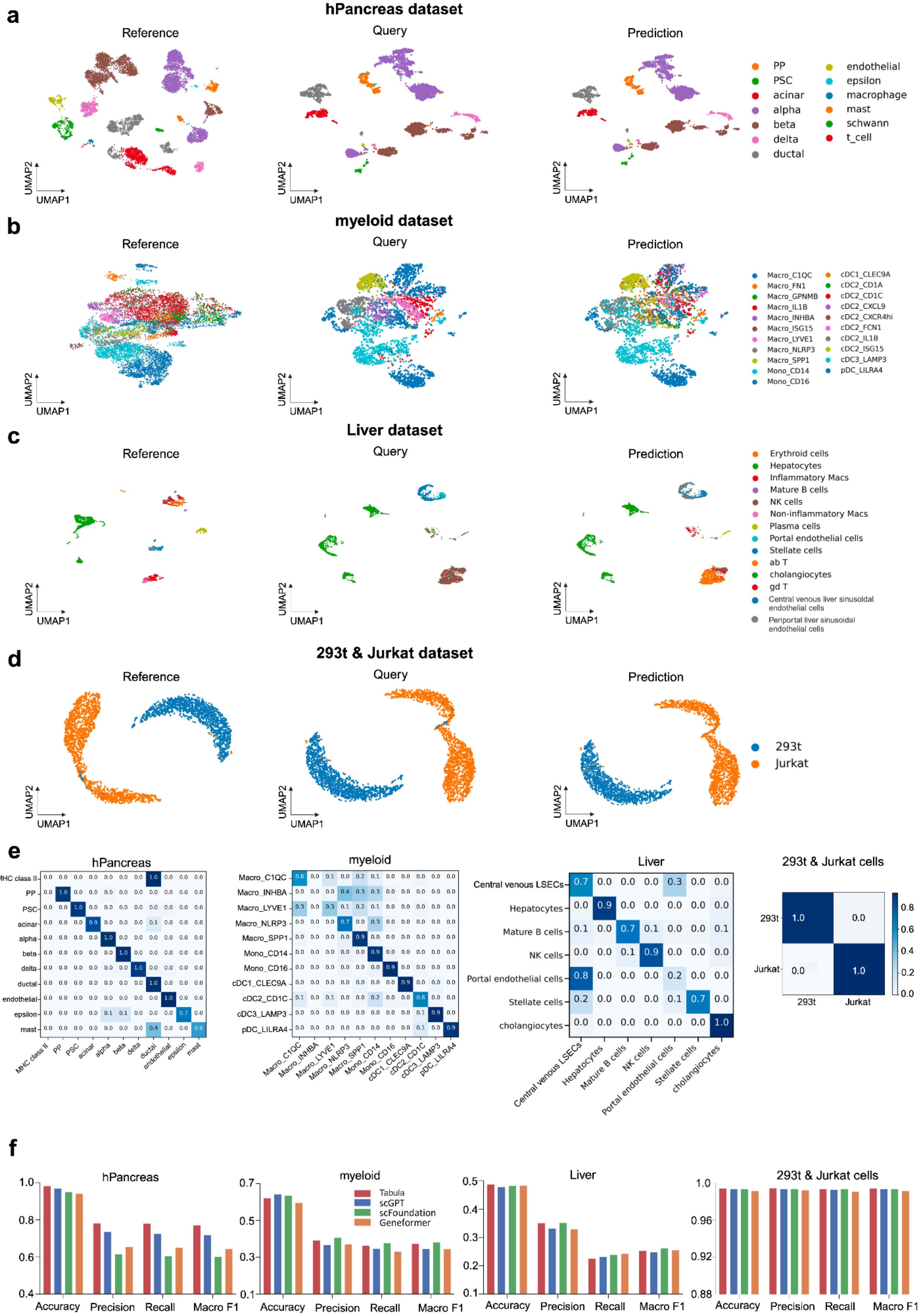
a
hPancreas dataset
Reference
Query
Prediction
PP
PSC
acinar
alpha
beta
delta
ductal
endothelial
epsilon
macrophage
mast
schwann
t_cell
b
myeloid dataset
c
Liver dataset
Erythroid cells
Hepatocytes
Inflammatory Macs
Mature B cells
NK cells
Non-inflammatory Macs
Plasma cells
Portal endothelial cells
Stellate cells
ab T
cholangiocytes
gd T
Central venous liver sinusoidal endothelial cells
Periportal liver sinusoidal endothelial cells
d
293t & Jurkat dataset
293t
Jurkat
e
hPancreas
myeloid
Liver
293t & Jurkat cells
f
Tabula
scGPT
scFoundation
Geneformer
Accuracy
Precision
Recall
Macro F1

## Supplementary Figure 3: Tabula generally outperforms state-of-the-art single cell foundation model in annotating cell types.

a. UMAP visualization of the hPancreas dataset, colored by cell types. The reference dataset (left) was used to fine-tune Tabula, which was evaluated on the query dataset (middle). The predicted annotations (right) demonstrate the model's annotation capability.
b. Similar to (a), but for the myeloid dataset.
c. Similar to (a), but for the Liver dataset.
d. Similar to (a), but for the 293t & Jurkat cells dataset.
e. Confusion matrices for cell type annotation prediction of the hPancreas, Myeloid, Liver, and 293T & Jurkat cells datasets, revealing a majority of correct prediction along the diagonal. The x-axis represents the predictions, while the y-axis corresponds to the ground truth.
f. Barplots of performance metrics (accuracy, precision, recall, Macro F1) of Tabula against scGPT, scFoundation, and Geneformer across the four datasets. Tabula consistently achieves superior or competitive results.

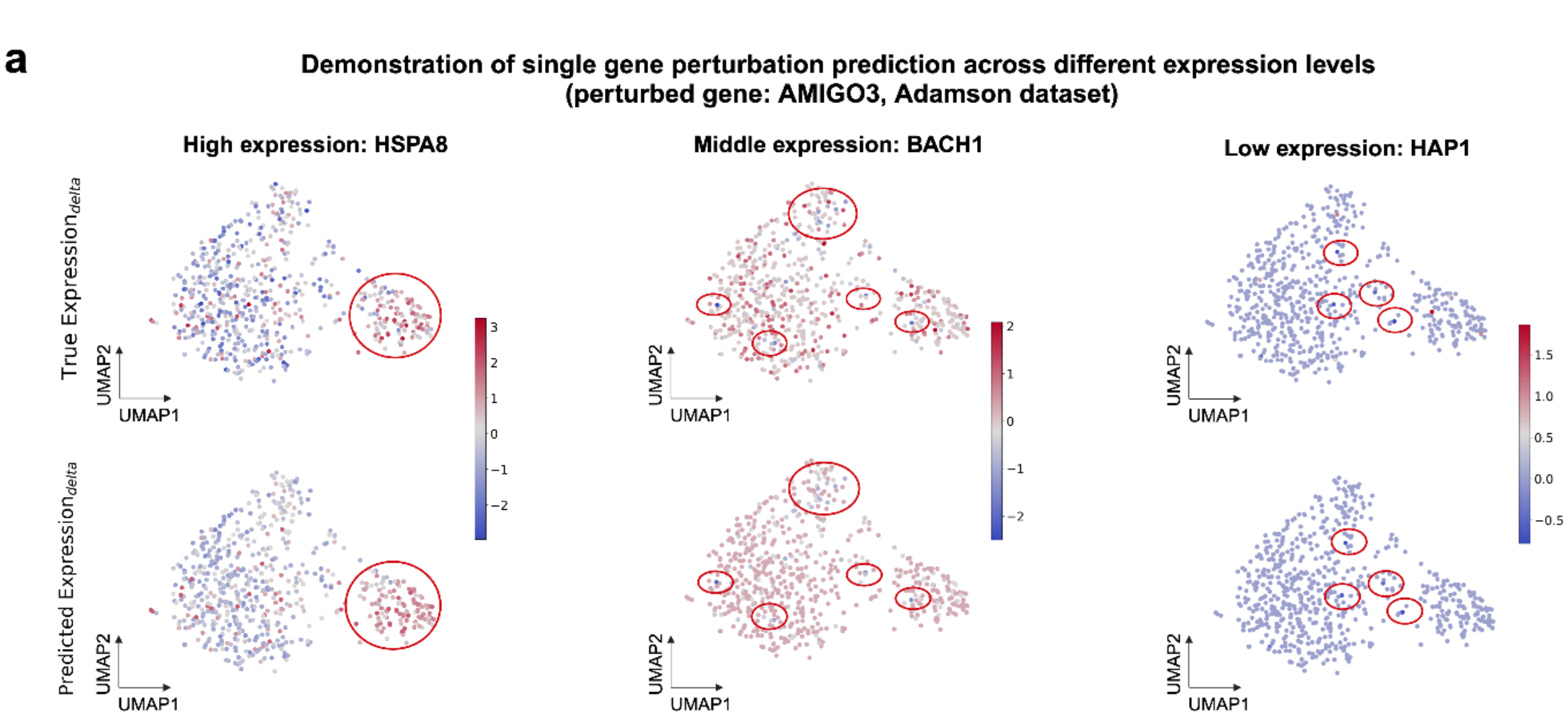
a
Demonstration of single gene perturbation prediction across different expression levels
(perturbed gene: AMIGO3, Adamson dataset)
High expression: HSPA8
Middle expression: BACH1
Low expression: HAP1
True Expression$_{delta}$
Predicted Expression$_{delta}$
UMAP1
UMAP2

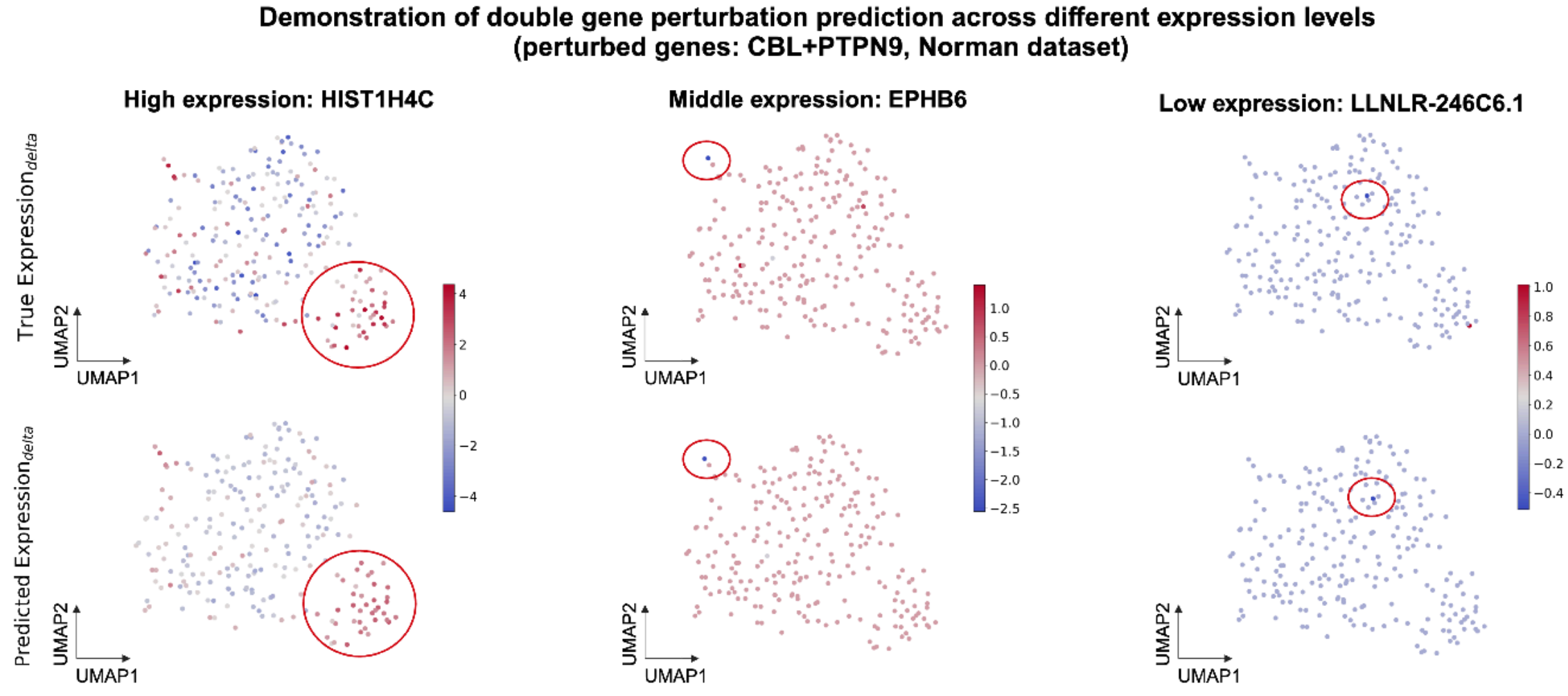
b
Demonstration of double gene perturbation prediction across different expression levels
(perturbed genes: CBL+PTPN9, Norman dataset)
High expression: HIST1H4C
Middle expression: EPHB6
Low expression: LLNLR-246C6.1
True Expression$_{delta}$
Predicted Expression$_{delta}$
UMAP1
UMAP2

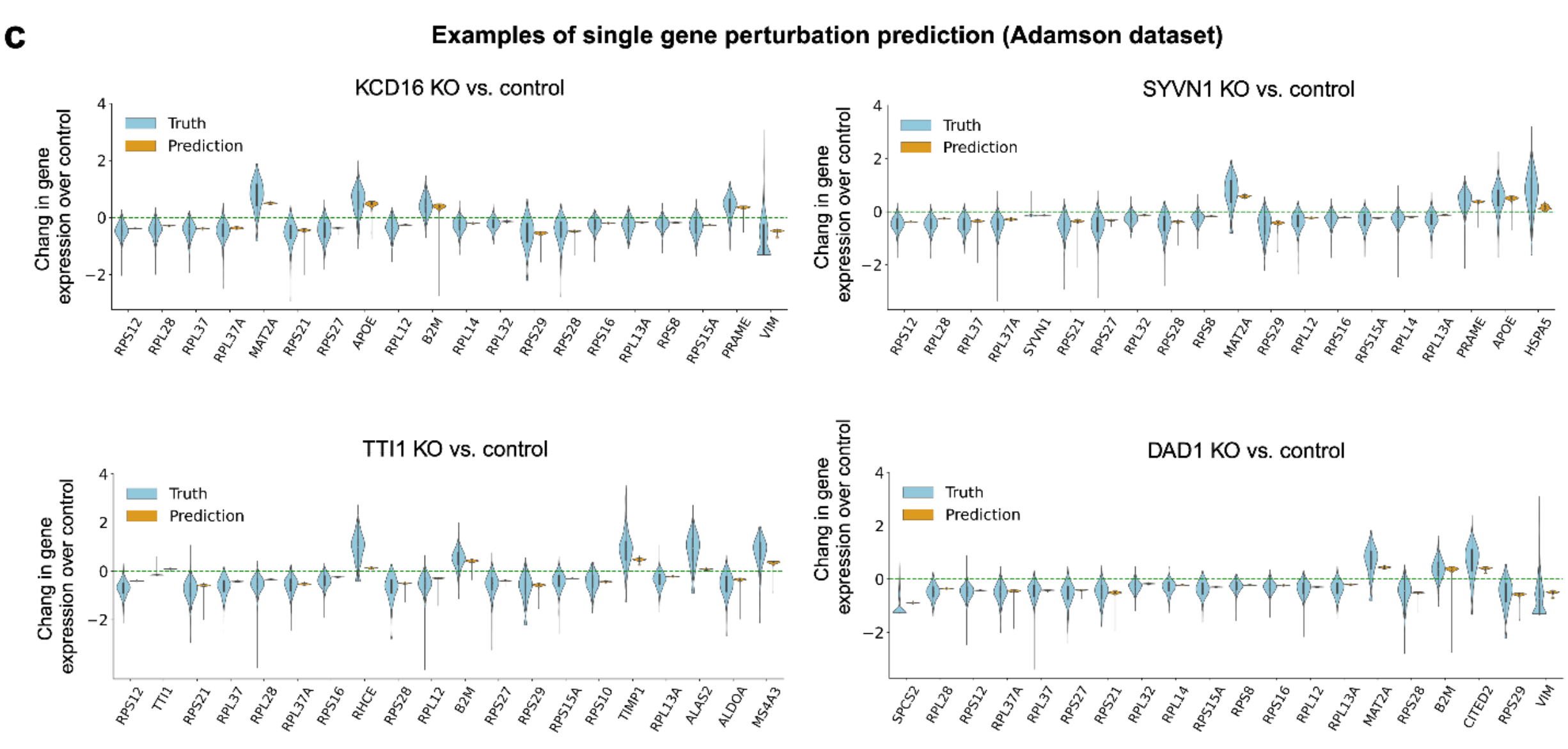
c
Examples of single gene perturbation prediction (Adamson dataset)
KCD16 KO vs. control
SYVN1 KO vs. control
TTI1 KO vs. control
DAD1 KO vs. control
Truth
Prediction
Chang in gene expression over control

## Supplementary Figure 4: Scatterplot of target genes after single- or double-gene perturbation and examples of gene expression changes in response to single-gene perturbation.

a. Scatterplots of example target genes with different expression levels after single-gene perturbation from the Adamson dataset [53]. The perturbed gene is AMIGO3, and the effects are displayed for three target individual genes with different expression levels: high (HSPA8), medium (BACH1), and low (HAP1).
b. Scatterplots of example target genes with different expression levels after double-gene perturbation from the Norman dataset [86]. The perturbed genes are CBL and PTPN9, and the effects are displayed for three target individual genes with different expression levels: high (HIST1H4C), medium (EPHB6), and low (LLNLR-246C6.1).
c. Comparison of the distribution of gene expression changes of top 20 DE genes from ground truth (blue) and Tabula's predictions (orange) in response to perturbations of KCD16, SYVN1, TTI1, and DAD1 relative to control conditions (similar to **Fig. 4d**).

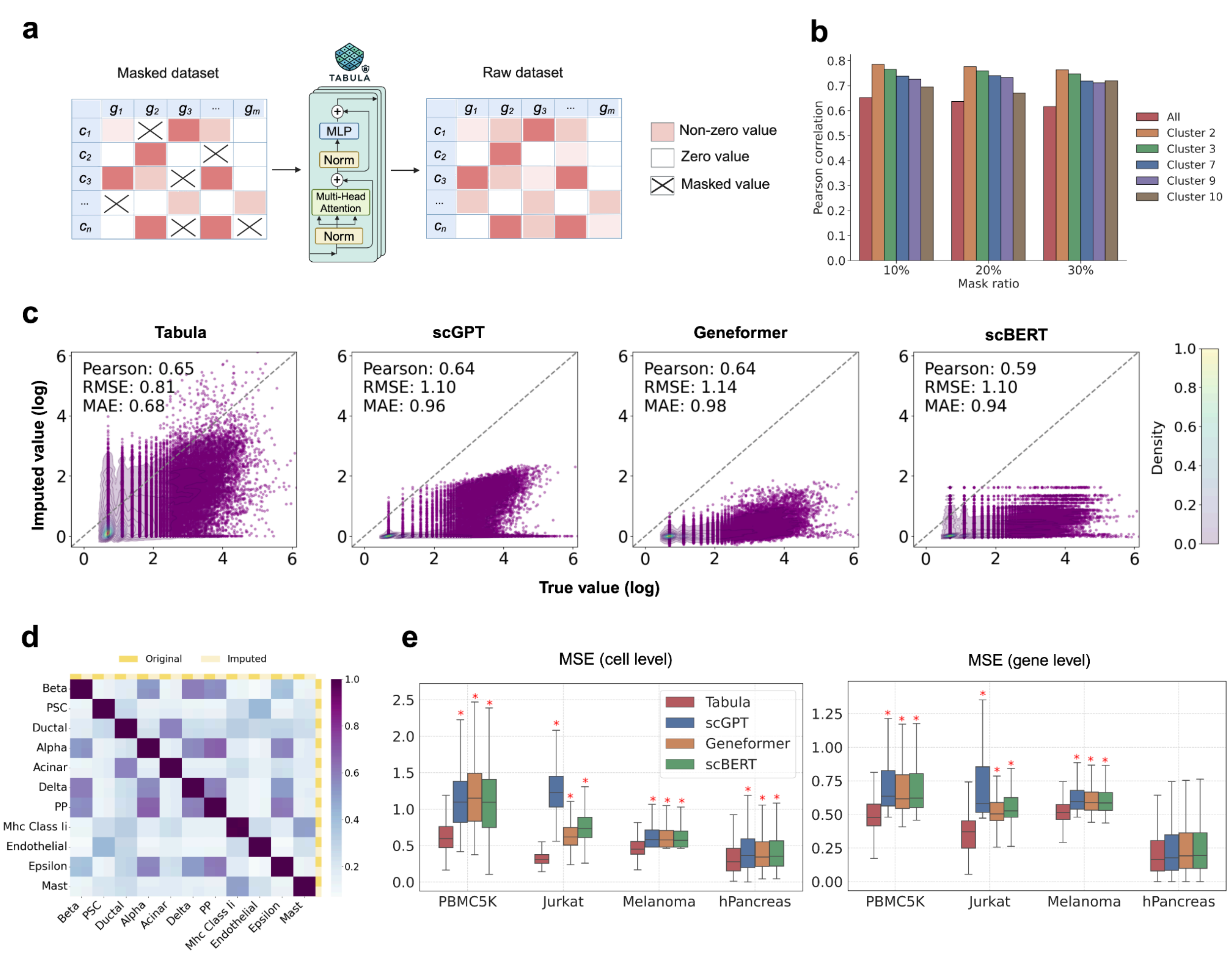

## Supplementary Figure 5: Tabula surpasses existing foundation models in gene expression imputation tasks across a variety of datasets.

a. Schematic of the data imputation task: The dropout probabilities derived from an exponential distribution of gene expression are utilized to weight and sample non-zero values from a subset of genes within each cell, masking their expression to zero to simulate gene dropout. The fine-tuned Tabula model is subsequently applied to recover the true expression values of these masked genes.
b. Robustness of the gene expression imputation by Tabula across varying gene masking ratios of 10%, 20%, and 30% on the PBMC5K dataset. The evaluation includes pearson correlation for both the overall dataset and the top five clusters.
c. Scatterplot comparison of ground truth (x-axis) vs. imputed values (y-axis) across Tabula, scGPT, Geneformer, and scBERT on the PBMC5K dataset. Log-transformed values are used for the density plot while the color of the points indicates the density of imputed values. Quantitative metrics, including Pearson correlation coefficient, Mean Absolute Error and Root Mean Square Error, are provided for each method.
d. Fidelity and consistency of cell type similarity before and after Imputation. The heatmaps visualizes the degree of similarity between cell types in the original and imputed data on hPancreas dataset. Coloured squares at the edges represent the original and imputed data, while the varying colours within the heatmap illustrate the similarity between cell types. Tabula preserves the similarity between original and imputed data within the same cell type, while also maintaining the distinctions between different cell types.
e. Boxplot of the mean squared error of the original and imputed values at both the cell-cell and gene-gene levels across four different datasets under four methods. Wilcoxon rank-sum test is used to determine the p-value between Tabula and other methods. $P < 0.05$ is denoted with an asterisk.

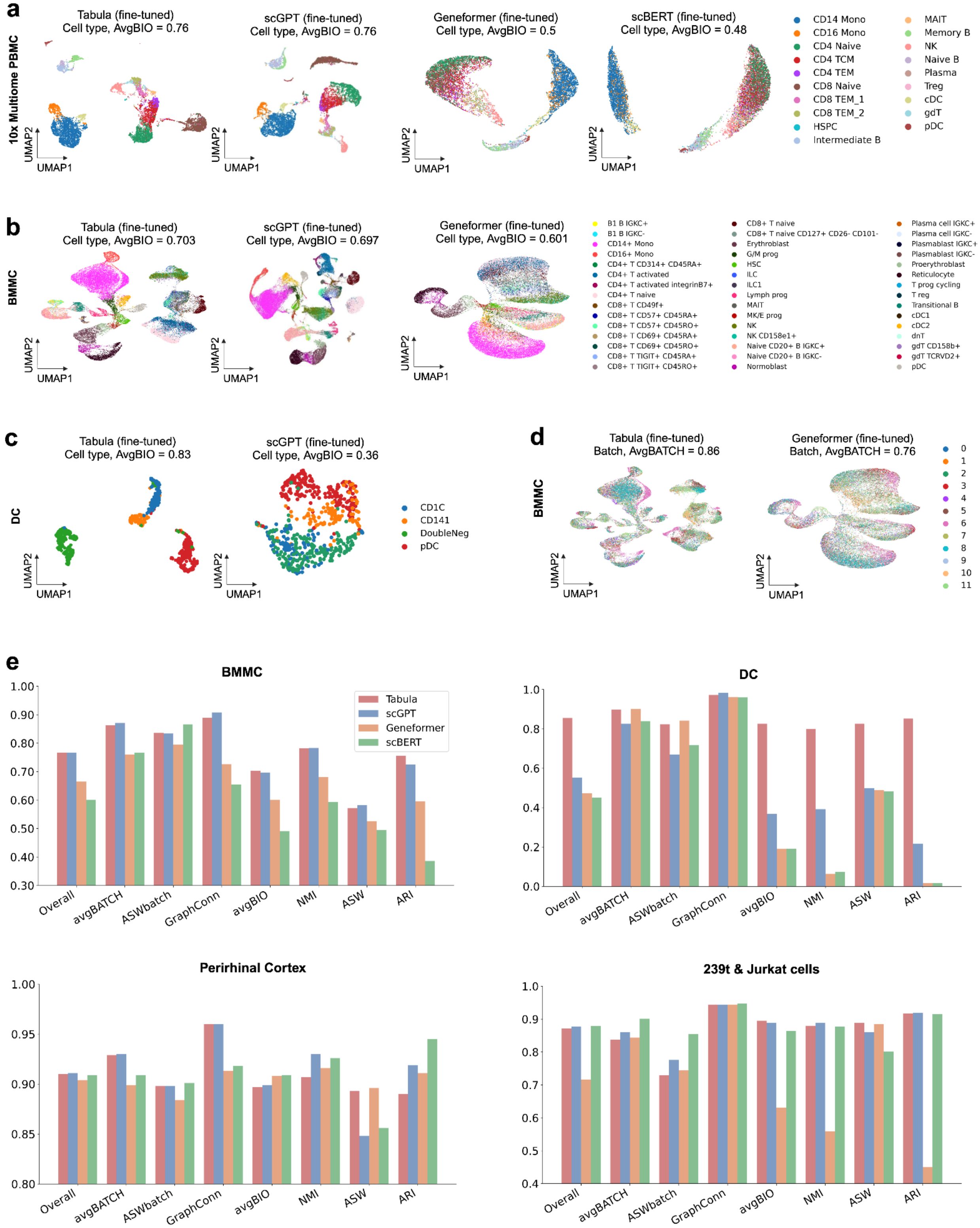
a
10x Multiome PBMC
Tabula (fine-tuned)
Cell type, AvgBIO = 0.76
scGPT (fine-tuned)
Cell type, AvgBIO = 0.76
Geneformer (fine-tuned)
Cell type, AvgBIO = 0.5
scBERT (fine-tuned)
Cell type, AvgBIO = 0.48
UMAP1
UMAP2
CD14 Mono
CD16 Mono
CD4 Naive
CD4 TCM
CD4 TEM
CD8 Naive
CD8 TEM_1
CD8 TEM_2
HSPC
Intermediate B
MAIT
Memory B
NK
Naive B
Plasma
Treg
cDC
gdT
pDC
b
BMMC
Tabula (fine-tuned)
Cell type, AvgBIO = 0.703
scGPT (fine-tuned)
Cell type, AvgBIO = 0.697
Geneformer (fine-tuned)
Cell type, AvgBIO = 0.601
c
DC
Tabula (fine-tuned)
Cell type, AvgBIO = 0.83
scGPT (fine-tuned)
Cell type, AvgBIO = 0.36
CD1C
CD141
DoubleNeg
pDC
d
BMMC
Tabula (fine-tuned)
Batch, AvgBATCH = 0.86
Geneformer (fine-tuned)
Batch, AvgBATCH = 0.76
e
BMMC
DC
Perirhinal Cortex
239t & Jurkat cells
Tabula
scGPT
Geneformer
scBERT
Overall
avgBATCH
ASWbatch
GraphConn
avgBIO
NMI
ASW
ARI

## Supplementary Figure 6: Multi-batch and multi-omics integration performance comparison.

a. UMAP scatterplots of 10x PBMC Multiome dataset showing multi-omics integration by Tabula, scGPT, Geneformer, and scBERT. Tabula and scGPT exhibit superior biological integration performance (AvgBIO = 0.76) thanGeneformer and scBERT.
b. Same as in panel **a** but for the BMMC dataset. Tabula demonstrates the highest AvgBIO (0.703), followed by scGPT (0.697) and then Geneformer (0.601).
c. Same as in panel **a** but for the DC dataset. Tabula achieves a significantly higher AvgBIO score than scGPT.
d. Same as in the panel **a** but for the BMMC dataset. Tabula achieves a significantly higher AvgBATCH score than Geneformer.
e. Barplots of multi-omics and multi-batch integration metrics (AvgBATCH, ASWbatch, GraphConn, AvgBIO, NMI, ASW, ARI) across BMMC, DC, Perirhinal Cortex, and 293T & Jurkat cells datasets and under different models. Tabula consistently achieves top or competitive performance in biological and batch integration.

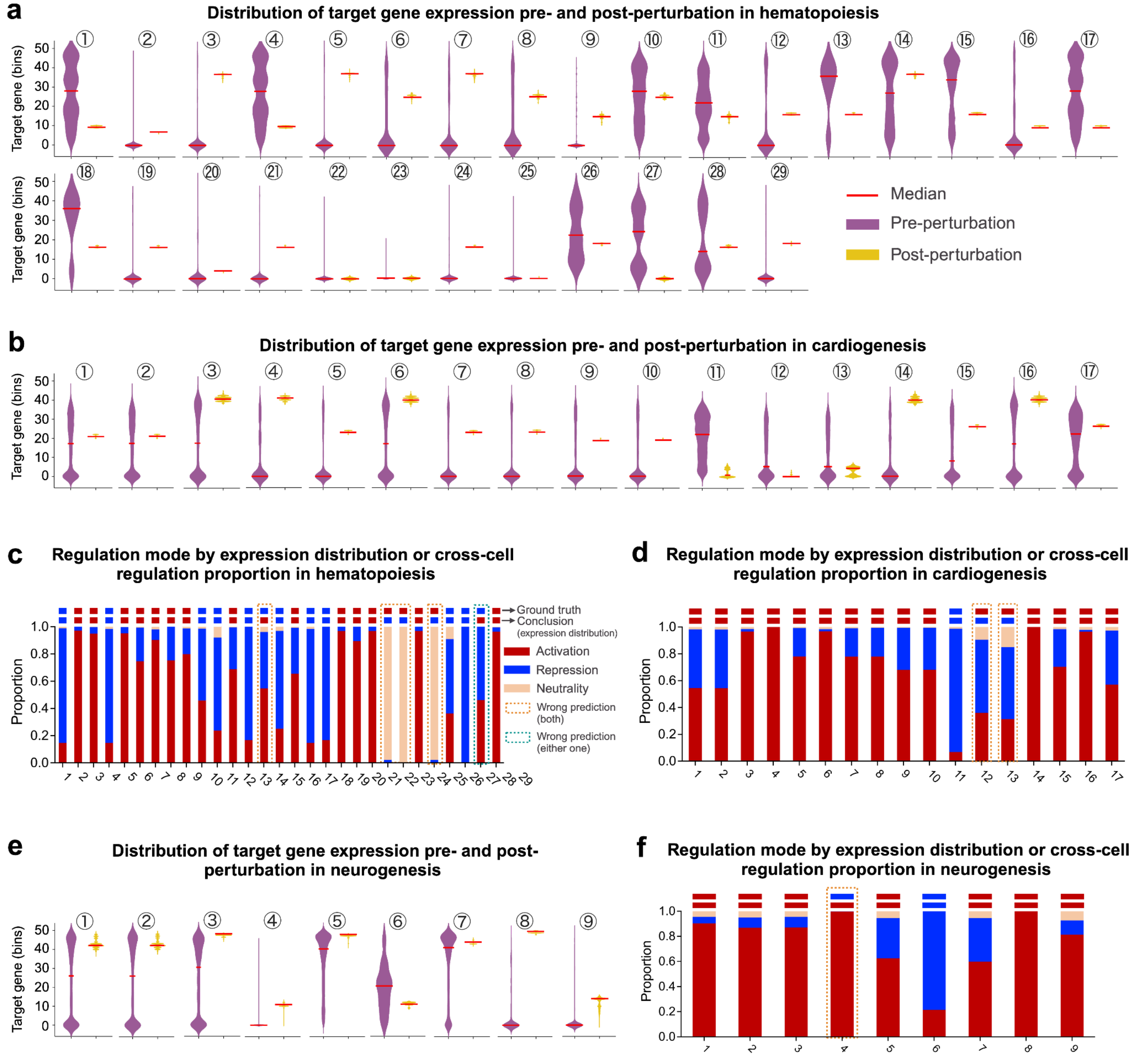

# Supplementary Figure 7: Pairwise regulation prediction by Tabula across hematopoiesis, cardiogenesis, and neurogenesis systems.

a. Distribution of target gene expression pre- and post-perturbation under Tabula prediction within a network motif of 11 genes and 29 edges from the hematopoiesis system [35] (similar to **Fig. 3d**).
b. Same as in panel a but for the network motif of 9 genes and 17 edges cardiogenesis system [37] (similar to **Fig. 3d**).
c. Comparison of regulation mode identified by expression distribution shift and cross-cell regulation proportions in hematopoiesis system [35] (similar to **Fig. 3e**).
d. Same as in panel c but for the cardiogenesis system [37].

e. Same as in panel a but for the network motif with seven genes and nine edges from the neurogenesis system [39].
f. Same as in c but for the neurogenesis system [39].

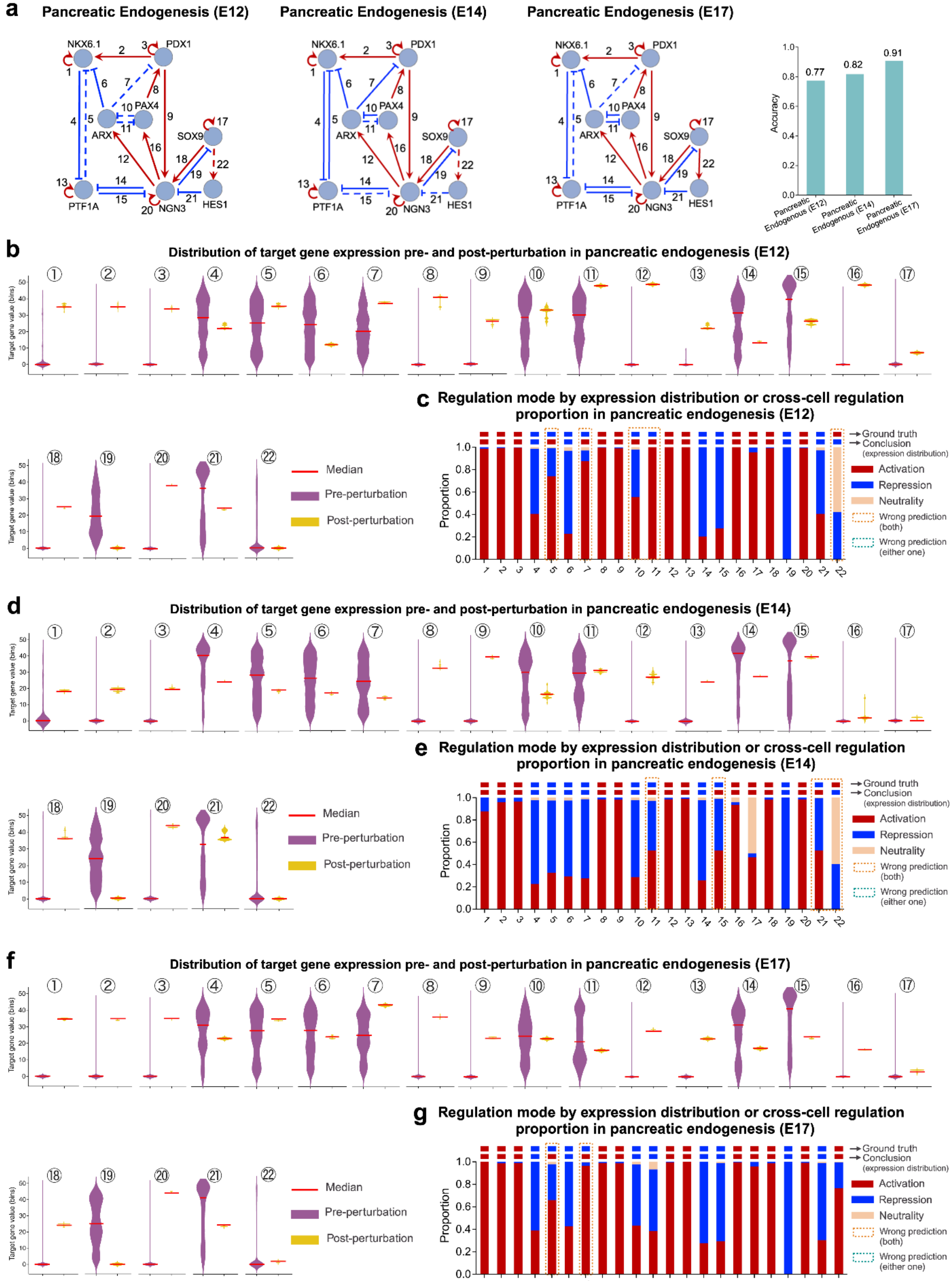
a
Pancreatic Endogenesis (E12)
Pancreatic Endogenesis (E14)
Pancreatic Endogenesis (E17)
NKX6.1
PDX1
PAX4
ARX
SOX9
PTF1A
NGN3
HES1
Accuracy
0.77
0.82
0.91
Pancreatic Endogenous (E12)
Pancreatic Endogenous (E14)
Pancreatic Endogenous (E17)
b
Distribution of target gene expression pre- and post-perturbation in pancreatic endogenesis (E12)
Target gene value (bins)
Median
Pre-perturbation
Post-perturbation
c
Regulation mode by expression distribution or cross-cell regulation proportion in pancreatic endogenesis (E12)
Proportion
Ground truth
Conclusion (expression distribution)
Activation
Repression
Neutrality
Wrong prediction (both)
Wrong prediction (either one)
d
Distribution of target gene expression pre- and post-perturbation in pancreatic endogenesis (E14)
e
Regulation mode by expression distribution or cross-cell regulation proportion in pancreatic endogenesis (E14)
f
Distribution of target gene expression pre- and post-perturbation in pancreatic endogenesis (E17)
g
Regulation mode by expression distribution or cross-cell regulation proportion in pancreatic endogenesis (E17)

## Supplementary Figure 8: Pairwise regulation prediction by Tabula in pancreatic endogenesis system across developmental stages E12, E14, and E17.

a. Pancreatic endogenesis system at three developmental stages (E12, E14, E17) with extensively validated regulatory networks that Tabula applied to and its performance in pairwise regulation prediction [41] (similar to **Fig. 3b**).
b. Distribution of target gene expression pre- and post-perturbation under Tabula prediction in the pancreatic endogenesis core network at developmental stage E12 featuring 22 edges [41] (similar to **Fig. 3d**).
c. Comparison of regulation mode identified by expression distribution shift and cross-cell regulation proportions in pancreatic endogenesis system at developmental stage E12 [41] (similar to **Fig. 3e**).
d. Same as in panel b but for the developmental stage E14.
e. Same as in panel c but for the developmental stage E14.
f. Same as in panel b but for the developmental stage E17.
g. Same as in panel c but for the developmental stage E17.

**a** **Combinatorial regulation on cardiogenesis**

CR 2: FGF8 = ¬MESP1 ∧ (FOXC2 ∨ TBX1)

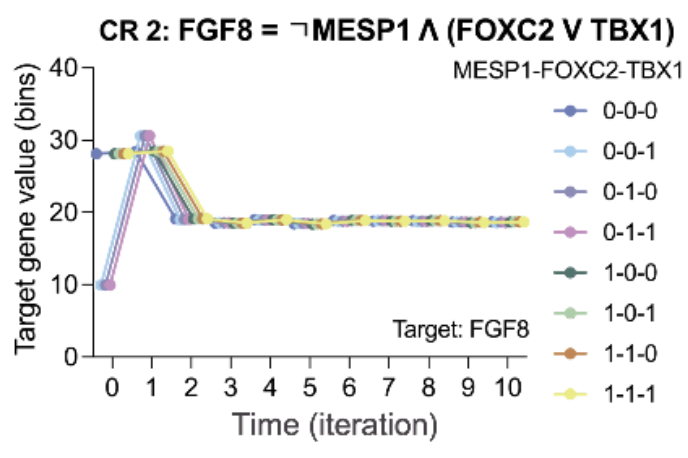

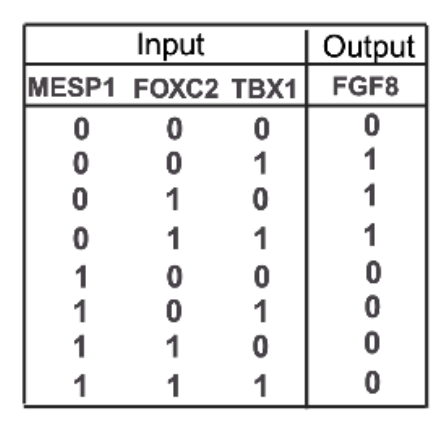

| Input | | | Output |
|---|---|---|---|
| MESP1 | FOXC2 | TBX1 | FGF8 |
| 0 | 0 | 0 | 0 |
| 0 | 0 | 1 | 1 |
| 0 | 1 | 0 | 1 |
| 0 | 1 | 1 | 1 |
| 1 | 0 | 0 | 0 |
| 1 | 0 | 1 | 0 |
| 1 | 1 | 0 | 0 |
| 1 | 1 | 1 | 0 |

CR 4: ISL1 = (TBX1 ∨ MESP1 ∨ FGF8 )

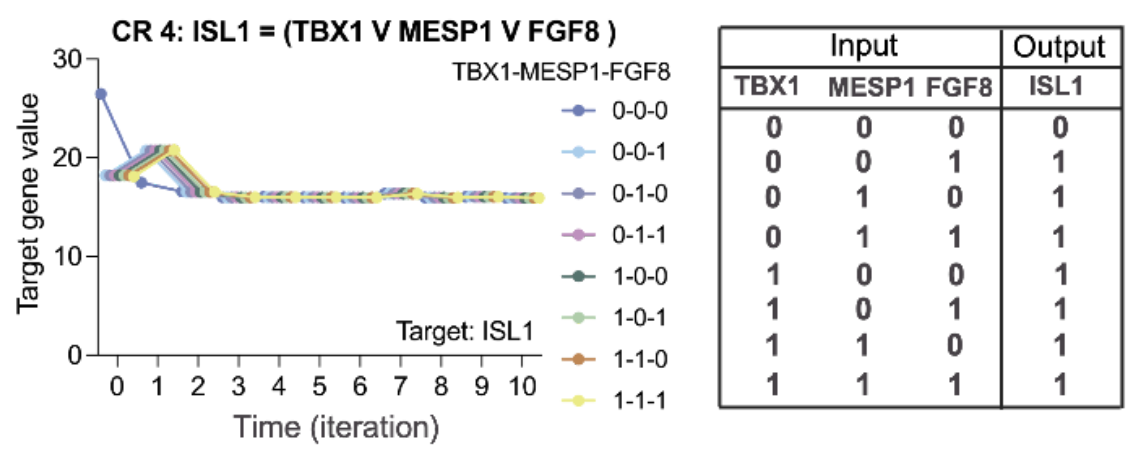

| Input | | | Output |
|---|---|---|---|
| TBX1 | MESP1 | FGF8 | ISL1 |
| 0 | 0 | 0 | 0 |
| 0 | 0 | 1 | 1 |
| 0 | 1 | 0 | 1 |
| 0 | 1 | 1 | 1 |
| 1 | 0 | 0 | 1 |
| 1 | 0 | 1 | 1 |
| 1 | 1 | 0 | 1 |
| 1 | 1 | 1 | 1 |

**b** **Combinatorial regulation on hematopoiesis**

CR 1: CEBPA = CEBPA ∧ ¬(GATA1 ∧ZFPM1 ∧TAL1)

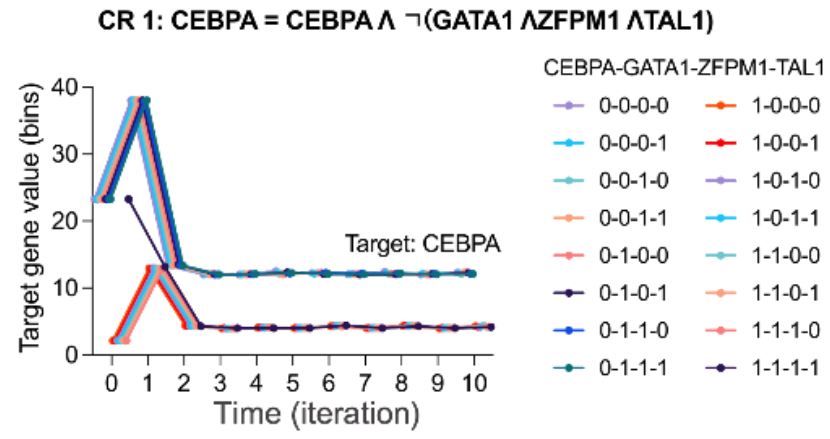

| Input | | | | Output |
|---|---|---|---|---|
| CEBPA | GATA1 | ZFPM1 | TAL1 | CEBPA |
| 0 | 0 | 0 | 0 | 0 |
| 0 | 0 | 0 | 1 | 0 |
| 0 | 0 | 1 | 0 | 0 |
| 0 | 0 | 1 | 1 | 0 |
| 0 | 1 | 0 | 0 | 0 |
| 0 | 1 | 0 | 1 | 0 |
| 0 | 1 | 1 | 0 | 0 |
| 0 | 1 | 1 | 1 | 0 |

| Input | | | | Output |
|---|---|---|---|---|
| CEBPA | GATA1 | ZFPM1 | TAL1 | CEBPA |
| 1 | 0 | 0 | 0 | 1 |
| 1 | 0 | 0 | 1 | 1 |
| 1 | 0 | 1 | 0 | 1 |
| 1 | 0 | 1 | 1 | 1 |
| 1 | 1 | 0 | 0 | 1 |
| 1 | 1 | 0 | 1 | 1 |
| 1 | 1 | 1 | 0 | 1 |
| 1 | 1 | 1 | 1 | 0 |

CR 2: SPI1 = (CEBPA ∨ SPI1) ∧ ¬(GATA1 ∨ GATA2)

| Input | | | | Output |
|---|---|---|---|---|
| CEBPA | SPI1 | GATA1 | GATA2 | SPI1 |
| 0 | 0 | 0 | 0 | 0 |
| 0 | 0 | 0 | 0 | 0 |
| 0 | 0 | 0 | 1 | 0 |
| 0 | 0 | 0 | 1 | 0 |
| 0 | 1 | 0 | 0 | 1 |
| 0 | 1 | 0 | 0 | 1 |
| 0 | 1 | 0 | 1 | 0 |
| 0 | 1 | 0 | 1 | 0 |

| Input | | | | Output |
|---|---|---|---|---|
| CEBPA | SPI1 | GATA1 | GATA2 | SPI1 |
| 1 | 0 | 1 | 0 | 0 |
| 1 | 0 | 1 | 0 | 0 |
| 1 | 0 | 1 | 1 | 0 |
| 1 | 0 | 1 | 1 | 0 |
| 1 | 1 | 1 | 0 | 0 |
| 1 | 1 | 1 | 0 | 0 |
| 1 | 1 | 1 | 1 | 0 |
| 1 | 1 | 1 | 1 | 0 |

CR 3: GATA2 = GATA2 ∧ ¬(GATA1 ∧ ZFPM1) ∧ ¬SPI1

| Input | | | | Output |
|---|---|---|---|---|
| GATA2 | GATA1 | ZFPM1 | SPI1 | GATA2 |
| 0 | 0 | 0 | 0 | 0 |
| 0 | 0 | 0 | 1 | 0 |
| 0 | 0 | 1 | 0 | 0 |
| 0 | 0 | 1 | 1 | 0 |
| 0 | 1 | 0 | 0 | 0 |
| 0 | 1 | 0 | 1 | 0 |
| 0 | 1 | 1 | 0 | 0 |
| 0 | 1 | 1 | 1 | 0 |

| Input | | | | Output |
|---|---|---|---|---|
| GATA2 | GATA1 | ZFPM1 | SPI1 | GATA2 |
| 1 | 0 | 0 | 0 | 1 |
| 1 | 0 | 0 | 1 | 0 |
| 1 | 0 | 1 | 0 | 1 |
| 1 | 0 | 1 | 1 | 0 |
| 1 | 1 | 0 | 0 | 1 |
| 1 | 1 | 0 | 1 | 0 |
| 1 | 1 | 1 | 0 | 0 |
| 1 | 1 | 1 | 1 | 0 |

CR 4: GATA1 = (GATA1 ∨ GATA2 ∨ FLI1) ∧ ¬SPI1

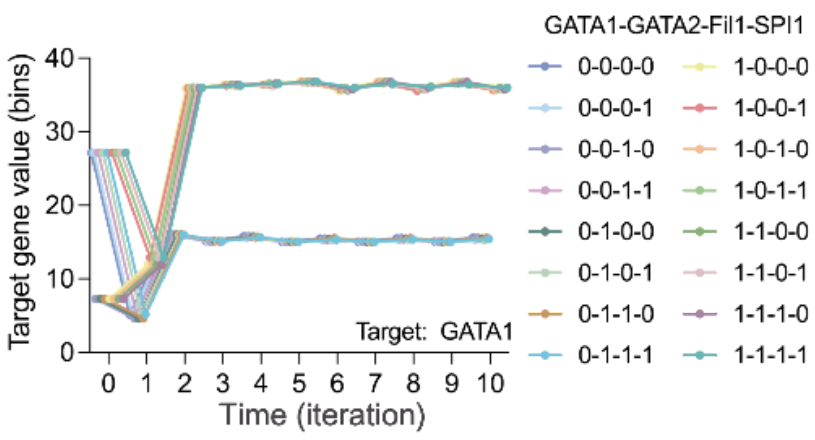

| Input | | | | Output |
|---|---|---|---|---|
| GATA1 | GATA2 | FLI1 | SPI1 | GATA1 |
| 0 | 0 | 0 | 0 | 0 |
| 0 | 0 | 0 | 1 | 0 |
| 0 | 0 | 1 | 0 | 1 |
| 0 | 0 | 1 | 1 | 0 |
| 0 | 1 | 0 | 0 | 1 |
| 0 | 1 | 0 | 1 | 0 |
| 0 | 1 | 1 | 0 | 1 |
| 0 | 1 | 1 | 1 | 0 |

| Input | | | | Output |
|---|---|---|---|---|
| GATA1 | GATA2 | FLI1 | SPI1 | GATA1 |
| 1 | 0 | 0 | 0 | 1 |
| 1 | 0 | 0 | 1 | 0 |
| 1 | 0 | 1 | 0 | 1 |
| 1 | 0 | 1 | 1 | 0 |
| 1 | 1 | 0 | 0 | 1 |
| 1 | 1 | 0 | 1 | 0 |
| 1 | 1 | 1 | 0 | 1 |
| 1 | 1 | 1 | 1 | 0 |

CR 7: FLI1 = GATA1 ∧ ¬EKLF

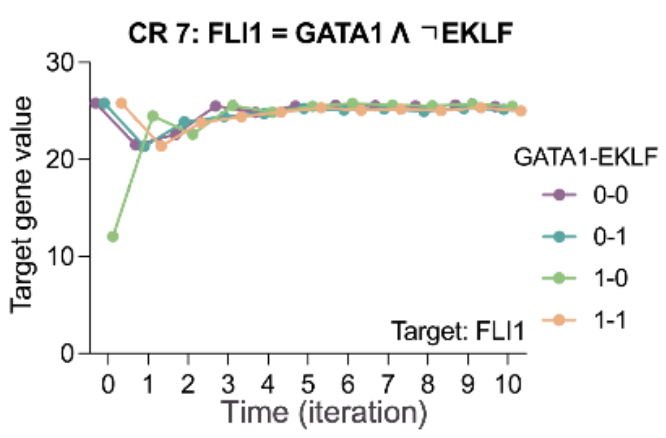

| Input | | Output |
|---|---|---|
| GATA1 | EKLF | FLI1 |
| 0 | 0 | 0 |
| 0 | 1 | 0 |
| 1 | 0 | 1 |
| 1 | 1 | 0 |

## Supplementary Figure 9: Combinatorial regulation predictions by Tabula for cardiogenesis and hematopoiesis systems.

a. Combinatorial regulation (CR) predictions of CR2 and CR4 for the cardiogenesis system [97] (similar to **Fig. 3f**).
b. Combinatorial regulation (CR) predictions of CR1-4 and CR7 for the hematopoiesis system [35] (similar to **Fig. 3g**).

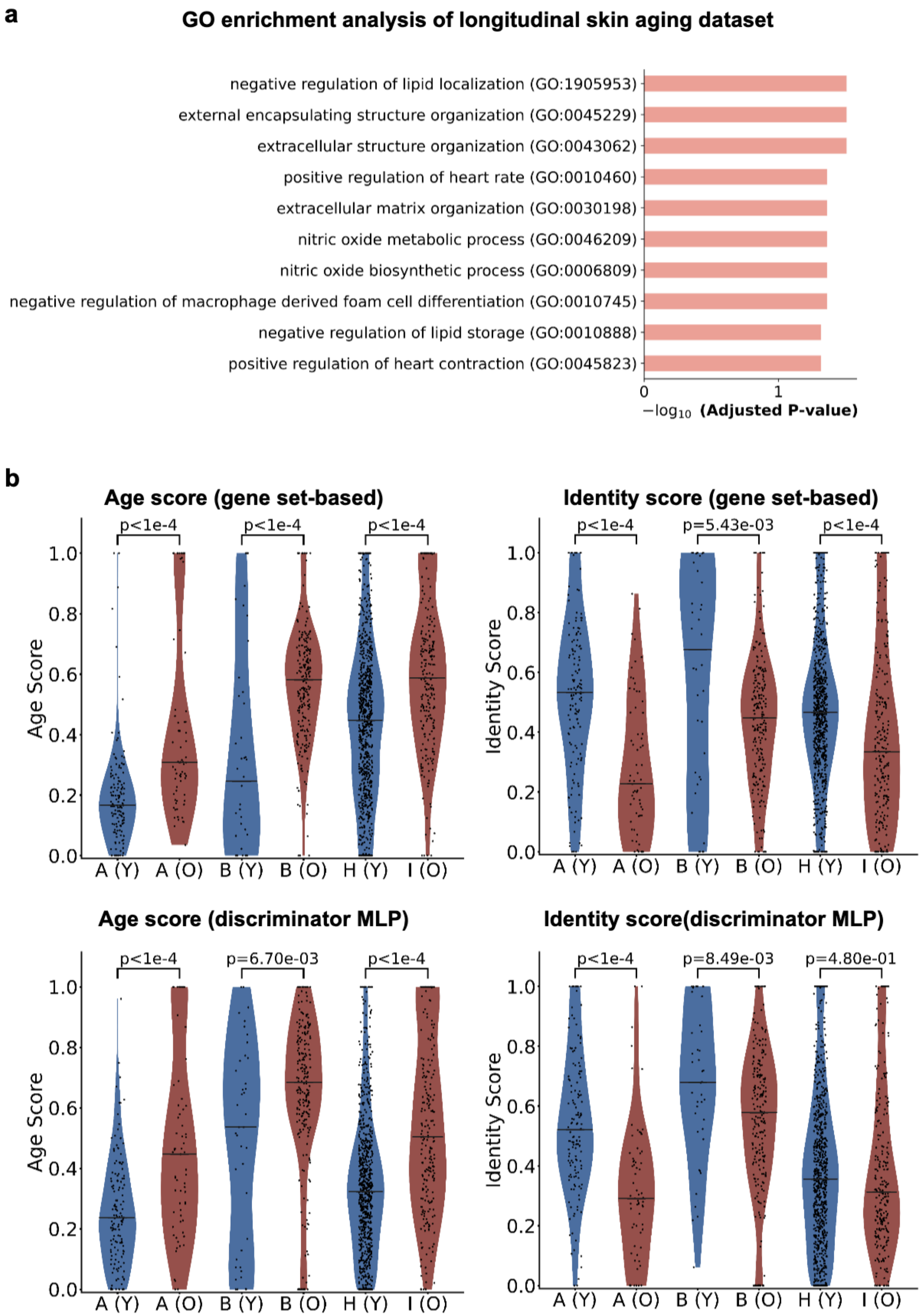

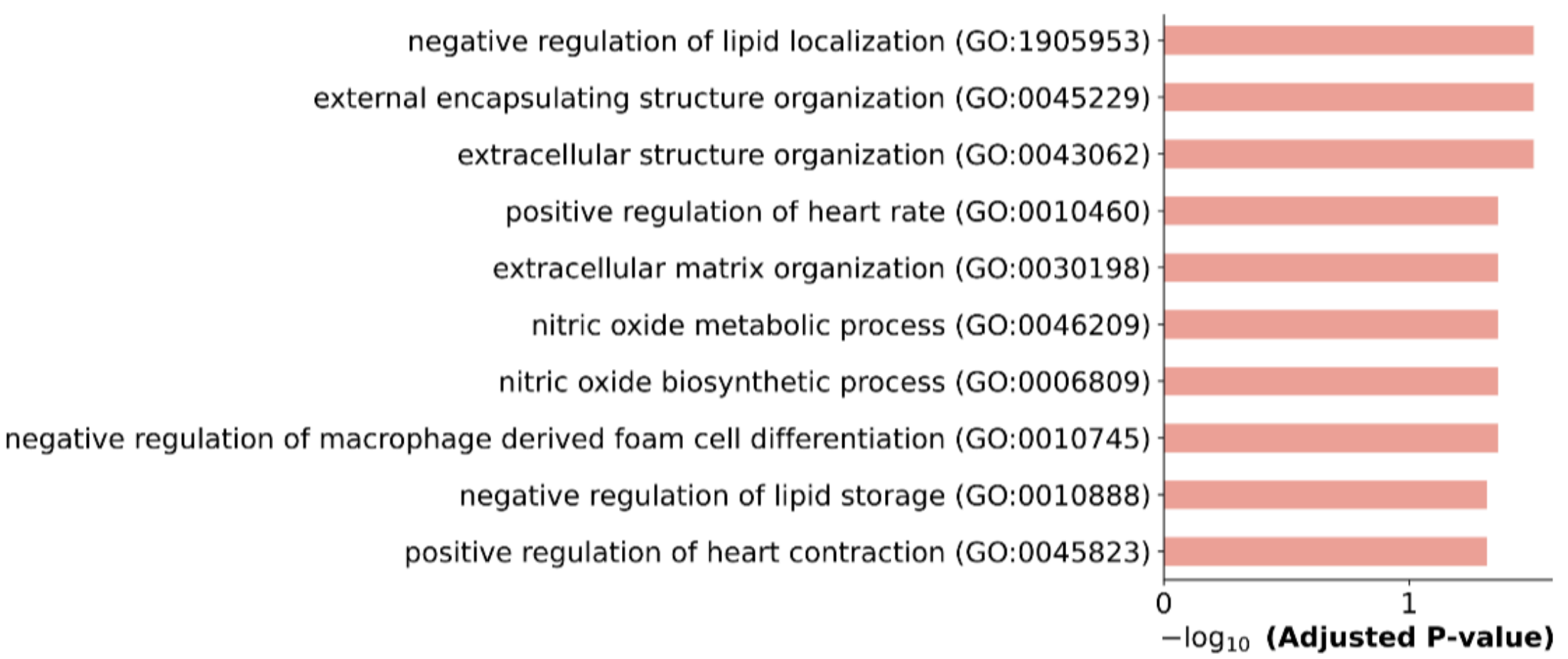

b

Age score (gene set-based)

p<1e-4 p<1e-4 p<1e-4

Age Score

A (Y) A (O) B (Y) B (O) H (Y) I (O)

Identity score (gene set-based)

p<1e-4 p=5.43e-03 p<1e-4

Identity Score

A (Y) A (O) B (Y) B (O) H (Y) I (O)

Age score (discriminator MLP)

p<1e-4 p=6.70e-03 p<1e-4

Age Score

A (Y) A (O) B (Y) B (O) H (Y) I (O)

Identity score(discriminator MLP)

p<1e-4 p=8.49e-03 p=4.80e-01

Identity Score

A (Y) A (O) B (Y) B (O) H (Y) I (O)

## Supplementary Figure 10: Construction and validation of age and identity scores in the longitudinal skin aging dataset.

a. Gene Ontology Biological Process enrichment analysis of the predefined age-associated gene set used for score construction in the longitudinal skin aging dataset. The top enriched terms include extracellular structure organization, extracellular matrix organization, nitric oxide metabolic process, and lipid-related regulatory processes. Bar length indicates adjusted p-value.
b. Age and identity score validation and MLP scoring in Tabula's reconstruction of gene expressions in fibroblast groups. Violin plots illustrate age and identity scores calculated from normalized gene expression using predefined gene sets, with all scores further rescaled to the 0–1 range. Aged cells consistently show elevated age scores and reduced identity scores relative to young cells across groups, supporting the use of these two scores as optimization objectives for identifying candidate rejuvenation factors. P values are shown above each comparison.

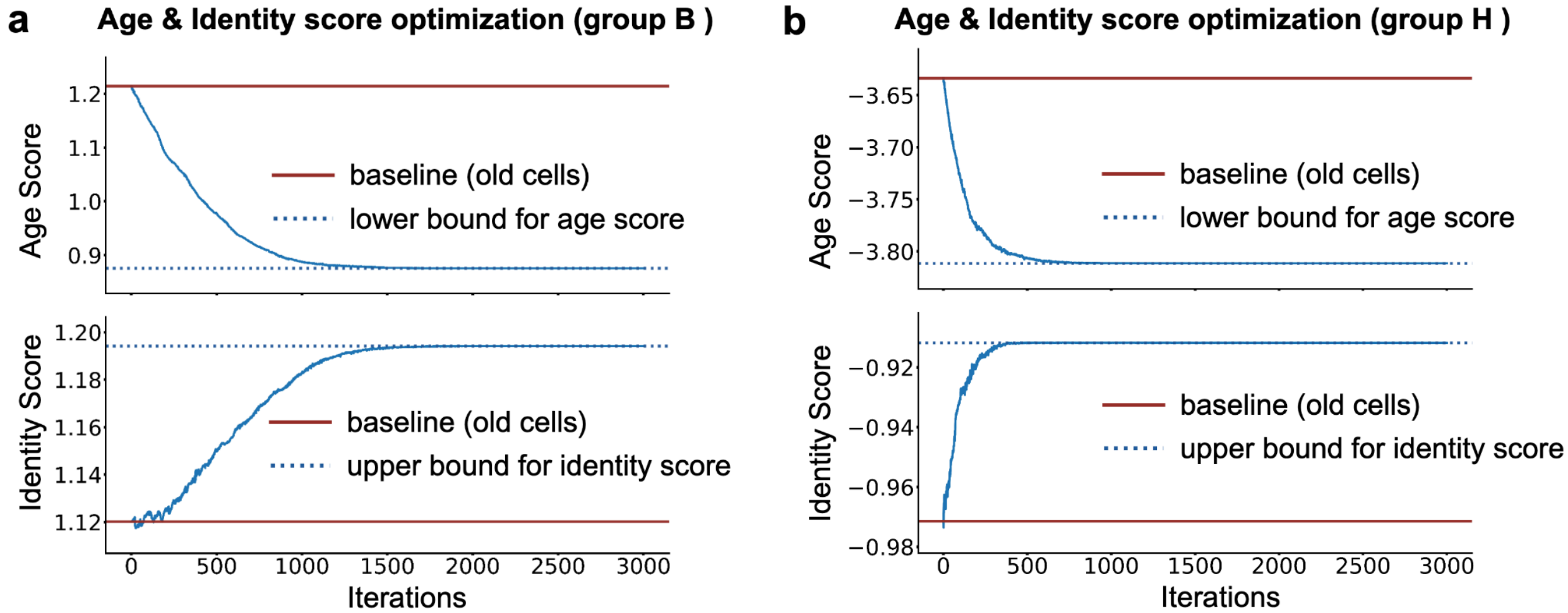

Supplementary Figure 11: Optimization of age and identity scores during Tabula-guided in silico reconstruction in fibroblast group B/H.

a. In fibroblast group B, the top panel shows that the age score progressively decreases over iterations from the baseline of old cells toward the predefined lower boundary, whereas the bottom panel shows that the identity score correspondingly increases toward the upper boundary defined by young cells.
b. The same analysis in fibroblast group H, the top panel shows that the age score progressively decreases over iterations from the baseline of old cells toward the predefined lower boundary, whereas the bottom panel shows that the identity score correspondingly increases toward the upper boundary defined by young cells.

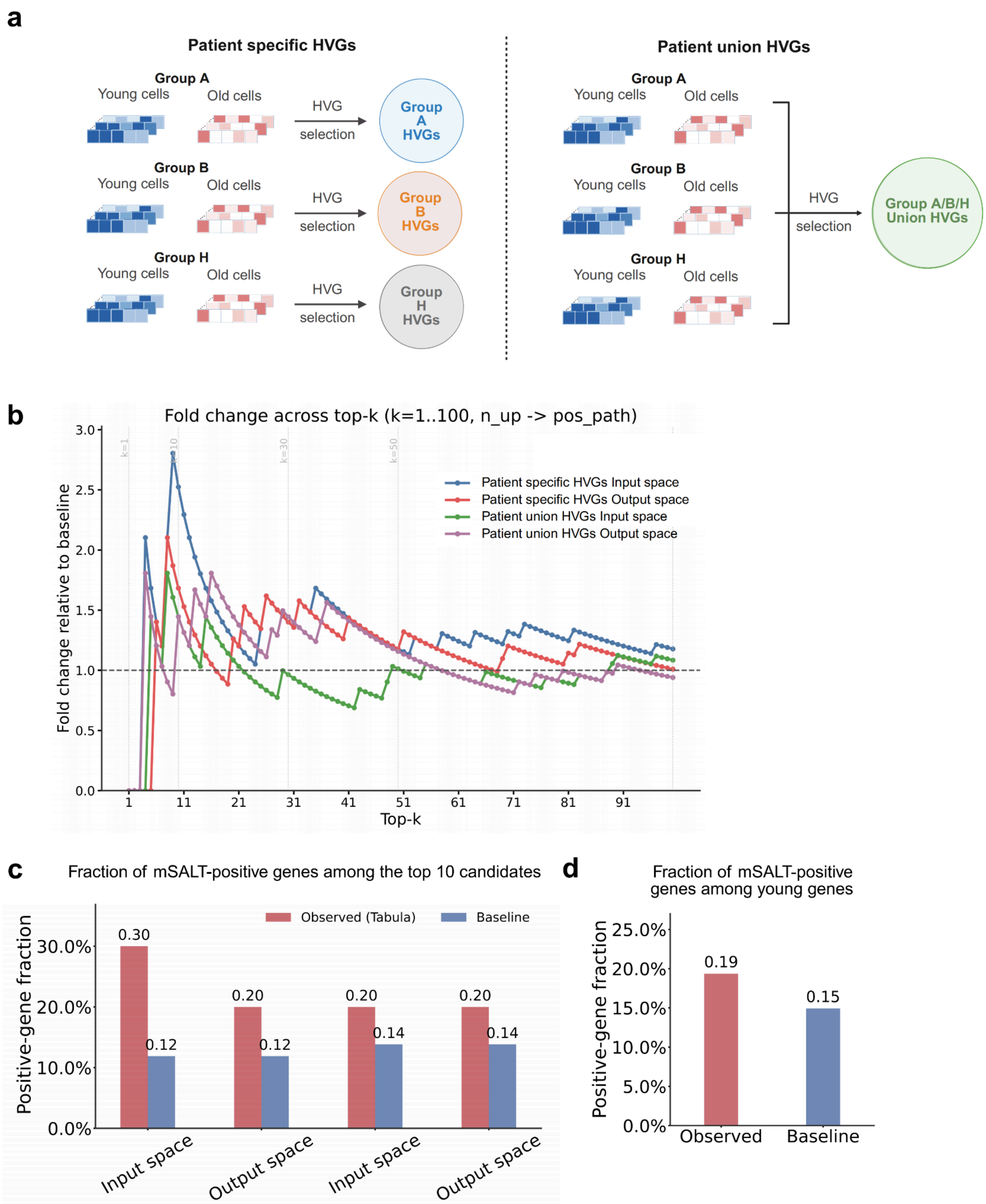

Supplementary Figure 12: Validation of Tabula-identified rejuvenation factors using the mSALT reference.

a. Schematic illustration of the two HVG selection strategies used for candidate identification. In the Patient specific HVGs setting, highly variable genes (HVGs) were selected separately within each matched young–old group, yielding group-specific HVG sets for groups A, B, and H. In the UnionHVG setting, cells from groups A, B, and H were pooled to define a shared HVG set used across all three groups.
b. Fold enrichment of mSALT-positive genes among the top-k Tabula-ranked candidates under different Tabula settings. mSALT-positive genes were defined using the mSALT-derived reference set[45] (slope > 0 and P < 0.05). Curves show the fold change relative to the matched baseline as k varies from 1 to 100, with genes ranked by upregulatory frequency (n_up) and ties broken by positive path accumulation score (pos_path). The baseline was calculated as the expected fraction of mSALT-positive genes in randomly selected, size-matched gene sets drawn from the corresponding HVG pool for each Tabula settingResults are shown for Patient specific HVGs Input space, Patient specific HVGs Output space, Patient union HVGs Input space, and Patient union HVGs Output space. The dashed horizontal line indicates baseline expectation (1×).
c. Fraction of mSALT-positive genes among the top 10 Tabula-ranked candidates, compared with the corresponding baseline fraction. The baseline was calculated as the expected fraction of mSALT-positive genes in randomly selected, size-matched gene sets drawn from the corresponding HVG pool for each Tabula setting. Bars show the observed fraction from Tabula and the matched baseline fraction for Patient specific HVGs and Patient union HVGs in both input and output spaces.
d. Fraction of mSALT-positive genes among 48 young genes (**Supplementary Table 1**). The observed fraction derived from the inferred young gene set is compared with the expected fraction of mSALT-positive genes in randomly selected, size-matched gene set drawn from the whole transcriptome.

**a**

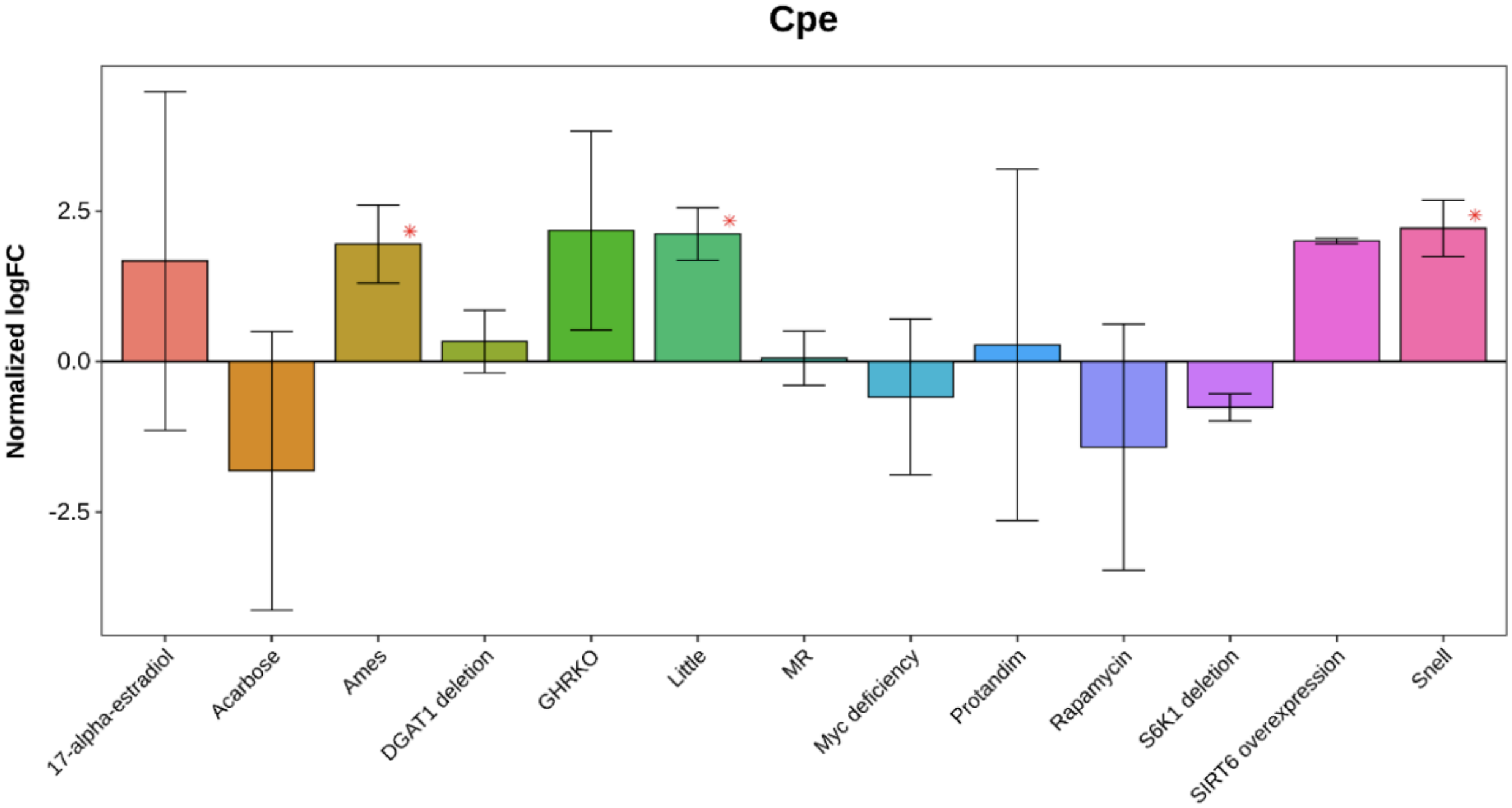

Cpe
Normalized logFC
2.5
0.0
-2.5
17-alpha-estradiol
Acarbose
Ames
DGAT1 deletion
GHRKO
Little
MR
Myc deficiency
Protandim
Rapamycin
S6K1 deletion
SIRT6 overexpression
Snell

**b**

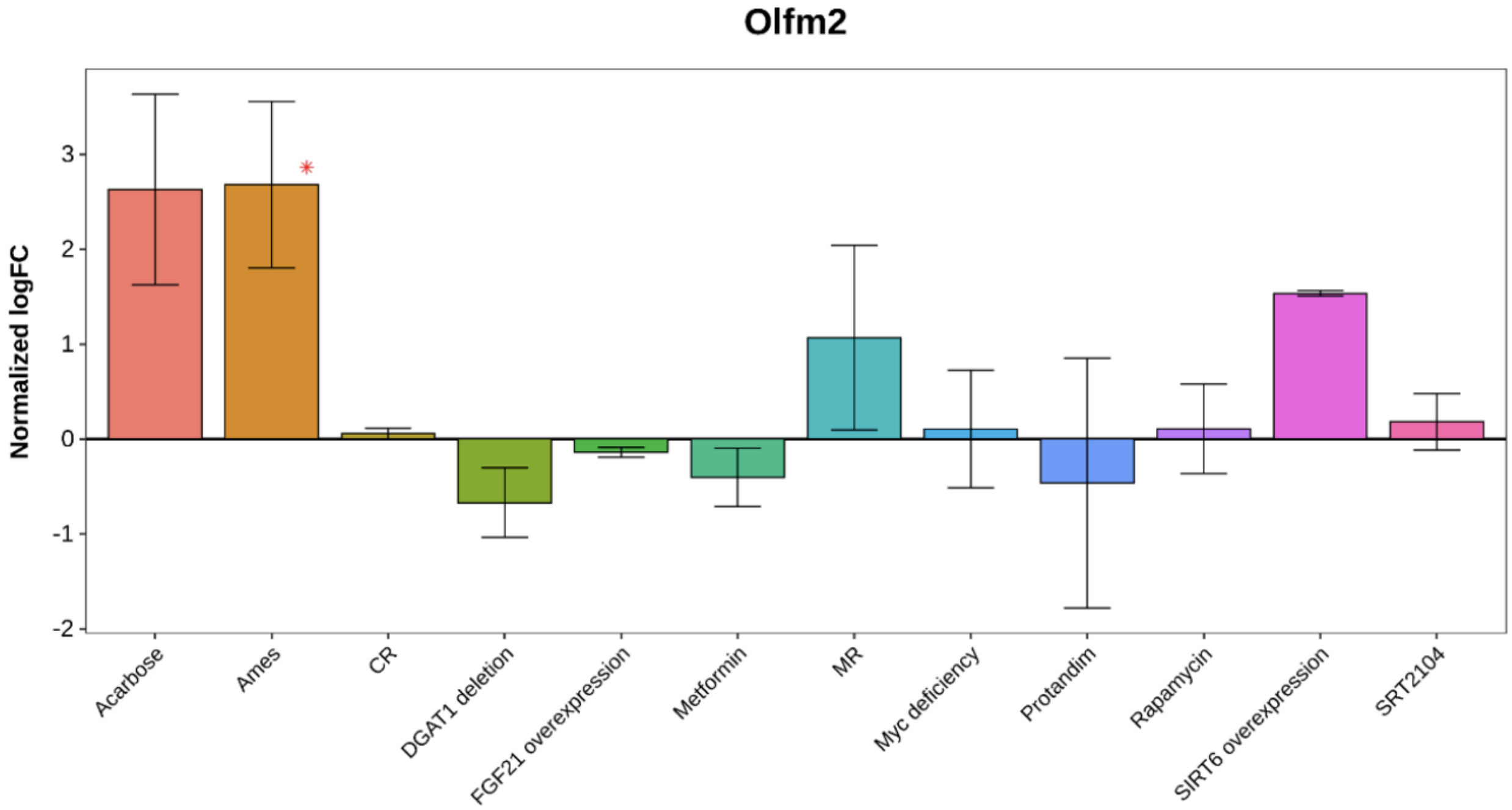

Olfm2
Normalized logFC
3
2
1
0
-1
-2
Acarbose
Ames
CR
DGAT1 deletion
FGF21 overexpression
Metformin
MR
Myc deficiency
Protandim
Rapamycin
SIRT6 overexpression
SRT2104

## Supplementary Figure 13: Expression responses of candidate rejuvenation genes across lifespan-extending interventions.

a. Aggregated normalized log fold change (logFC) of *Cpe* in mouse liver across multiple lifespan- and healthspan-extending interventions. Bar height indicates the aggregated normalized logFC for each intervention, and error bars indicate the corresponding variation across datasets. Red asterisks denote interventions with adjusted p-values below 0.05.
b. Aggregated normalized log fold change (logFC) of *Olfm2* in mouse liver across multiple lifespan- and healthspan-extending interventions. Bar height indicates the aggregated normalized logFC for each intervention, and error bars indicate the corresponding variation across datasets. Red asterisks denote interventions with adjusted p-values below 0.05.

**a**

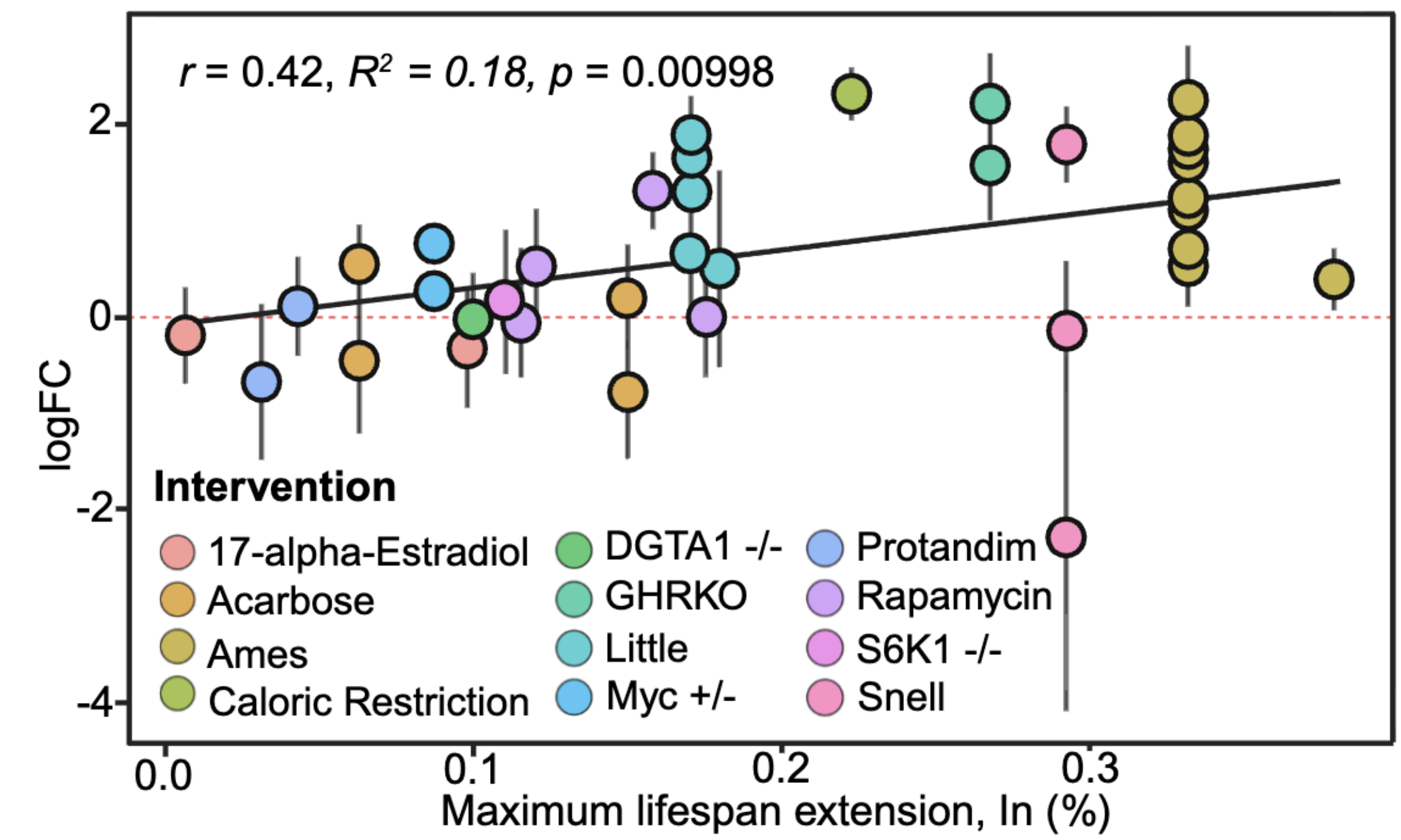
Postn
r = 0.42, $R^2$ = 0.18, p = 0.00998
logFC
2
0
-2
-4
Intervention
17-alpha-Estradiol
Acarbose
Ames
Caloric Restriction
DGTA1 -/-
GHRKO
Little
Myc +/-
Protandim
Rapamycin
S6K1 -/-
Snell
0.0
0.1
0.2
0.3
Maximum lifespan extension, ln (%)

**b**

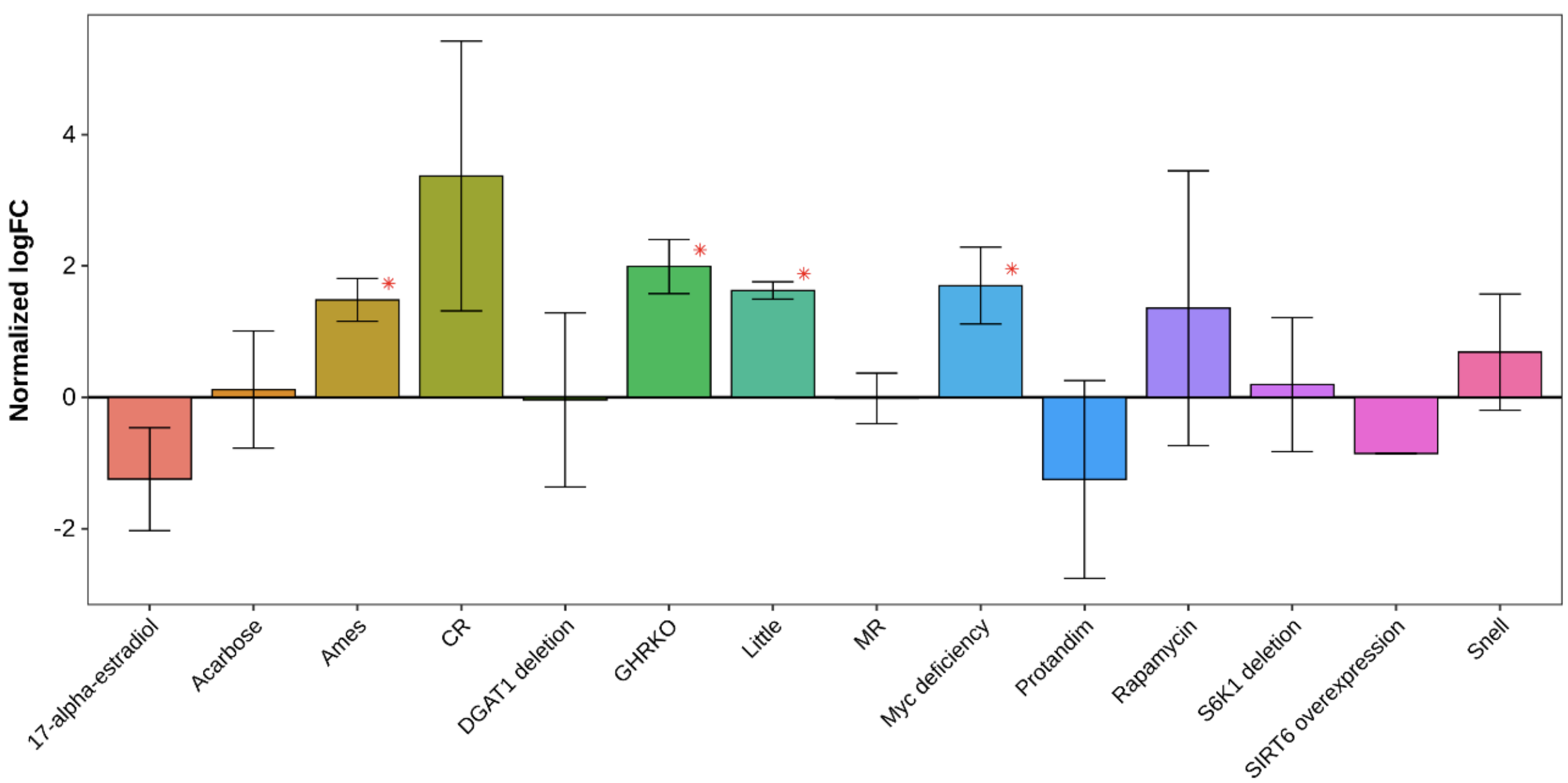
Postn
Normalized logFC
4
2
0
-2
17-alpha-estradiol
Acarbose
Ames
CR
DGAT1 deletion
GHRKO
Little
MR
Myc deficiency
Protandim
Rapamycin
S6K1 deletion
SIRT6 overexpression
Snell

## Supplementary Figure 14: Association of the candidate gene Postn with lifespan extension effects across independent mouse longevity interventions based on the mSALT database.

a. Each dot represents a single dataset and corresponds to the mean hepatic log fold change of *Postn* under a specific intervention, with error bars indicating standard errors. The x-axis denotes the maximum lifespan extension achieved by the intervention. *Postn* expression change is positively associated with lifespan extension effect ($r = 0.42$, $R^2 = 0.18$, $p = 0.00998$).
b. Aggregated normalized log fold change (logFC) of *Postn* in mouse liver across multiple lifespan- and healthspan-extending interventions. Bar height indicates the aggregated normalized logFC for each intervention, and error bars indicate the corresponding variation across datasets. Red asterisks denote interventions with adjusted *P* values below 0.05.

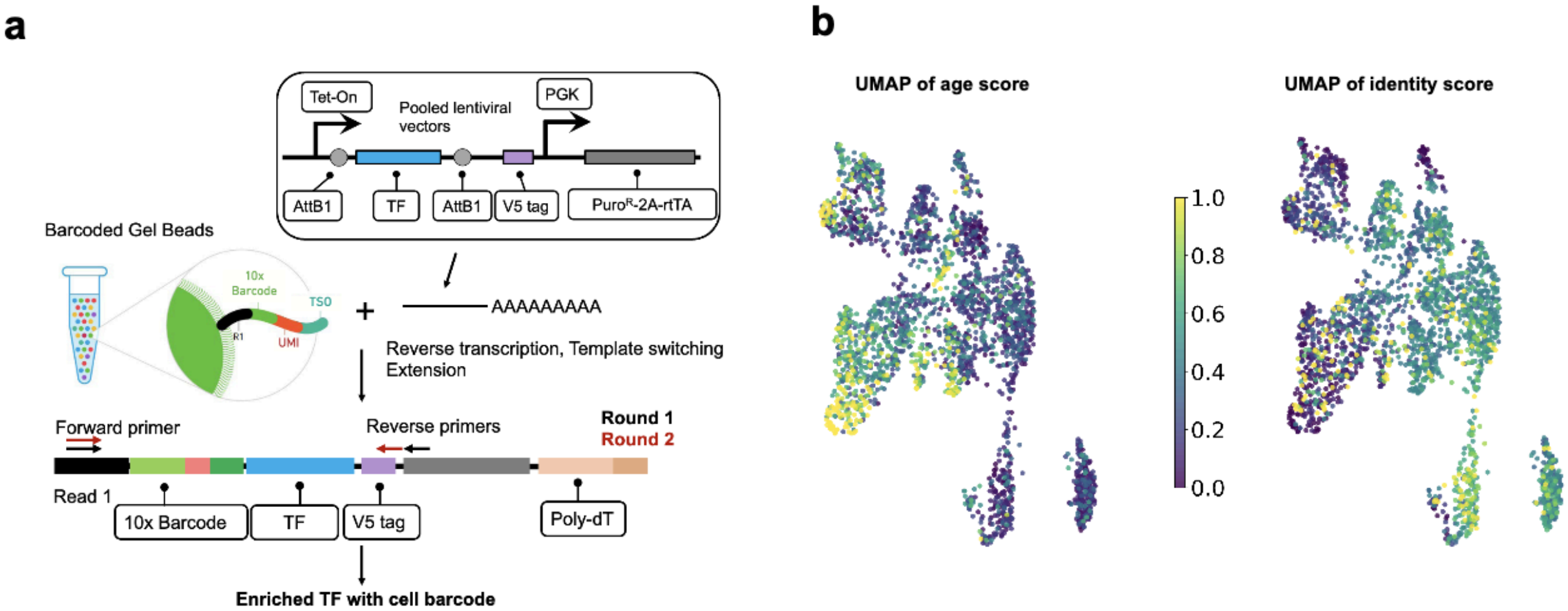

# Supplementary Figure 15: Perturb-seq design and score projection for reprogramming-factor perturbations in aged fibroblasts.

a. Schematic illustration of the inducible overexpression vector and Perturb-seq workflow used to quantify the effects of single and combinatorial reprogramming factors. A pooled lentiviral vector library carrying transcription factor (TF) cassettes was introduced into aged fibroblasts, and TF identity was captured together with single-cell transcriptomes through a Perturb-seq-compatible barcode design, enabling simultaneous measurement of genetic perturbations and their transcriptional consequences.

b. UMAP visualization of perturbed fibroblast cells colored by normalized age score (left) and normalized identity score (right). Scores were projected onto the single-cell manifold to summarize the transcriptional effects of perturbations in the two-dimensional space defined by rejuvenation-associated and fibroblast identity-associated transcriptional programs.

| Gene set category | Gene symbols |
| --- | --- |
| Young gene set (n=48) | *RPS29, CLEC2B, ITGA11, COL8A1, PCSK7, ANTXR1, VCL, ADM, PHLDA1, NQO1, COMP, HSPB7, VIM-AS1, CLIC3, EIF5B, SSTR1, SAMD9, TCEAL4, LINC01705, NEXN, HIST1H1C, NFKBIA, TCEAL3, CREB3, OLFML3, AC011603.2, CD302, GADD45B, H1F0, CTHRC1, MEG3, EDN1, ADIRF, ITGAV, MALL, ARL6IP1, SLC9A3R2, TMX4, SYNE2, SPON2, NREP, HMGB2, SOD2, CAPRIN2, ELN, MT-ND3, GADD45A, DDAH2* |
| Age gene set (n=9) | *NDUFA4L2, PI16, STMN2, NOVA1, RARRES2, MT1M, SFRP1, PTGDS, SLC22A18AS* |
| Fibroblast gene set (n=19) | *COL3A1, FSP-1, TGFB3, TGFB2, COL1A2, ITGA1, DDR2, P4HA3, THY1, FAP, CD248, VIM, COL1A1, ITGA5, P4HA1, P4HA2, TGFB1, HSP47, CD34* |

Supplementary Table 1: Gene sets used for age and identity score construction.

# SUPPLEMENTARY NOTES

## S1 Evaluation metrics for cell type annotation

To evaluate the performance of cell type annotation tasks, we use standard classification metrics: accuracy, precision, recall, and the F1 score. These metrics are defined as follows:

$$\text{Accuracy} = \frac{TP + TN}{TP + TN + FP + FN},$$

where $TP$ is the number of true positives, $TN$ is the number of true negatives, $FP$ is the number of false positives, and $FN$ is the number of false negatives.

$$\text{Precision} = \frac{TP}{TP + FP},$$

where precision quantifies the proportion of correctly predicted positive cell types among all predicted positives.

$$\text{Recall} = \frac{TP}{TP + FN},$$

where recall measures the ability to identify all true positive instances of a cell type.

$$\text{F1 Score} = 2 \cdot \frac{\text{Precision} \cdot \text{Recall}}{\text{Precision} + \text{Recall}},$$

where F1 score provides a balanced measure that accounts for both precision and recall, especially useful for imbalanced datasets.

## S2 Evaluation metrics for multi-omics and multi-batch integration

To evaluate the performance of integration tasks, we adopt the widely used evaluation pipeline in the field [74] and selectively choose five metrics as conducted by scGPT [17]. For multi-omics integration, metrics that measure conservation of biological variance of cell type, including $NMI_{cell}$, $ARI_{cell}$, and $ASW_{cell}$, are considered. For multi-batch integration, graph connectivity ($Graph_Conn$) and $ASW_{batch}$ are used for evaluation.

NMI (Normalized Mutual Information) measures the agreement between predicted and true cluster labels, providing an indication of how well cell types are preserved post-integration. It ranges from 0 (no agreement) to 1 (perfect agreement).

ARI (Adjusted Rand Index) evaluates the similarity between clustering results and true cell type annotations, adjusted for random chance. A higher ARI indicates better preservation of biological groupings.

ASW (Average Silhouette Width) quantifies the separation of cells within their assigned clusters compared to other clusters. A higher ASW indicates better-defined and more biologically meaningful clusters.

The batch-specific ASW evaluates how well batch effects are removed while maintaining meaningful biological groupings. A lower batch ASW indicates better integration across different batches.

Graph connectivity measures the continuity of the integrated data in a shared embedding space. Higher connectivity reflects improved integration, ensuring that cells from the same biological group form a cohesive cluster regardless of batch origin.

For systematically assessing the performance of integration tasks, $AvgBIO$ and $Avg_{batch}$ are employed to evaluate multi-omics integration and multi-batch integration, respectively. Additionally, $Overall$ is used to measure the comprehensive performance of both tasks when the dataset is used to simultaneously evaluate the two tasks. The formulas are defined as follows:

$$AvgBIO = \frac{\mathrm{ARI_{cell}} + \mathrm{NMI_{cell}} + \mathrm{ASW_{cell}}}{3}$$

$$Avg_{batch} = \frac{ASW_{batch} + Graph_Conn}{2}$$

$$Overall = 0.6 * AvgBIO + 0.4 * Avg_{batch}$$

## S3 Evaluation metrics for genetic perturbation prediction

To evaluate the performance of genetic perturbation prediction tasks, we use Pearson correlation coefficient, which quantifies the linear relationship between predicted and actual expression changes following perturbation. The Pearson coefficient computed across all predicted gene expression changes for a cell is formulated as follows:

$$Pearson_{delta} = \frac{\sum_{i=1}^{n}(x_i - \bar{x})(y_i - \bar{y})}{\sqrt{\sum_{i=1}^{n}(x_i - \bar{x})^2}\sqrt{\sum_{i=1}^{n}(y_i - \bar{y})^2}}$$

Here, $x_i$ and $y_i$ represent the predicted and actual gene expression changes for gene $i$, respectively. $\bar{x}$ and $\bar{y}$ are the mean values of the predicted and actual expression changes across all genes. $n$ is the total number of genes.

Similar to $Pearson_{delta}$, the Pearson coefficient computed across top 20 differentially expressed gene expression changes for a cell, $\mathrm{DEG\ Pearson}_{delta}$, is formulated as follows:

$$\mathrm{DEG\ Pearson}_{delta} = \frac{\sum_{i=1}^{20}(x_i - \bar{x})(y_i - \bar{y})}{\sqrt{\sum_{i=1}^{20}(x_i - \bar{x})^2}\sqrt{\sum_{i=1}^{20}(y_i - \bar{y})^2}}.$$

Here, $x_i$ and $y_i$ represent the predicted and actual gene expression changes for the top 20 differentially expressed genes (DEGs), respectively. $\bar{x}$ and $\bar{y}$ denote the mean values of the predicted and actual expression changes for these DEGs.

## S4 Evaluation metrics for imputation

To evaluate the performance of imputation tasks, we use multiple metrics, including the Pearson correlation coefficient, Root Mean Squared Error (RMSE), and Mean Absolute Error (MAE). The Pearson correlation coefficient quantifies the linear relationship between imputed and true gene expression values, while RMSE and MAE measure the differences between them. all metrics are computed across all imputed gene expression values and formulated as follows:

$$Pearson = \frac{\sum_{i=1}^{N}(x_i - \bar{x})(y_i - \bar{y})}{\sqrt{\sum_{i=1}^{N}(x_i - \bar{x})^2}\sqrt{\sum_{i=1}^{N}(y_i - \bar{y})^2}}$$

$$RMSE = \sqrt{\frac{1}{N}\sum_{i=1}^{N}(x_i - y_i)^2}$$

$$MAE = \frac{1}{N}\sum_{i=1}^{N}|x_i - y_i|$$

Here, $x_i$ and $y_i$ represent the imputed and true gene expression values for masked gene $i$, respectively. $\bar{x}$ and $\bar{y}$ are the mean values of the imputed and true gene expression values across all masked genes. $N$ denotes the total number of masked genes across all cells.

To evaluate the consistency between the original and imputed data, we calculate the Mean Squared Error (MSE) at two levels: cell-cell and gene-gene. The definitions are as follows:
$MSE_{cell-cell}$ measures the differences between the imputed and true values for all masked genes within a given cell. The formula is:

$$MSE_{cell-cell} = \frac{1}{m}\sum_{j=1}^{m}(x_j - y_j)^2$$

Here, $x_j$ and $y_j$ represent the imputed and true gene expression values for masked gene $j$ in a given cell. $m$ is the total number of masked genes in this cell.

Similar to $MSE_{cell-cell}$, $MSE_{gene-gene}$ measures the differences between the imputed and true gene expression values for all masked cells within a given gene. The formula is:

$$MSE_{gene-gene} = \frac{1}{n}\sum_{i=1}^{n}(x_i - y_i)^2$$

Here, $x_i$ and $y_i$ represent the imputed and true gene expression values for masked cell $i$ in a given gene. $n$ is the total number of masked cells in this gene.